\documentclass[journal]{IEEEtran}
\RequirePackage{silence}
\usepackage{times}
\usepackage{epsfig}
\usepackage{graphicx}
\usepackage{amsmath}
\usepackage{amssymb}
\usepackage{array}
\usepackage{multirow}
\usepackage{subcaption}
\usepackage{color}
\usepackage{float}
\usepackage{graphicx}
\usepackage{mathtools}
\usepackage{booktabs}
\usepackage{url}
\usepackage{cite}
\usepackage{hyperref}
\usepackage{mathtools}
\usepackage{listings}
\usepackage{xcolor}
\usepackage{colortbl}
\usepackage{array}
\usepackage[font=scriptsize]{caption}  
\usepackage[norelsize,linesnumbered,longend,ruled,lined,boxed,commentsnumbered]{algorithm2e}
\usepackage{algpseudocode}
\newcases{Rcases}{$\mskip\thickmuskip$}{%
  \hfil$\m@th{##}$}{$\m@th{##}$\hfil}{.}{\rbrace}

\usepackage{tikz}
\usetikzlibrary{matrix,positioning,fit,backgrounds,intersections}
\makeatletter
\newcommand\HUGES{\@setfontsize\Huge{14}{25}}
\newcommand\HUGESS{\@setfontsize\Huge{24}{45}}
\newcommand\HUGESSS{\@setfontsize\Huge{15}{45}}
\newcommand\HUGESSSS{\@setfontsize\Huge{30}{90}}
\makeatother    

\usepackage[normalem]{ulem}
\definecolor{darkgreen}{rgb}{0,.4,0}
\definecolor{darkcyan}{rgb}{0,.4,.4}
\newcommand{\REMOVE}[1]%
          {{\color{blue}\sout{#1}}}
\newcommand{\ADD}[1]{{\color{red}{#1}}}

\newcommand{\COMMENT}[1]%
          {{\color{darkgreen}\textbf{{Editor: }} {#1}}}

\definecolor{Tangdaowan_data_Rubbertrack!}{RGB}{255   0   0}
\definecolor{Tangdaowan_data_Flaggingv!}{RGB}{ 0, 255,   0}
\definecolor{Tangdaowan_data_Sandy!}{RGB}{ 0,   0 ,255}
\definecolor{Tangdaowan_data_Asphalt!}{RGB}{255 ,255,   0}
\definecolor{Tangdaowan_data_Boardwalk!}{RGB}{255 ,  0,255}
\definecolor{Tangdaowan_data_Rockyshallows!}{RGB}{ 0 255 255}
\definecolor{Tangdaowan_data_Grassland!}{RGB}{200, 100 ,  0}
\definecolor{Tangdaowan_data_Bulrush!}{RGB}{ 0, 200, 100}
\definecolor{Tangdaowan_data_Gravelroad!}{RGB}{100 ,  0, 200}
\definecolor{Tangdaowan_data_Ligustrumvicaryi!}{RGB}{200 ,  0, 100}
\definecolor{Tangdaowan_data_Coniferouspine!}{RGB}{100 ,200,   0}
\definecolor{Tangdaowan_data_Spiraea!}{RGB}{  0 ,100, 200}
\definecolor{Tangdaowan_data_Baresoil!}{RGB}{150 , 75 , 75}
\definecolor{Tangdaowan_data_Buxussinica!}{RGB}{ 75 ,150 , 75}
\definecolor{Tangdaowan_data_Photiniaserrulata!}{RGB}{75 , 75 ,150}
\definecolor{Tangdaowan_data_Populus!}{RGB}{255, 100, 100}
\definecolor{Tangdaowan_data_UlmuspumilaL!}{RGB}{100, 255, 100}
\definecolor{Tangdaowan_data_Seawater!}{RGB}{100 ,100, 255}

\definecolor{Qingyun_data_Trees!}{RGB}{255,   0 ,  0}
\definecolor{Qingyun_data_Concretebuilding!}{RGB}{ 0 ,255,   0}
\definecolor{Qingyun_data_Car!}{RGB}{ 0 ,  0 ,255}
\definecolor{Qingyun_data_Ironhidebuilding!}{RGB}{255, 255 ,  0}
\definecolor{Qingyun_data_Plasticplayground!}{RGB}{255 ,  0 ,255}
\definecolor{Qingyun_data_Asphaltroad!}{RGB}{0, 255, 255}

\definecolor{Pingan_data_Ship!}{RGB}{255   0   0}
\definecolor{Pingan_data_Seawater!}{RGB}{ 0, 255,   0}
\definecolor{Pingan_data_Trees!}{RGB}{ 0,   0 ,255}
\definecolor{Pingan_data_Concretestructurebuilding!}{RGB}{255 ,255,   0}
\definecolor{Pingan_data_Floatingpier!}{RGB}{255 ,  0,255}
\definecolor{Pingan_data_Brickhouses!}{RGB}{ 0 255 255}
\definecolor{Pingan_data_Steelhouses!}{RGB}{200, 100 ,  0}
\definecolor{Pingan_data_Wharfconstructionland!}{RGB}{ 0, 200, 100}
\definecolor{Pingan_data_Car!}{RGB}{100 ,  0, 200}
\definecolor{Pingan_data_Road!}{RGB}{200 ,  0, 100}

\begin{document}
%


\title{How to Learn More? Exploring Kolmogorov-Arnold Networks for Hyperspectral Image Classification}


\author{
       Ali Jamali,
       Swalpa Kumar Roy,~\IEEEmembership{Senior Member,~IEEE,} 
       Danfeng Hong,~\IEEEmembership{Senior Member,~IEEE}, \\
       Bing Lu, and 
       Pedram Ghamisi,~\IEEEmembership{Senior Member,~IEEE,}
       
\thanks{(\textit{Corresponding author: Ali Jamali})}
\thanks{A. Jamali is with the Department of Geography, Simon Fraser University, 8888 University Dr, Burnaby, BC V5A 1S6, Canada. (e-mail: alij@sfu.ca).}
\thanks{S. K. Roy is with the Department of Computer Science and Engineering, Alipurduar Government Engineering and Management College, West Bengal 736206, India (e-mail: swalpa@agemc.ac.in).}
\thanks{D. Hong is with the Aerospace Information Research Institute, Chinese Academy of Sciences, 100094 Beijing, China, and also with the School of Electronic, Electrical and Communication Engineering, University of Chinese Academy of Sciences, 100049 Beijing, China. (e-mail: hongdf@aircas.ac.cn).}
\thanks{B. Lu is with the Department of Geography, Simon Fraser University, 8888 University Dr, Burnaby, BC V5A 1S6, Canada. (e-mail: alij@sfu.ca).}
\thanks{P. Ghamisi is with the Helmholtz-Zentrum Dresden-Rossendorf (HZDR), Helmholtz Institute Freiberg for Resource Technology, Machine Learning Group 09599 Freiberg, Germany, and also with Lancaster University, LA1 4YR Lancaster, U.K. (e-mail: p.ghamisi@gmail.com).}

}

\markboth{Submitted to IEEE}
 {Jamali \MakeLowercase{\textit{et al.}}: Bare Demo of IEEEtran.cls for Journals}

\maketitle

\begin{abstract}
Convolutional Neural Networks (CNNs) and vision transformers (ViTs) have shown excellent capability in complex hyperspectral image (HSI) classification. However, these models require a significant number of training data and are computational resources. On the other hand, modern Multi-Layer Perceptrons (MLPs) have demonstrated great classification capability. These modern MLP-based models require significantly less training data compared to CNNs and ViTs, achieving the state-of-the-art classification accuracy. Recently, Kolmogorov-Arnold Networks (KANs) were proposed as viable alternatives for MLPs. Because of their internal similarity to splines and their external similarity to MLPs, KANs are able to optimize learned features with remarkable accuracy in addition to being able to learn new features. Thus, in this study, we assess the effectiveness of KANs for complex HSI data classification. Moreover, to enhance the HSI classification accuracy obtained by the KANs, we develop and propose a Hybrid architecture utilizing 1D, 2D, and 3D KANs. To demonstrate the effectiveness of the proposed KAN architecture, we conducted extensive experiments on three newly created HSI benchmark datasets: QUH-Pingan, QUH-Tangdaowan, and QUH-Qingyun. The results underscored the competitive or better capability of the developed hybrid KAN-based model across these benchmark datasets over several other CNN- and ViT-based algorithms, including 1D-CNN, 2DCNN, 3D CNN, VGG-16, ResNet-50, EfficientNet, RNN, and ViT. The code are publicly available at \url{https://github.com/aj1365/HSIConvKAN}

\end{abstract}


\IEEEpeerreviewmaketitle

\section{Introduction}

\IEEEPARstart{H}{yperspectral} remote sensing has drawn a lot of interest lately for a variety of Earth observation uses \cite{hong2024spectralgpt,Ullah46,Li2024Learning}. Mapping the physical, biological, or geographical dimensions of ecosystems is necessary to monitor the temporal and spatial patterns of earth surface activities and comprehend how they work. Because each pixel contains a wealth of spectral information, hyperspectral imaging (HSI) has been applied extensively in a variety of real-world applications, including precision agriculture \cite{KHAN202678}, military object detection \cite{Ke27}, and land use land cover mapping \cite{hong2023cross, Roy184}. Because it offers precise and detailed information about the physical and chemical properties of objects that are imaged, HSI has grown to be an essential tool in the industry. Notably, the detailed features produce effective classification results that are too intricate for conventional methods, i.e., a nonlinear correlation among the obtained spectrum data and the corresponding object, such as buildings \cite{Ullah46}.

As opposed to standard panchromatic and multi-spectral imagery captured by satellites, HSI supplies hundreds of contiguous narrow spectral bands, offering an improved detailed and accurate technique for discerning Earth objects \cite{Landgrebe718}. HSI is especially useful for more refined classification because of its capacity to identify subtle spectral characteristics that standard imagery is unable to detect \cite{Li0085}. The majority of techniques used in the early stages of HSI classification research concentrated on handcrafted extraction of features, such as extended morphological profiles (EMPs) \cite{Fauvel3943}. However, these conventional classification techniques are limited in their ability to retrieve high-level characteristics of images, and they are associated with ``shallow'' models. As a result, these techniques typically fall short of achieving greater accuracy. Recently, it has been established that deep learning (DL) is a powerful feature extractor that effectively recognizes the nonlinear problems that have emerged in a variety of computer vision tasks. This promotes the encouraging outcomes of using DL for HSI data classification \cite{he2023foundation, He33, li2023lrr}.

Convolutional neural networks (CNNs), because of their superior local contextual modeling capabilities, are widely used in spectral-spatial HSI data classification. While CNN-based methods are advantageous for spatial-contextual identification, they suffer greatly from handling spectral sequential data because long-range dependencies are often difficult for CNNs to capture correctly \cite{Xue059}. While the existing CNN-based techniques have shown promising results \cite{7514991}, they continue to encounter several difficulties. For instance, the receptive field is constrained, data is lost during the downsampling phase, and deep networks require a large amount of processing power \cite{yu2024hypersinet}. On the other hand, in the field of computer vision, vision transformers (ViTs) have demonstrated significant promise recently \cite{Li7135, Lee78, ran2023deep, yao2023extended, Yang0197}. By means of the incorporation of a multi-layer perceptron (MLP) and a multi-headed self-attention (MHSA) module, ViTs are capable of acquiring global long-range data interactions in the input sequential data. Because of this capability, the application of transformers to the classification of HSI data is expanding rapidly \cite{Liang137, Hong27165, Mei5238,li2024casformer}. 

However, due to their quadratic computational complexity, transformers need a substantially higher amount of training data than CNNs and have a relatively high computational cost \cite{NEUcba0a4ee}. ViTs and CNNs have been surpassed in image classification tasks by modern MLP algorithms, such as MLP-Mixer \cite{NEUcba0a4ee} and ResMLP \cite{Touvron2023}, which have demonstrated excellent classification capability. These modern MLP models require significantly less training data compared to CNNs and ViTs, achieving state-of-the-art classification accuracy \cite{Ali888}. In addition, SpectralMamba \cite{yao2024spectralmamba} was proposed for hyperspectral image classification to further reduce computational complexity while effectively improving classification performance. This work is notable as the first to introduce the Mamba framework into the hyperspectral remote sensing field.

In the past few months, Kolmogorov-Arnold Networks (KANs), which are inspired by the Kolmogorov-Arnold representation theorem, were proposed as viable alternatives for MLPs \cite{liu2024kan}. KANs employ feature learnable activation functions on edges, or ``weights,'' in opposed to MLPs, which have fixed activation functions on nodes, or ``neurons''. KANs do not use any linear weights at all; instead, a uni-variate function with spline parameterization serves as a substitute for each weight parameter. Thus, in this research, we assess and evaluate the capability and effectiveness of KAN models for complex HSI data classification over several other CNN- and vision-based models. The contributions of this paper can be summarized as follows:

\begin{itemize}
   \item We introduce a hybrid architecture based on KANs, a technique that achieves competitive or better HSI classification accuracy over several well-known CNN- and ViT-based algorithms.
   \item We incorporate 1-D, 2-D, and 3-D KAN modules to enhance the ability of linear KANs in image classification tasks. This hybrid architecture increases the discriminative capability of the KAN architecture.
   \item We conduct extensive experiments on a brand-new, complex HSI dataset called Qingdao UAV-borne HSI (QUH), including QUH-Tangdaowan, QUH-Qingyun, and QUH-Pingan \cite{FU2023115}. These experiments prove the effectiveness of the proposed KAN architecture.
\end{itemize}

The remainder of the paper is structured as follows. In Section \ref{sec:method}, we examine the structure and various modules developed in the proposed KAN models-based architecture. Subsequently, we conduct comprehensive experiments, including a thorough discussion of the obtained HSI data classification results, as detailed in Section \ref{sec:results}. The paper concludes with a summary provided in Section \ref{sec:Concl}.

\section{Proposed Methodology}
\label{sec:method}

Multilayer perceptrons (MLPs) are the foundation of many modern deep learning models. KANs were recently presented as an alternative to MLPs \cite{liu2024kan}. KANs are motivated by the Kolmogorov-Arnold representation theorem \cite{kolmogorov1957representation}, whereas MLPs are inspired by the universal approximation theorem. Similar to MLPs, KANs have fully-connected structures. But MLPs employ fixed activation functions on nodes (referred to as ``neurons''), while KANs place learnable activation functions on edges (referred to as ``weights''). Instead of using linear weight matrices, KANs use a learnable 1D function parametrized as a spline for each weight parameter. Nodes in KANs do nothing more than add up incoming signals without using any non-linearities. The straightforward modification of KANs to use an activation function on the edges allows them to surpass MLPs in terms of accuracy as well as interpretability on small-scale machine learning challenges. In function-fitting tasks, smaller KANs can attain accuracy levels that are comparable to or higher than larger MLPs. KANs are known to have faster neural scaling laws than MLPs, both in theory and in practice \cite{liu2024kan}. Splines can be easily adjusted locally, are precise for low-dimensional functions, and can transition between different resolutions. However, due to their limited ability to take advantage of compositional structures, splines suffer greatly from the curse of dimensionality (COD). In contrast, MLPs are less prone to COD because of their feature learning capabilities. However, in low dimensions, their accuracy is inferior to splines due to their incapacity to optimize univariate functions. It should be noted that KANs are just combinations of splines and MLPs, utilizing their respective advantages and avoiding their respective disadvantages, despite their sophisticated mathematical interpretation. In order to correctly learn a function, the model must be able to approximate the univariate functions (internal degrees of freedom) as well as learn the compositional structure (external degrees of freedom). Because of their internal similarity to splines and their external similarity to MLPs, KANs are able to optimize learned features with remarkable accuracy in addition to being able to learn new features. 

\textbf{Are KANs similar to MLP?} An MLP can be expressed as stacking $N$ layers and each layer may be expressed as a linear combination of weight matrix ($W$) followed by non-linear operations ($\delta$) for input $X\in\mathcal{R}^{p_{in}}$:

\begin{equation}
MLP(x)= (W_{N-1} \circ \delta \circ W_{N-2} \circ \cdots  W_{1} \circ \delta \circ W_{0})~x
\end{equation}

\begin{figure}[!t]
    \centering
    \includegraphics[clip=true, trim = 05 480 05 05, width=0.95\textwidth]{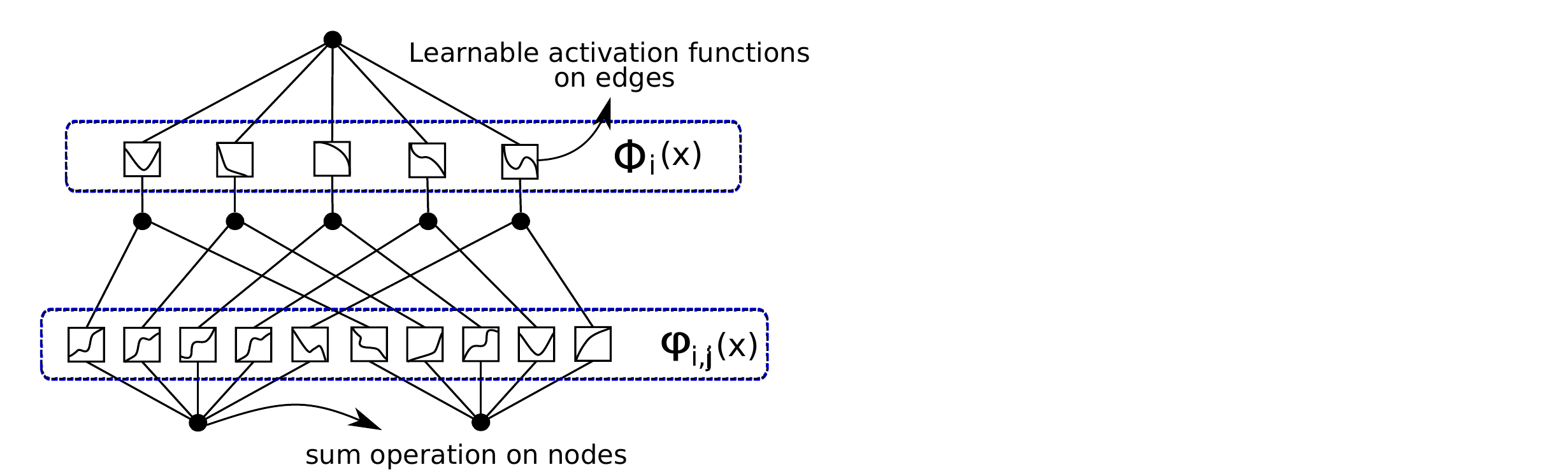} 
    \caption{The overall architecture of the Kolmogorov-Arnold Networks.}
    \label{fig:kat}
\end{figure}

\begin{figure*}[!ht]
\centering
\begin{tikzpicture}[mmat/.style={matrix of math nodes,column sep=-\pgflinewidth/2,
row sep=-\pgflinewidth/2,cells={nodes={draw,inner sep=7pt,thin}},draw=#1,thick,inner sep=.5pt},
mmat/.default=black,
node distance=0.5em]
 \matrix[mmat](mat1){
         0 & 1 & 1 & x_{11} & x_{12} & x_{13} & 0 \\ 
         0 & 0 & 1 & x_{21} & x_{22} & x_{23} & 0 \\ 
         0 & 0 & 0 & x_{31} & x_{32} & x_{33} & 0 \\ 
         0 & 0 & 0 & 1 & 1 & 0 & 0 \\ 
         0 & 0 & 1 & 1 & 0 & 0 & 0 \\ 
         0 & 1 & 1 & 0 & 0 & 0 & 0 \\ 
         0 & 1 & 0 & 0 & 0 & 0 & 0 \\ 
         };
 \def\myarray{{1,0,1},{0,1,0},{1,0,1}}       
 \foreach \X in {0,1,2}
 {\foreach \Y in {0,1,2}
  {\pgfmathsetmacro{\myentry}{{\myarray}[\Y][\X]}
  \path (mat1-\the\numexpr\Y+1\relax-\the\numexpr\X+4\relax.south east)
  node[anchor=south east,blue,scale=0.3,inner sep=2pt]{$\times\myentry$};
  }}         
 \node[fit=(mat1-1-4)(mat1-3-6),inner sep=0pt,draw,red,thick,name path=fit](f1){};      
 \node[right=of mat1] (mul) {$\circ$};      
 \matrix[mmat=blue,fill=blue!20,right=of mul,name path=mat2](mat2){    
     \phi_{11}(\cdot) & \phi_{12}(\cdot) & \phi_{13}(\cdot) \\ 
     \phi_{21}(\cdot) & \phi_{22}(\cdot) & \phi_{23}(\cdot) \\ 
     \phi_{31}(\cdot) & \phi_{32}(\cdot) & \phi_{33}(\cdot) \\ };
 \node[right=of mat2] (eq) {$=$};       
 \matrix[mmat,right=of eq](mat3){    
     o_{11} & o_{12} & o_{13} & |[draw=green,thick,fill=green!20,alias=4]|o_{14} & o_{15} \\ 
     o_{21} & o_{22} & o_{23} & o_{24} & o_{25} \\ 
     o_{31} & o_{32} & o_{33} & o_{34} & o_{35} \\ 
     o_{41} & o_{42} & o_{43} & o_{44} & o_{45} \\ 
     o_{51} & o_{52} & o_{53} & o_{54} & o_{55} \\
 };
 \foreach \Anchor in {south west,north west,south east,north east}
 {\path[name path=test] (f1.\Anchor) -- (mat2.\Anchor);
 \draw[blue,densely dotted,name intersections={of=test and fit,total=\t}]
 \ifnum\t>0 (intersection-\t) -- (mat2.\Anchor) \else
  (f1.\Anchor) -- (mat2.\Anchor)\fi;
 \path[name path=test2]  (4.\Anchor) -- (mat2.\Anchor);  
 \draw[green,densely dotted,name intersections={of=test2 and mat2,total=\tt}] 
 \ifnum\tt>0 (intersection-1) -- (4.\Anchor) \else
    (mat2.\Anchor) --  (4.\Anchor)\fi;
    }
 \path (mat1.south) node[below] {$\mathbf{X}$}
  (mat2|-mat1.south) node[below] {$\mathbf{\Phi_{i}}$}
  (mat3|-mat1.south) node[below] {$\mathbf{X}\circ\mathbf{\Phi}$};
 \begin{scope}[on background layer]
  \fill[red!20] (f1.north west) rectangle (f1.south east);
 \end{scope}
\end{tikzpicture}
\caption{Pictorial representation of KAN Convolution operation where $x$, and $\Phi$ represent the input sub-patch and B-splines, respectively. The output of $o_{14} = x \circ \Phi$ can be calculated as $\phi_{11}(x_{11})+\phi_{12}(x_{12})+\phi_{13}(x_{13})+\phi_{21}(x_{21})+\phi_{22}(x_{22})+\phi_{23}(x_{23})+\phi_{31}(x_{31})+\phi_{32}(x_{32})+\phi_{33}(x_{33})$.}
\label{fig:kan_cov}
\end{figure*}
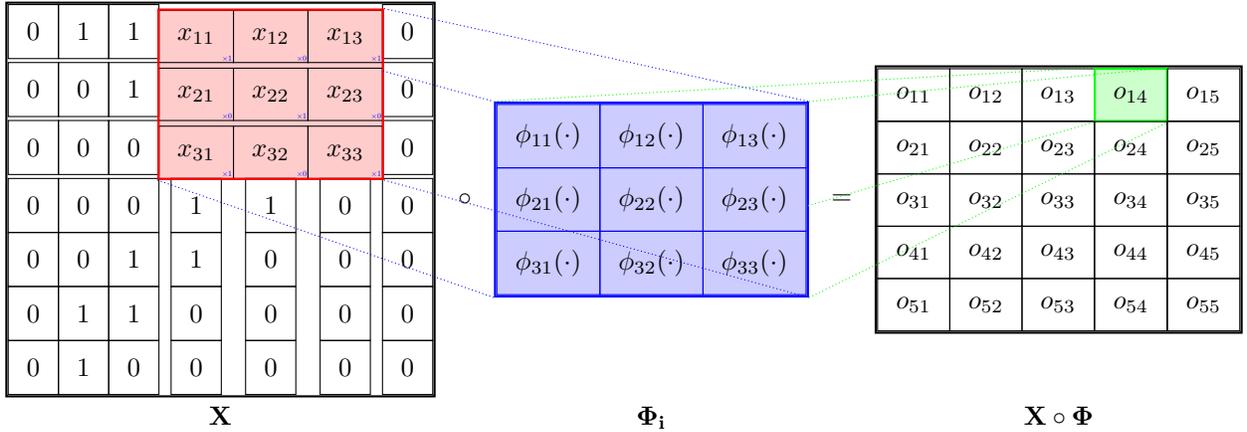

On the other hand, a general KAN model consists of nesting $N$ layers and the output map can be defined as:

\begin{equation}
KAN(x)= (\Phi_{N-1} \circ \Phi_{N-2} \circ \cdots  \Phi_{1} \circ \Phi_{0})~x
\end{equation}

\noindent where $\Phi_i$ represents the $i$-th layer of whole KAN models. Let the $p_{in}$ and $p_{out}$ be the dimension of input and output for each KAN layer, then $\Phi$ consists of $p_{in}\times p_{out}$ 1-D learnable activation function $\phi$:

\begin{equation}
    \Phi = \{\phi_{i,j} \} ~~~~~ i = 1, 2, \ldots p_{in}, ~~~~ j = 1, 2, \ldots p_{out}
\end{equation}

The  outcome of KAN models while computing from layer $n$ to layer $n$ + 1 may be shown in matrix form as follows:

\begin{equation}
   X_{n+1} = \underbrace{\left[\begin{matrix}
\phi_{n,1,1}(\cdot) & \phi_{n,1,2}(\cdot) & \ldots & \phi_{n,1,p_{n}}(\cdot) \\
\phi_{n,2,1}(\cdot) & \phi_{n,2,2}(\cdot) & \ldots & \phi_{n,2,p_{n}}(\cdot) \\
\vdots &\vdots & \vdots & \vdots \\
\phi_{n,p_{n+1},1}(\cdot) & \phi_{n,p_{n+1},2}(\cdot) & \ldots & \phi_{n,p_{n+1},p_{n}}(\cdot) 
\end{matrix}\right]}_{\Phi_n}~X_n
\end{equation}

It is evident that KANs treat non-linearities and linear transformations collectively in $\Phi$, whereas MLPs treat them separately as $W$ and $\delta$. To ensure the representation power of $\phi_{i,j}$ and $\Phi_i$ as shown in Fig. \ref{fig:kat}, in the KAN models a basis function $b(x)$ (similar to that of residual connections) is included such that the activation function $\phi(x)$ is the sum of the many spline function and the basis function $b(x)$, as defined by:

\begin{equation}
\phi(x)= w_b ~b(x) + w_s ~spline(x)
\end{equation}



\noindent where $b(x)= silu(x)= x/ (1 + e^{-x})$, spline(x)= $\sum_{i}\rm{c_i ~ B_i (x)}$, and $c_i$s are trainable. For more details refer to Liu et al.\cite{liu2024kan}.

\begin{figure*}[!ht]
    \centering
    \includegraphics[clip=true, trim = 05 600 05 05, width=0.99\textwidth]{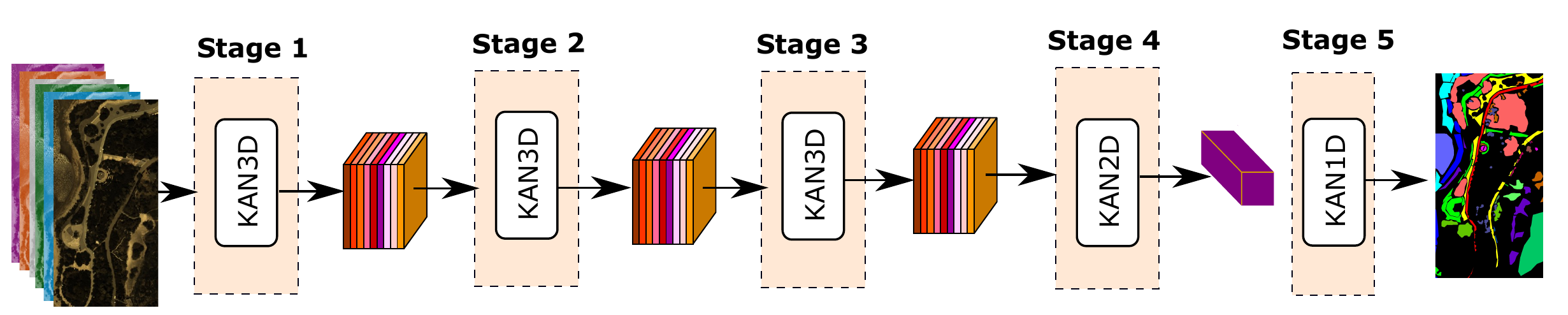} 
    \caption{The overall architecture of the proposed Hybrid KAN.}
    \label{fig:KAN}
\end{figure*}

\textbf{Classical vs. KAN Convolution:} KAN convolutions are perhaps similar to traditional convolutions operation, except that each element is given to a learnable non-linear activation function, which is then added to the kernel and the associated pixels in the image patch, rather than the dot product between the two. The kernel of the KAN convolution is equivalent to a KAN linear Layer of 9 inputs and 1 output neuron (shown in Fig.~\ref{fig:kan_cov}). The output pixel of that convolution step is the sum of $\phi_i(x_i)$ for each input $i$ to which we have applied a $\phi_i$ learnable function. To visualize the difference between classical vs KAN convolution consider the input image patch $X\in\mathbf{R}^{W\times{H}}$, the output $O\in\mathbf{R}^{H'\times{W'}}$, the kernel $K$, and $\Phi$ for convolutional
\begin{equation}
X = \left[\begin{matrix}
x_{11} & x_{12} & x_{13} & \ldots & x_{1w}\\
x_{21} & x_{22} & x_{23} & \ldots & x_{2w}\\
x_{31} & x_{32} & x_{33} & \ldots & x_{3w}\\
\vdots & \vdots & \vdots & \vdots & \vdots\\
x_{h1} & x_{h2} & x_{h3} & \ldots & x_{hw}\\
\end{matrix} \right]_{H\times{W}} 
\label{eqn:patch}
\end{equation}
and KAN kernel are defined in Eqn.~(\ref{eqn:kernel}), respectively. 
\begin{equation}
K = 
\left[\begin{matrix}
k_{11} & k_{12} & k_{13}\\
k_{21} & k_{22} & k_{23}\\
k_{31} & k_{32} & k_{33}\\
\end{matrix} \right] 
~and~~~ \Phi =
\left[\begin{matrix}
\phi_{11}(\cdot) & \phi_{12}(\cdot) & \phi_{13}(\cdot)\\
\phi_{21}(\cdot) & \phi_{22}(\cdot) & \phi_{23}(\cdot)\\
\phi_{31}(\cdot) & \phi_{32}(\cdot) & \phi_{33}(\cdot)\\
\end{matrix} \right] 
\label{eqn:kernel}
\end{equation} 
The output of the classical convolutional operation \ADD{($*$)} can be obtained as follows:
\begin{equation}
o_{i,j} = \sum_{m,n=0}^{K-1} x_{i+m,j+n}K_{m,n}
\end{equation}

In case of KAN convolution, the inner function $\phi(\cdot)$ may be represented as a matrix containing several activation functions as shown in Eqn~(\ref{eqn:kernel}). We also have an input matrix ($X$) that will cycle through each activation function and has $n\times{n}$ characteristics. It should be noted that $\phi(\cdot)$ here denotes the activation function rather than the weights. These activation functions are called B-splines. Let add all functions which are just basic polynomial curves and these curves are dependent upon the $X$ input. The output of the KAN convolutional operation ($\circ$) can be obtained as follows:

\begin{equation}
o_{i,j} = \sum_{m,n} \phi_{m,n}(x_{i+m,j+n})
\label{eqn:kan_conv}
\end{equation}

Similarly, the above Eqn.~(\ref{eqn:kan_conv}) can easily be extended for input image $X\in\mathbf{R}^{H\times{W}\times{C_{in}}}$ with $C_{in}$ channels by applying a set of KAN kernel $\Phi$ and produces output $O\in\mathbf{R}^{H'\times{W'}\times{C_{out}}}$ as follows:

\begin{equation}
o_{i,j,c} = \sum_{m,n,c} \phi_{m,n,c}(x_{i+m,j+n})
\end{equation}
 
\textbf{HybridSN an Embedding by KAN Layer}: We experimentally selected a KAN architecture similar to Hybrid spectral network \cite{Roy6016}, as seen in Fig. \ref{fig:KAN}. Hybrid spectral network was proposed in 2020 and consider to be a successful architecture in hyperspectral feature extraction and classification. Considering an input hyperspectral image of $X_{in}\in\mathbf{R}^{H\times{W}\times{B}}$, where $H$, $W$ and $B$ indicate the height, width and number of spectral bands, respectively. We first utilized a principle component analysis (PCA) algorithm to reduce the number of input channels/bands in all HSI datasets to $D$, as expressed as: 

\begin{equation}
X^{'} = {f}^{PCA} (X_{in})
\end{equation}

To enhance the HSI classification accuracy obtained by the KAN models, we develop and propose a Hybrid KAN network based architecture consisting of three consecutive 3D KANs with $8$, $16$, and $32$ number of output channels (feature maps), as expressed as:

\begin{equation}
X^{''} = KAN_{3D} (KAN_{3D}(KAN_{3D}(X)))
\end{equation}

Then one 2D KAN layer with an output channel (output map) of $64$ is employed immediately after the third 3D KAN. The resulting feature maps are then flattened and sent to a 1D KAN layer with a hidden layer of $32$ and output map/channel equivalent to the number classes in the HSI data, as expressed as:

\begin{equation}
class = KAN_{1D} (KAN_{2D}(X))
\end{equation}

The architecture of the proposed KAN-based model layer-wised is presented in Table \ref{tab:architecture}.

\begin{table}[!ht]
\centering
\caption{The layer-wise summary of the proposed HybridKAN architecture with window size 9×9. The last layer is based on the Tangdaowan dataset.)}
\resizebox{0.99\linewidth}{!}{
\begin{tabular}{|c|c|c|c|c|c|c|} \hline
Layer (type)   &  kernel size & Stride & Number of Kernels/Filters& Output Shape \\ 
\hline
\hline
KAN3D-1  &  1 & 1& 8 & (8, 9, 9, 1)   \\
KAN3D-2  &   1 & 1& 16 & (16, 9, 9, 1) \\
KAN3D-3  &   1 & 1& 32 & (32, 9, 9, 1) \\
Reshape &  - & -& -&(32, 9, 9)           \\
KAN2D-1  &   3 & 2& 64 &(64, 5 , 5) \\
Max pooling &   3 & 3 & - &(64, 1 , 1)         \\
Flatten &  - & -& -& (64, 1)           \\
KAN1D-1  &  - &-& 32& (64, 32, 18) \\
\hline
\end{tabular}}
\label{tab:architecture}
\end{table}


\section{Datasets}

The HSI data benchmarks that are being used are located in Qingdao City, Shandong Province, China's West Coast New Area. This city is close to China's Yellow Sea coast and features a wealth of both natural and artificial surroundings, as well as rapid urbanization. Because the morphology and distribution of each region's land cover are so complex, it is not easy to classify them precisely. An unmanned aerial vehicle (UAV) equipped with hyperspectral sensors was used to collect these datasets. More specifically, the UAV platform was the DJI M600 Pro. A hyperspectral sensor called the Gaiasky mini2-VN imaging spectrometer was used. Image mosaicking, radiometric calibration, and atmospheric and geometric corrections were all carried out using the instrument manufacturer's SpecView software \cite{FU2023115}.

\subsection{QUH-Tangdaowan}

\begin{figure}[!t]
\centering
	\begin{subfigure}{0.15\textwidth}
		\includegraphics[width=0.99\textwidth]{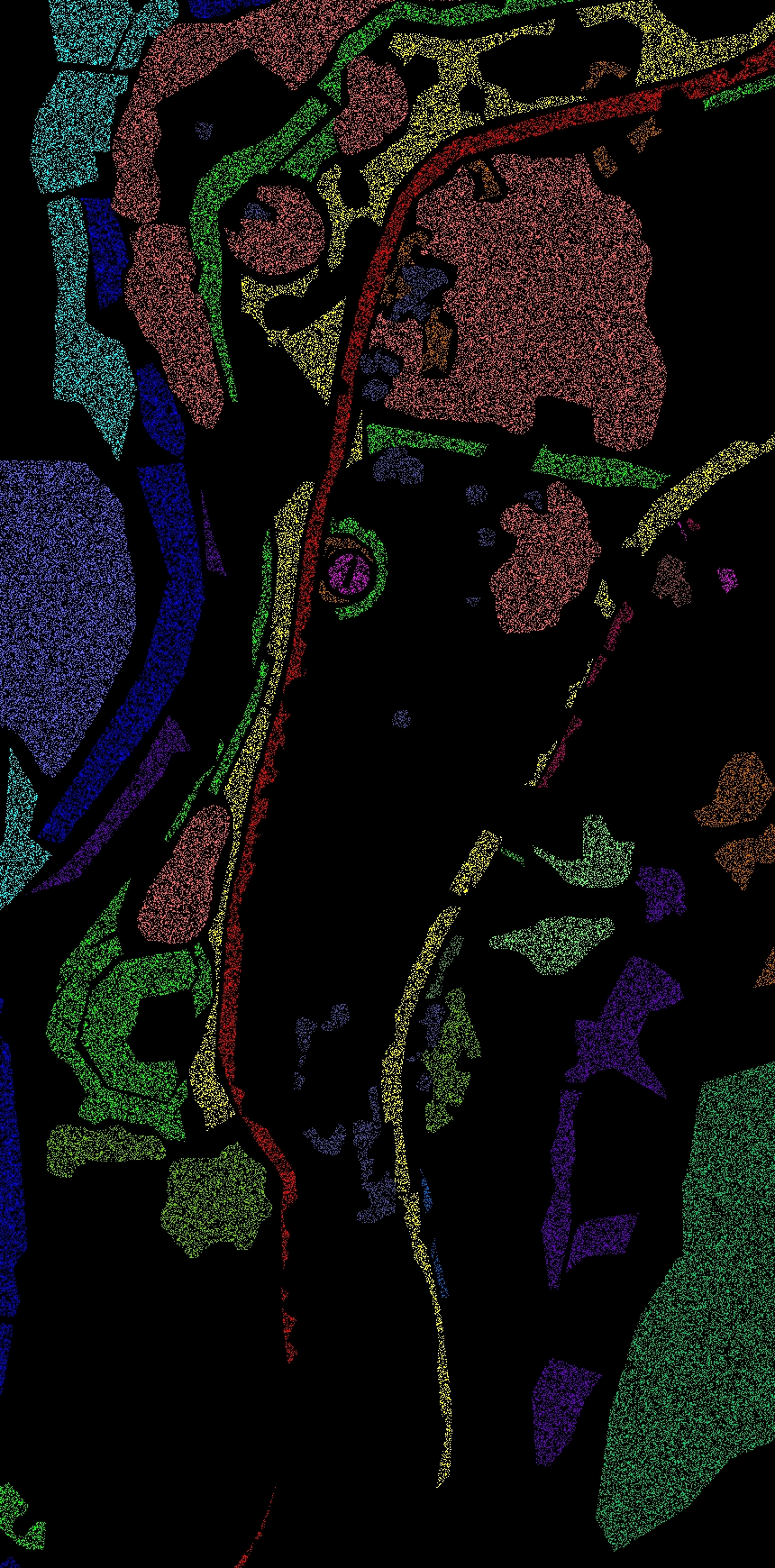}
		\caption{Train data}
	\end{subfigure}
         \begin{subfigure}{0.15\textwidth}
		\includegraphics[width=0.99\textwidth]{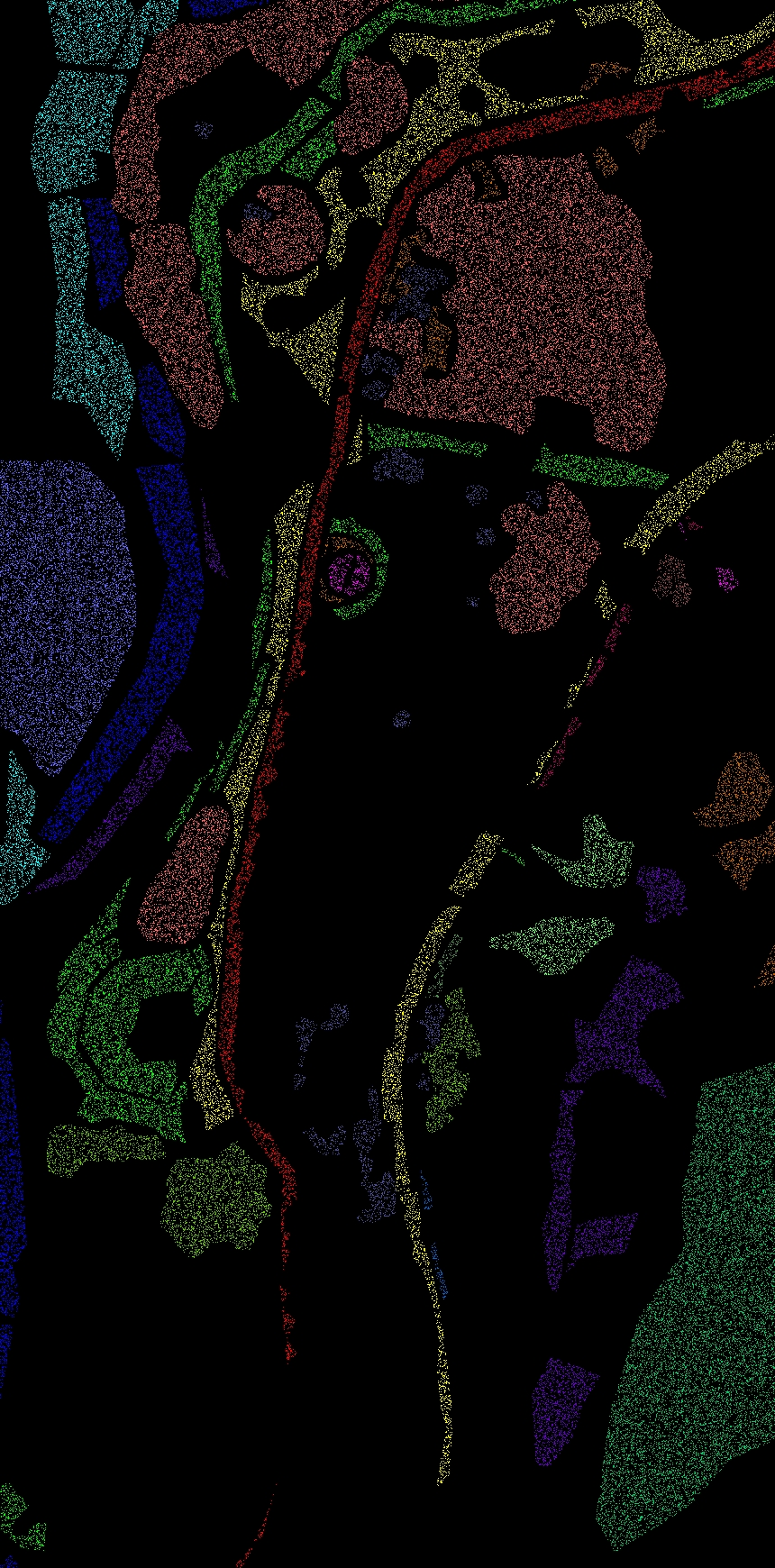}
		\caption{Validation data}
	\end{subfigure}
	\begin{subfigure}{0.15\textwidth}
		\includegraphics[width=0.99\textwidth]{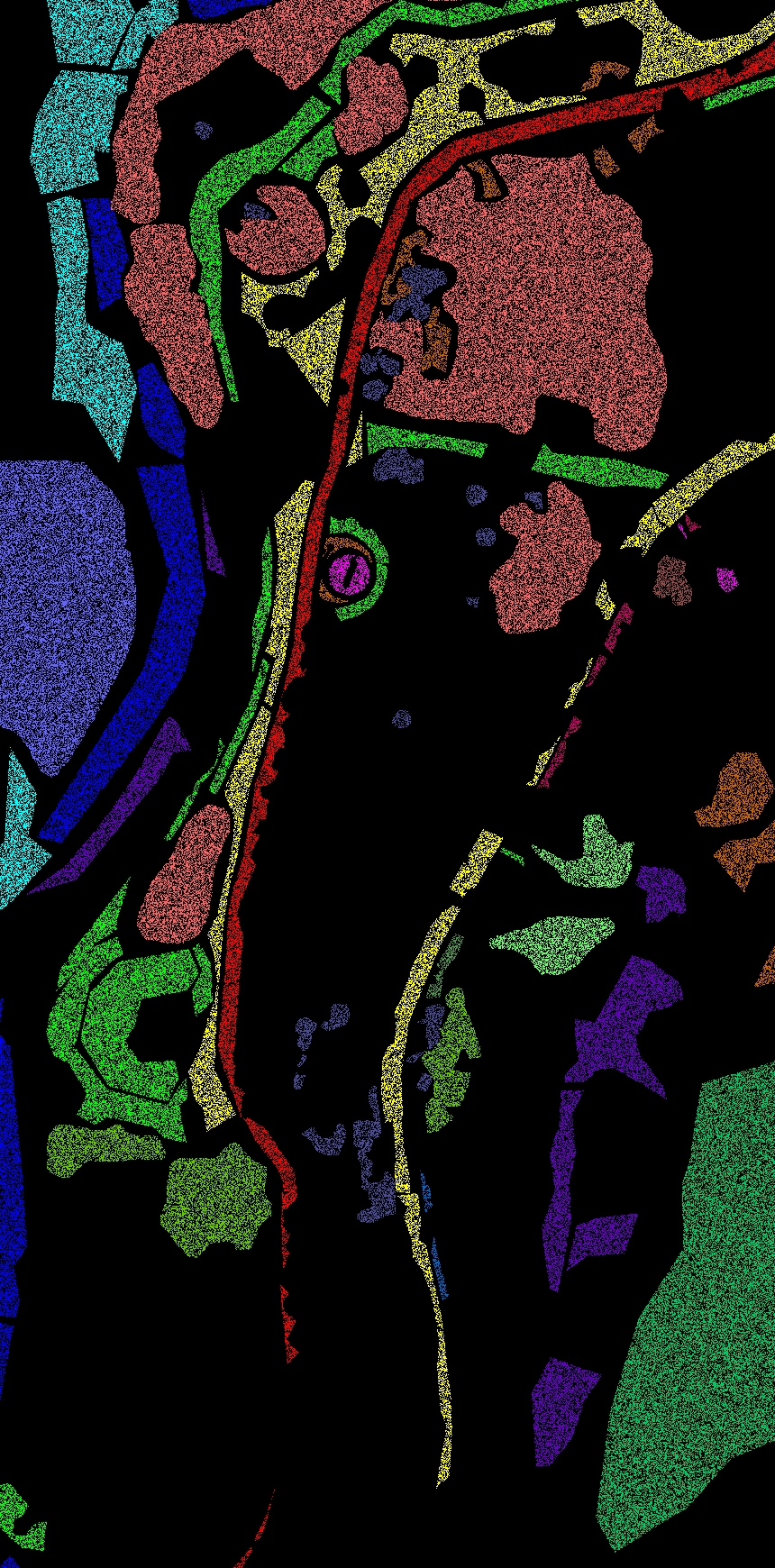}
		\caption{Test data}
	\end{subfigure}
\caption{Pictorial view of the QUH-Tangdaowan data benchmark: (a) the annotation of the training samples, (b) the annotation of the validation samples, and (c) the test samples.}
\label{fig:TangdaowanData}
\end{figure}

The QUH-Tangdaowan dataset was surveyed on May 18, 2021, in Tangdao Bay National Wetland Park, Qingdao, China. The UAV operated at a height of 300 meters with a spatial resolution of approximately 0.15 meters. This dataset comprises 176 bands with a wavelength range of 400–1000 nm and an image pixel size of 1740 × 860. Table~\ref{tab:TangdaowanData} and Fig\ref{fig:TangdaowanData} illustrate the number of training, validation, and test data in this data set.

\begin{table}[!ht]
\centering
\caption{Number of training, validation, and test ground truth data in QUH-Tangdaowan dataset.)}
\resizebox{0.99\linewidth}{!}{
\begin{tabular}{|c|c|c|c|c|c|c|} \hline
Class No. &  Color &Class & Train & Validation & Test & Total\\ 
\hline
\hline
1	&  \fcolorbox{Tangdaowan_data_Rubbertrack!}{Tangdaowan_data_Rubbertrack!}{\null} & Rubber track	&  7755	&   5170	&  12924	&  25849  
\\
2	&   \fcolorbox{Tangdaowan_data_Flaggingv!}{Tangdaowan_data_Flaggingv!}{\null} & Flaggingv	&  16666	&  11111 	&      27776 	&      55553   
\\
3		&    \fcolorbox{Tangdaowan_data_Sandy!}{Tangdaowan_data_Sandy!}{\null} & Sandy	&              10211	&        6807  	&     17019    	&   34037 
\\
4		&   \fcolorbox{Tangdaowan_data_Asphalt!}{Tangdaowan_data_Asphalt!}{\null} & Asphalt  	&           18207 	&      12138 	&      30345  	&     60690 
\\
5		&    \fcolorbox{Tangdaowan_data_Boardwalk!}{Tangdaowan_data_Boardwalk!}{\null} & Boardwalk	&             559  	&       372 	&        931	&         1862 
\\
6		&    \fcolorbox{Tangdaowan_data_Rockyshallows!}{Tangdaowan_data_Rockyshallows!}{\null} & Rocky shallows  	&        11137 	&       7425 	&      18563	&       37125 
\\
7		&   \fcolorbox{Tangdaowan_data_Grassland!}{Tangdaowan_data_Grassland!}{\null} & Grassland  	&           4238  	&      2825 	&       7064	&       14127  
\\
8	 	&   \fcolorbox{Tangdaowan_data_Bulrush!}{Tangdaowan_data_Bulrush!}{\null} & Bulrush	&             19226  	&     12817	&       32044	&       64087 
\\
9		&  \fcolorbox{Tangdaowan_data_Gravelroad!}{Tangdaowan_data_Gravelroad!}{\null} & Gravel road  	&          9208  	&      6139    	&   15348   	&    30695  
\\
10	  &  \fcolorbox{Tangdaowan_data_Ligustrumvicaryi!}{Tangdaowan_data_Ligustrumvicaryi!}{\null} & Ligustrum vicaryi 	&        535  	&       357  	&       891	&         1783  
\\
11	  &  \fcolorbox{Tangdaowan_data_Coniferouspine!}{Tangdaowan_data_Coniferouspine!}{\null} & Coniferous pine	&          6371	&        4247	&       10618  	&     21236   
\\
12	  &  \fcolorbox{Tangdaowan_data_Spiraea!}{Tangdaowan_data_Spiraea!}{\null} & Spiraea     	&         225 	&        150 	&        374    	&     749   
\\
13	  & \fcolorbox{Tangdaowan_data_Baresoil!}{Tangdaowan_data_Baresoil!}{\null} & Bare soil    	&         506    	&     337 	&        843 	&        1686  
\\
14	  & \fcolorbox{Tangdaowan_data_Buxussinica!}{Tangdaowan_data_Buxussinica!}{\null} &  Buxus sinica    	&        266  	&       177 	&        443	&         886    
\\
15	  &  \fcolorbox{Tangdaowan_data_Photiniaserrulata!}{Tangdaowan_data_Photiniaserrulata!}{\null} & Photinia serrulata  	&       4206	&        2804 	&       7010  	&     14020   
\\
16	  &   \fcolorbox{Tangdaowan_data_Populus!}{Tangdaowan_data_Populus!}{\null} & Populus 	&            42271  	&     28181 	&      70452	&       140904  
\\
17	  &  \fcolorbox{Tangdaowan_data_UlmuspumilaL!}{Tangdaowan_data_UlmuspumilaL!}{\null} & Ulmus pumila L   	&        2940 	&       1961 	&       4901   	&     9802   
\\
18	  &  \fcolorbox{Tangdaowan_data_Seawater!}{Tangdaowan_data_Seawater!}{\null} & Seawater  	&           12682 	&       8455 	&      21138 	&      42275   

\\
\hline
- 	  & -&Total    	&       167209  	&    111473  	&    278684  	&    557366  
\\
\hline
\end{tabular}}
\label{tab:TangdaowanData}
\end{table}

\subsection{QUH-Qingyun}

\begin{figure}[!t]
\centering
	\begin{subfigure}{0.15\textwidth}
		\includegraphics[width=0.99\textwidth]{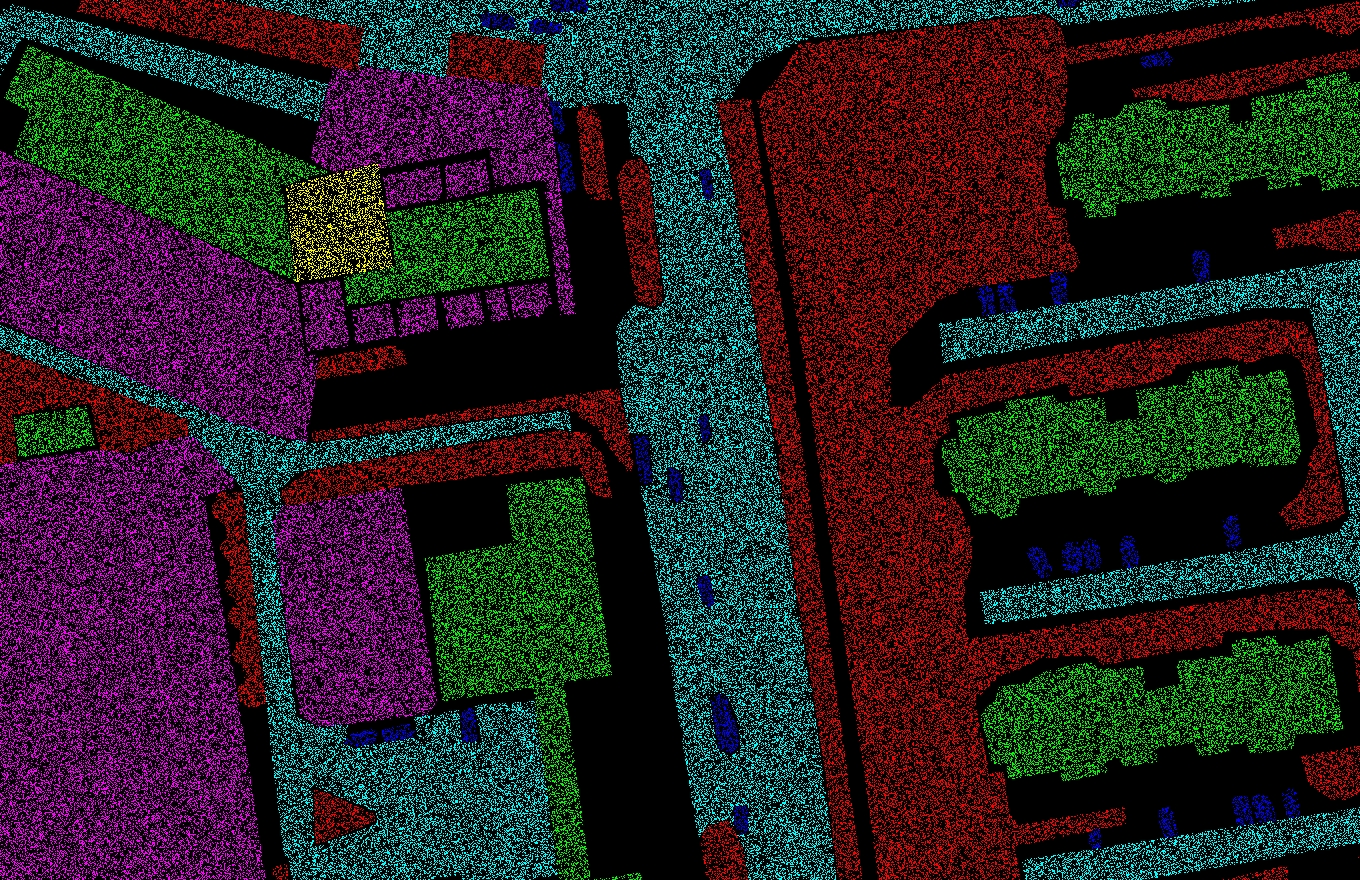}
		\caption{Train data}
	\end{subfigure}
         \begin{subfigure}{0.15\textwidth}
		\includegraphics[width=0.99\textwidth]{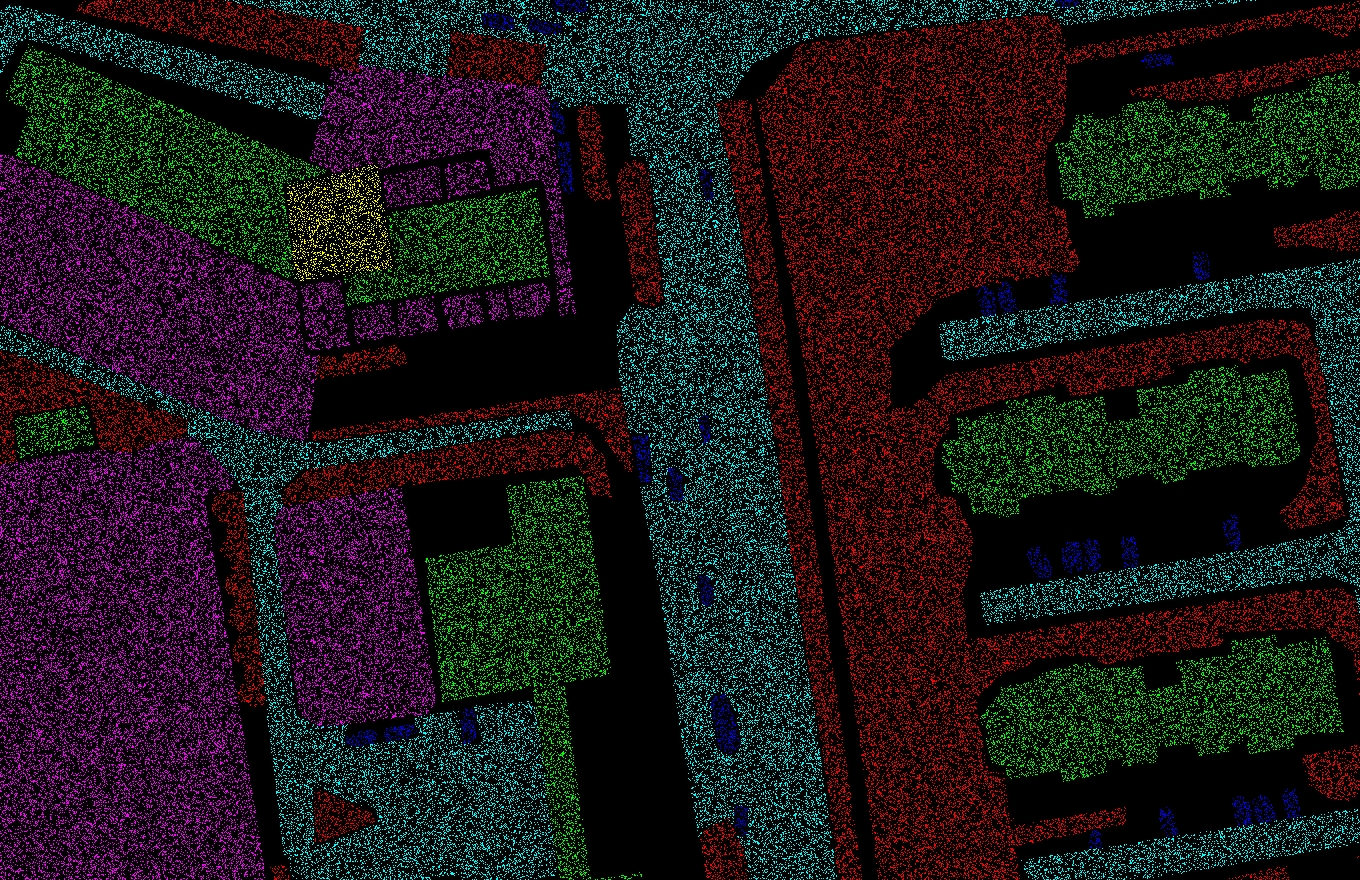}
		\caption{Validation data}
	\end{subfigure}
	\begin{subfigure}{0.15\textwidth}
		\includegraphics[width=0.99\textwidth]{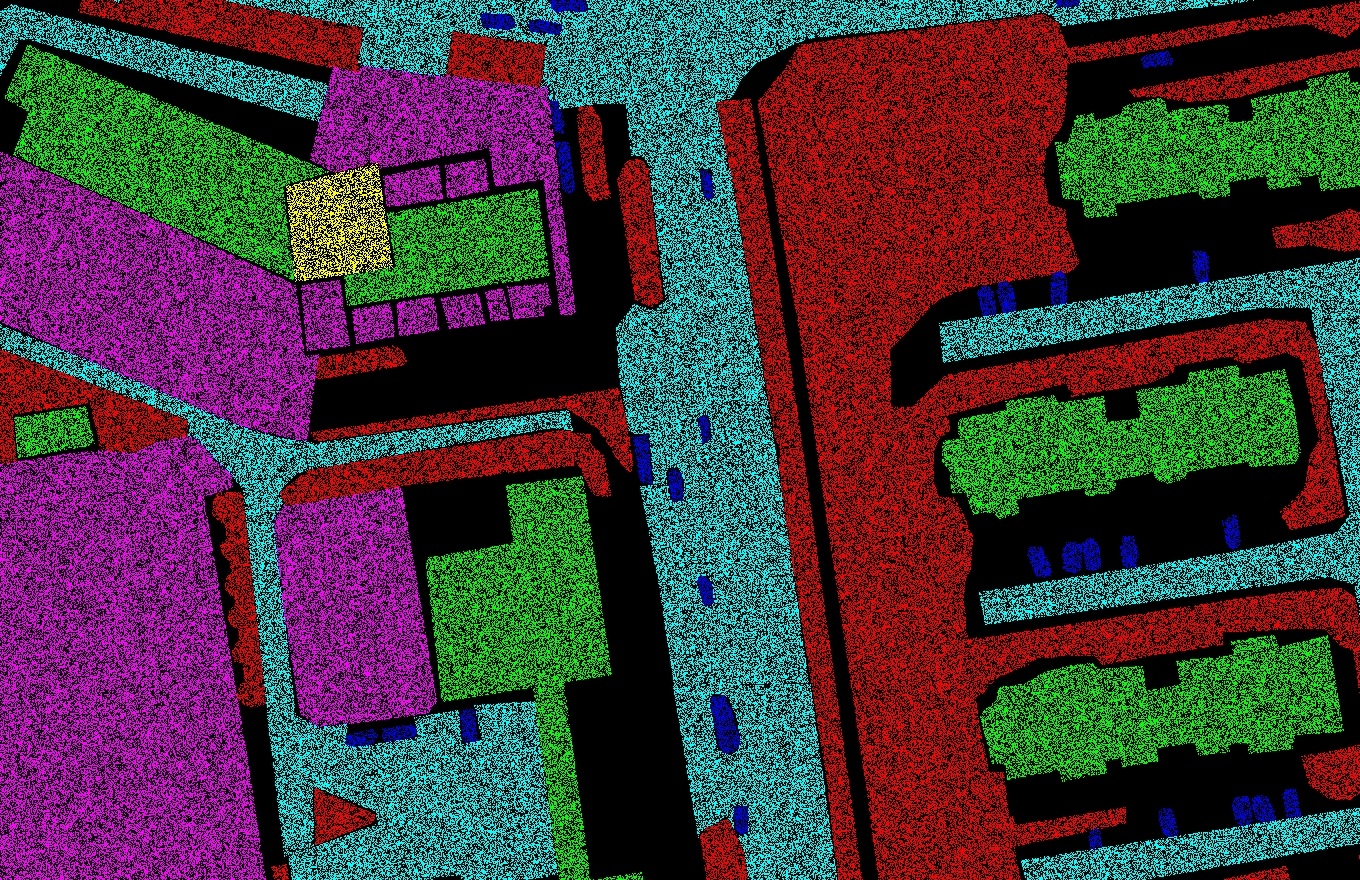}
		\caption{Test data}
	\end{subfigure}
\caption{Pictorial view of the QUH-Qingyun data benchmark: (a) the annotation of the training samples, (b) the annotation of the validation samples, and (c) the test samples.}
\label{fig:QingyunData}
\end{figure}

The QUH-Qingyun dataset was surveyed on May 18, 2021, in the vicinity of the Qingyun Road primary school and residential area in Qingdao, China. The UAV captured images with an image pixel size of 880 x 1360, 270 bands ranging from 400 to 1000 nm at a height of 300 meters with a spatial resolution of approximately 0.15 meters. Table~\ref{tab:QingyunData} and Fig\ref{fig:QingyunData} illustrate the number of training, validation, and test data in this data set.

\begin{table}[!ht]
\centering
\caption{Number of training, validation, and test ground truth data in QUH-Qingyun dataset.)}
\resizebox{0.99\linewidth}{!}{
\begin{tabular}{|c|c|c|c|c|c|c|} \hline
Class No. & Color &Class & Train & Validation & Test & Total\\ 
\hline
\hline
1	&   \fcolorbox{Qingyun_data_Trees!}{Qingyun_data_Trees!}{\null} & Trees  	&             83445   	&     55630  	&      139075   	&    278150  
\\
2	&    \fcolorbox{Qingyun_data_Concretebuilding!}{Qingyun_data_Concretebuilding!}{\null} & Concrete building   	&      53853   	&     35902 	&       89757	&        179512
\\
3		&  \fcolorbox{Qingyun_data_Car!}{Qingyun_data_Car!}{\null} &  Car   	&              4135   	&      2757 	&        6891  	&      13783   
\\
4		&  \fcolorbox{Qingyun_data_Ironhidebuilding!}{Qingyun_data_Ironhidebuilding!}{\null} & Ironhide building  	&        2930  	&       1953   	&      4884   	&      9767   
\\
5		&  \fcolorbox{Qingyun_data_Plasticplayground!}{Qingyun_data_Plasticplayground!}{\null} &  Plastic playground  	&       65320 	&       43547 	&       108868  	&     217735  
\\
6		& \fcolorbox{Qingyun_data_Asphaltroad!}{Qingyun_data_Asphaltroad!}{\null} &      Asphalt road  	&          76784  	&      51189	&        127973  	&     255946  

\\
\hline
- 	  & - &Total    	&       286467  	&     190978 	&      477448 	&      954893   
\\
\hline
\end{tabular}}
\label{tab:QingyunData}
\end{table}

\subsection{QUH-Pingan}

\begin{figure}[!t]
\centering
	\begin{subfigure}{0.15\textwidth}
		\includegraphics[width=0.99\textwidth]{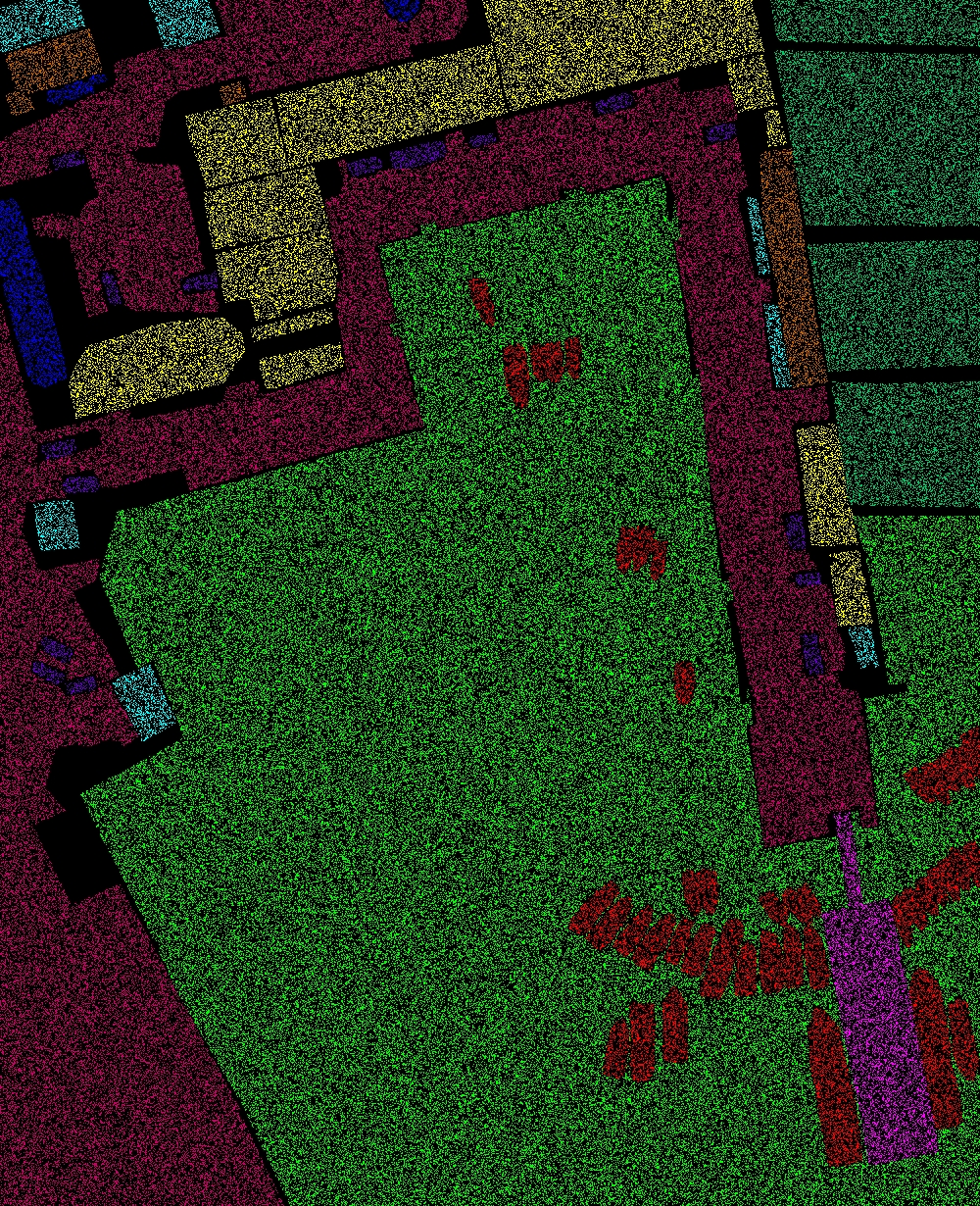}
		\caption{Train data}
	\end{subfigure}
         \begin{subfigure}{0.15\textwidth}
		\includegraphics[width=0.99\textwidth]{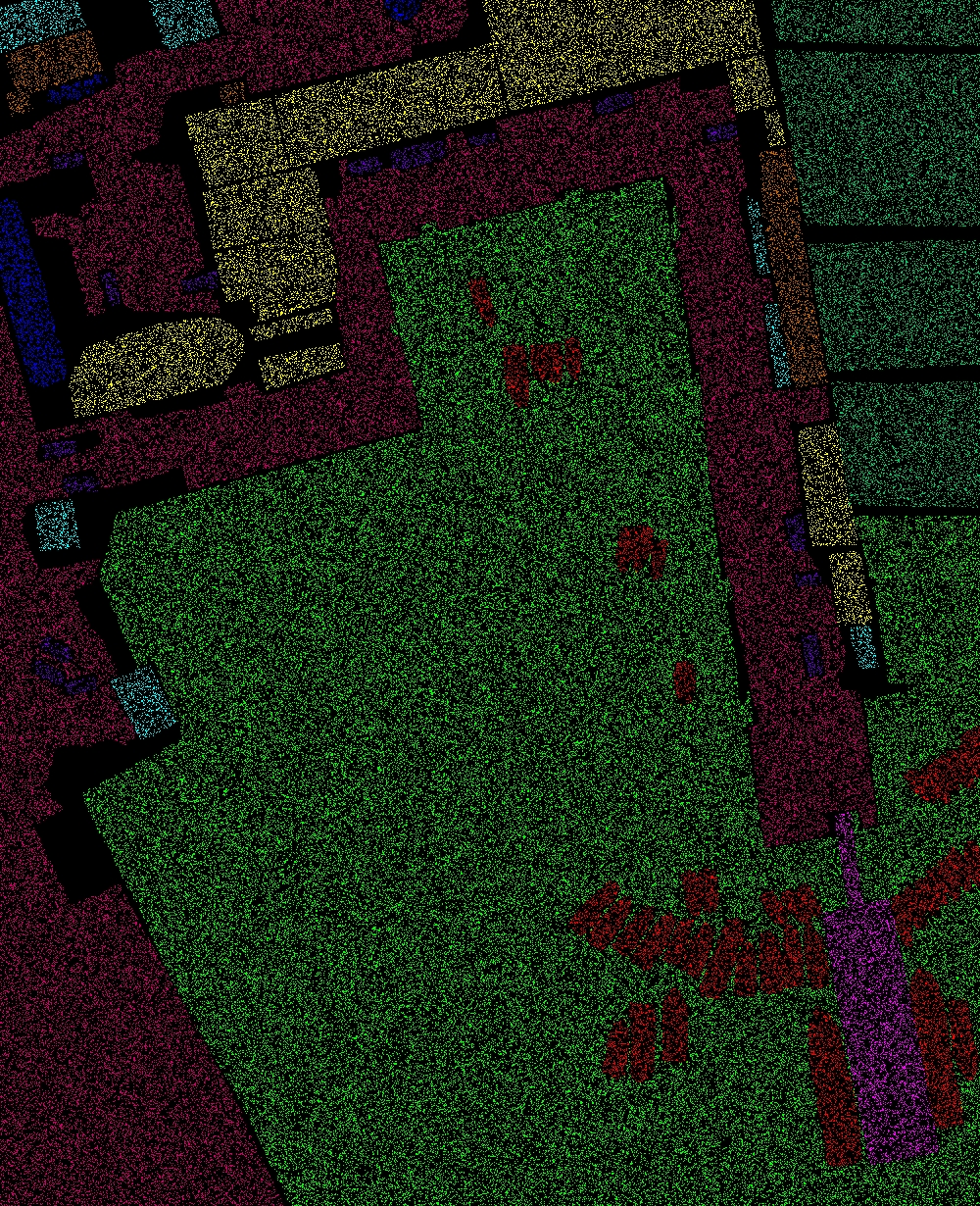}
		\caption{Validation data}
	\end{subfigure}
	\begin{subfigure}{0.15\textwidth}
		\includegraphics[width=0.99\textwidth]{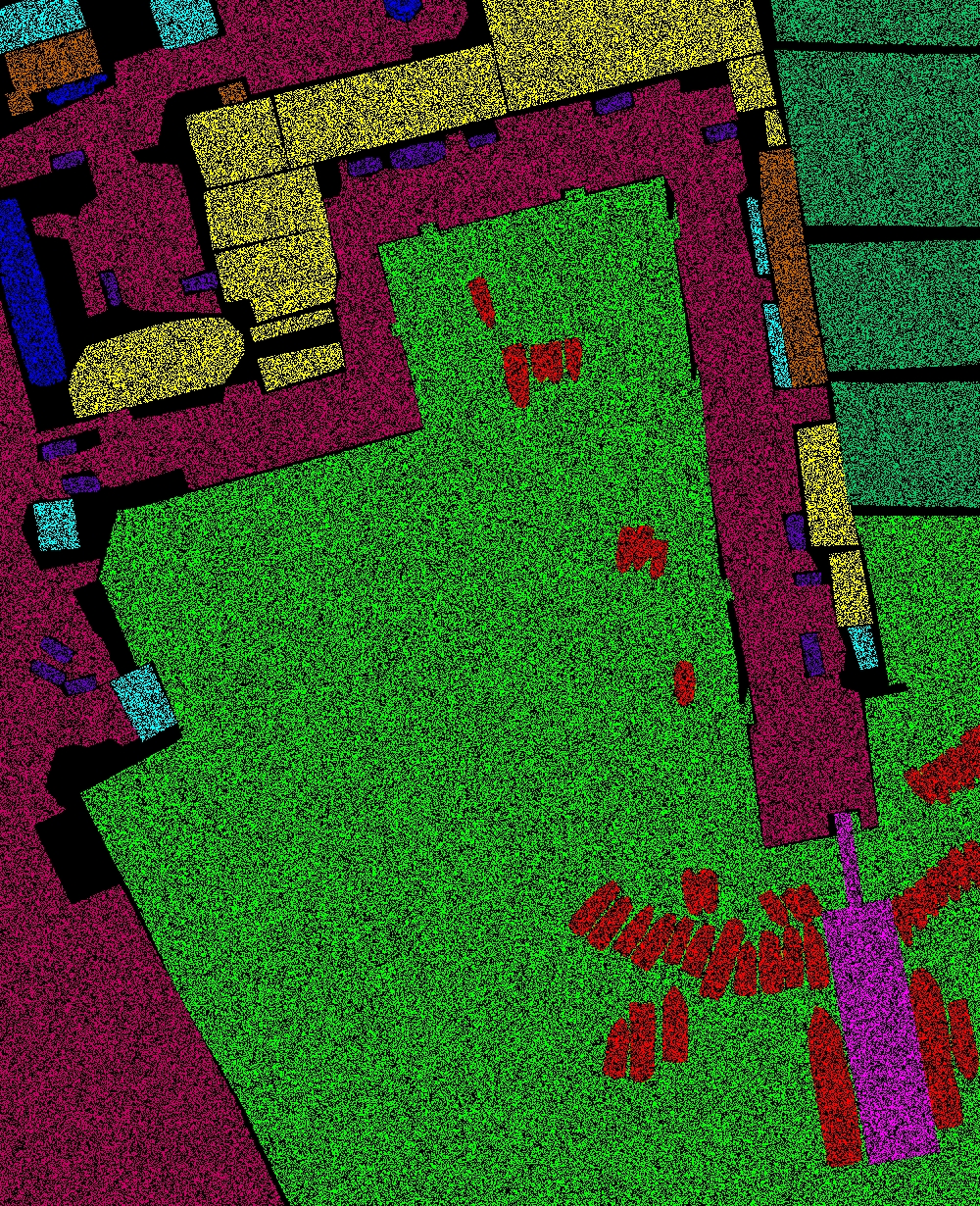}
		\caption{Test data}
	\end{subfigure}
\caption{Pictorial view of the QUH-Pingan data benchmark: (a) the annotation of the training samples, (b) the annotation of the validation samples, and (c) the test samples.}
\label{fig:PinganData}
\end{figure}

On May 19, 2021, at Huangdao Pingan Passenger Ship Terminal in Qingdao, China, the QUH-Pingan dataset was collected. The UAV operated at a height of 200 meters above the ground with a spatial resolution of approximately 0.10 meters. This dataset comprises 176 bands with a wavelength range of 400–1000 nm and an image pixel size of 1230 × 1000. Table~\ref{tab:PinganData} and Fig\ref{fig:PinganData} present the number of training, validation, and test data in this HSI data set.

\begin{table}[!ht]
\centering
\caption{Number of training, validation, and test ground truth data in QUH-Pingan dataset.)}
\resizebox{0.99\linewidth}{!}{
\begin{tabular}{|c|c|c|c|c|c|c|} \hline
Class No. & Color &Class & Train & Validation & Test & Total\\ 
\hline
\hline
1 	&     \fcolorbox{Pingan_data_Ship!}{Pingan_data_Ship!}{\null} &Ship       	&           14680 	&          9787	&          24468    	&      48935 \\  
2	&     \fcolorbox{Pingan_data_Seawater!}{Pingan_data_Seawater!}{\null} & Seawater    	&            173434	&         115622 	&        289057   	&      578113 \\ 
3 	&     \fcolorbox{Pingan_data_Trees!}{Pingan_data_Trees!}{\null} &Trees   	&               2504  	&         1669	&           4172  	&        8345 \\  
4  	&    \fcolorbox{Pingan_data_Concretestructurebuilding!}{Pingan_data_Concretestructurebuilding!}{\null} &Concrete structure building   	&    26692  	&        17794   	&       44487   	&       88973  \\   
5 	&    \fcolorbox{Pingan_data_Floatingpier!}{Pingan_data_Floatingpier!}{\null} & Floating pier	&              6228 	&          4152 	&         10379  	&        20759 \\  
6 	&    \fcolorbox{Pingan_data_Brickhouses!}{Pingan_data_Brickhouses!}{\null} & Brick houses 	&              4226	&           2817 	&          7043  	&        14086\\   
7	&     \fcolorbox{Pingan_data_Steelhouses!}{Pingan_data_Steelhouses!}{\null} & Steel houses 	&              4197  	&         2798  	&         6996  	&        13991\\   
8	&      \fcolorbox{Pingan_data_Wharfconstructionland!}{Pingan_data_Wharfconstructionland!}{\null} &Wharf construction land   	&     24934  	&        16623   	&       41556  	&        83113\\ 
9 	&    \fcolorbox{Pingan_data_Car!}{Pingan_data_Car!}{\null} & Car     	&              2432   	&       1622 	&          4054 	&          8108 \\  
10 	&     \fcolorbox{Pingan_data_Road!}{Pingan_data_Road!}{\null} &Road   	&               82954  	&        55303  	&        138257  	&       276514  \\     
\hline
- 	  &- & Total    	&       342281   	&      228187  	&       570469   	&     1140937
\\
\hline
\end{tabular}}
\label{tab:PinganData}
\end{table}

\subsection{Experimental Setting}
\label{subsec:exp_set}

This section describes the comparative approaches and experimental settings used to evaluate the proposed KAN-based model. overall accuracy (OA), Average accuracy (AA), Kappa accuracy ($\kappa$), and per-class accuracies are calculated across all HSI datasets. The percentage of accurately mapped samples is the main focus of overall and average accuracy. On the other hand, kappa ($\kappa$) accuracy comes from statistical testing and offers information about how well classification models function in comparison to random selection. Essentially, the accuracy of Kappa ($\kappa$) depends on the number of classes in the dataset and the probability that sample points will be assigned a random label. As such, it functions as a more reliable accuracy metric than OA and AA, which could be deceptive in instances of unbalanced datasets. Comparative analysis against state-of-the-art methods was conducted to assess the effectiveness of KAN models. In more detail, the HSI classification results obtained by the KAN models are evaluated to several other models, including 1D-CNN, 2DCNN, 3D CNN, VGG-16 \cite{simonyan2015deep}, ResNet-50 \cite{He_2016_CVPR}, EfficientNet \cite{Koonce2021}, RNN \cite{Mou752}, and ViT \cite{Alexey20}.

\begin{table*}[!t]
\centering
\caption{Classification results in terms of OA, AA, and Kappa (in \%) obtained on the Tangdaowan dataset.}
\resizebox{0.99\linewidth}{!}{
\begin{tabular}{|c|c|c|c|c|c|c|c|c|c|c|c|c|c|c|}
\hline
Class No. & 1DCNN & 2DCNN & 3DCNN& VGG16 \cite{simonyan2015deep}& ResNet50\cite{He_2016_CVPR} & EfficientNet\cite{Koonce2021} & RNN\cite{Mou752} & ViT\cite{Alexey20} & 1DKAN \cite{liu2024kan}& 2DCKAN  & 3DKAN  & HybridKAN\\ 
\hline
\hline
1& 99.80& 99.82   &   99.45   &   99.96 & 99.93& 99.93&99.93& \textbf{100.0}&  99.95   & 99.91&     98.81 & 99.85\\

2&98.42&  99.49&  87.00  &    \textbf{99.75} & 97.49&  99.48&  97.83&  99.56&  99.52    & 99.52  &    95.91 & 99.48\\

3 &94.98&  96.62&   86.46  &   97.63& \textbf{98.54}& 97.02& 92.37 &  92.96&   97.31   & 96.74   &   92.33 & 97.24\\

4&99.17&  99.89&  92.02 &    \textbf{99.96}& 99.89& 99.92 &  98.26&  99.20&   99.78   & 99.60   &   98.52 & 99.71\\

5&  95.38&  97.42&  87.20  &    99.78& \textbf{99.89}& 99.24& 97.63&  97.85&   97.88   & 97.63   &   38.86 & 99.35\\

6&  88.54&   94.66&  86.84  &    94.38& 93.82& 92.71&  90.92&  94.81&  \textbf{97.21}   & 96.87   &   90.61 & 97.34\\ 

7&81.14&90.74&  40.00  &   81.92& 94.70& 96.06&  78.70 &  91.25   &   93.84   & 90.72   &   72.72 & \textbf{95.78}\\ 

8&  99.81&   99.92&  98.34  &   99.83& \textbf{99.99}& 99.93&  99.84&  99.80&   99.97   & 99.94    &   98.85& 99.89\\ 

9&   96.89&  99.35&  90.84  &   92.96& 99.79& \textbf{99.95}& 97.55&  99.70&  99.45    & 99.44&   94.33& 99.51\\

10&   92.81& \textbf{98.54}&  82.57  &   98.42&  96.18& 85.18&  97.75&  93.93&  95.22  & 95.72  & 70.38  & 97.75\\ 

11&   64.80&  84.77&  46.78  &   94.76& 84.95& \textbf{96.83}&  64.72&  86.68&  87.24   & 91.83  &   45.16& 90.34\\

12&  77.54&  86.89&  30.87  &   93.04& 92.78& 87.43& 89.83& 78.87&  94.18    & 91.25   &   56.99 & \textbf{95.85}  \\

13&  98.22&   \textbf{100.0}&  91.91  &   \textbf{100.0}& 99.88&  99.52&  99.76  &   98.81&   99.70    & 99.53  &   96.07 & 99.06\\

14& 83.74&   98.41&  37.09  &   89.16& 95.25& \textbf{100.0}&  86.00&  98.64&   97.69   & 95.67  &  45.74 & 97.25\\

15& 79.71&   96.40&  79.31  &   \textbf{97.08}& 96.81&  95.83& 83.05&  96.16&   94.07    & 93.68 &  77.42 & 95.04\\

16& 93.56&   96.35&  86.28  &   96.85& \textbf{98.99}&  96.38&  92.78&  95.66 & 96.52      & 97.16  &   88.58 & 97.20\\ 

17&  65.53 &    92.51&  74.53  &   \textbf{97.44}& 97.12& 96.89 & 67.41& 91.83& 90.79     & 94.20     &    71.58 & 93.66\\ 

18 &  99.84& 99.93&  98.54 & 99.90&  99.91& \textbf{99.99}  & 99.83&   99.69&   99.86   & 99.94 &  98.76& 99.91 \\ 

\hline
\hline
OA&   93.81&   97.32    & 87.63  &  97.43&  \textbf{98.09}&  97.89 &   93.59&  96.90          &97.68 &  97.90  &   91.11     & 98.08 \\

AA&   89.44 &   96.21   & 75.61  &  96.27& 96.99&   96.79&  90.79&  95.30 & 96.15  & 96.47  &  76.14  & \textbf{97.12}\\

$\kappa (\times 100)$&    92.93&  96.95   &  85.82 &  97.08&  \textbf{97.82} &   97.61&    92.69&  96.48&    97.36 & 97.61  &  89.80  & 97.81 \\
\hline
\end{tabular}}
\label{tab:Tangdaowan}
\end{table*}

\begin{table*}[!t]
\centering
\caption{Classification results in terms of OA, AA, and Kappa (in \%) obtained on the Pingan dataset.}
\resizebox{0.99\linewidth}{!}{
\begin{tabular}{|c|c|c|c|c|c|c|c|c|c|c|c|c|c|c|}
\hline
Class No. & 1DCNN & 2DCNN & 3DCNN& VGG16 \cite{simonyan2015deep}& ResNet50\cite{He_2016_CVPR} & EfficientNet\cite{Koonce2021} & RNN\cite{Mou752} & ViT\cite{Alexey20} & 1DKAN \cite{liu2024kan} &2DKAN  & 3DKAN  & HybridKAN\\ 
\hline
\hline
1&  78.73& 93.17  & 73.12    &   \textbf{97.04}& 95.23&  89.84&  73.72 & 85.26&  91.44  & 91.14 &   75.78 & 92.91\\

2 &  98.84& 99.59&  98.58    & 99.36 &  99.59& 99.56 &  98.91& \textbf{99.61}&   99.47  & 99.36 &  99.18  & 99.50\\

3& 95.18& 99.08&  80.16   & 99.25&  \textbf{99.83}&  97.69 &  95.97&  95.39 &   97.96  & 93.64  &  89.33  & 96.98\\

4& 83.38& 98.11&  60.58   &  \textbf{99.69}& 98.27&  98.59&  80.03& 97.88&  96.84  & 96.45    &   83.21 & 97.54\\

5& 68.01&  95.00&   58.77   & \textbf{96.03}&  94.44&  93.94&  67.41&  95.36&  91.39  & 88.50&  72.72  & 93.42\\

6 & 88.20&  \textbf{98.35}&   53.40  &   98.75&   98.42&  97.58&  89.49& 97.48& 97.54 & 95.81&   86.52 & 97.30\\ 

7 &  87.36&  98.77&   81.34    & \textbf{98.88}&   99.02 &  97.29&  85.37& 98.69 & 98.63 & 96.68 &   79.45 & 98.15\\ 

8& 86.74 &  97.22&  76.71    & 98.39&   \textbf{98.97}&  98.27&  84.12& 94.06& 96.53  & 95.91& 83.43 & 96.87\\ 

9& 54.66&  90.94&   30.67   &  \textbf{94.35}&   91.39&  89.93 &  43.88&91.36& 86.52  & 82.81 &  50.59  & 89.11\\ 

10&  95.12 &  \textbf{99.36}&   88.00  &   99.18&  98.67&  99.05 &  95.48 & 98.89 & 98.74  & 98.30&  94.45  & 98.91\\ 

\hline
\hline
OA & 93.82&  98.79 &  88.67    &\textbf{99.06}&  98.86& 98.61&  93.18&98.08   &  98.24  & 97.84&  93.36  & 98.48\\

AA &  83.62&  96.96 &   67.37   & \textbf{98.09}&  97.38&96.18&81.44&  95.40 &  95.42  & 94.14   &   81.53 & 95.95\\

$\kappa (\times 100)$ & 90.77& 98.20 &  83.01   & \textbf{98.61}&  98.31&97.93 &   89.80& 97.13 & 97.38  & 96.78& 90.09  & 97.74\\
\hline
\end{tabular}}
\label{tab:Pingan}
\end{table*}

\section{Results}
\label{sec:results}

\subsection{Statistical Results}

\begin{figure*}[!t]
\centering
	\begin{subfigure}{0.12\textwidth}
		\includegraphics[width=0.99\textwidth]{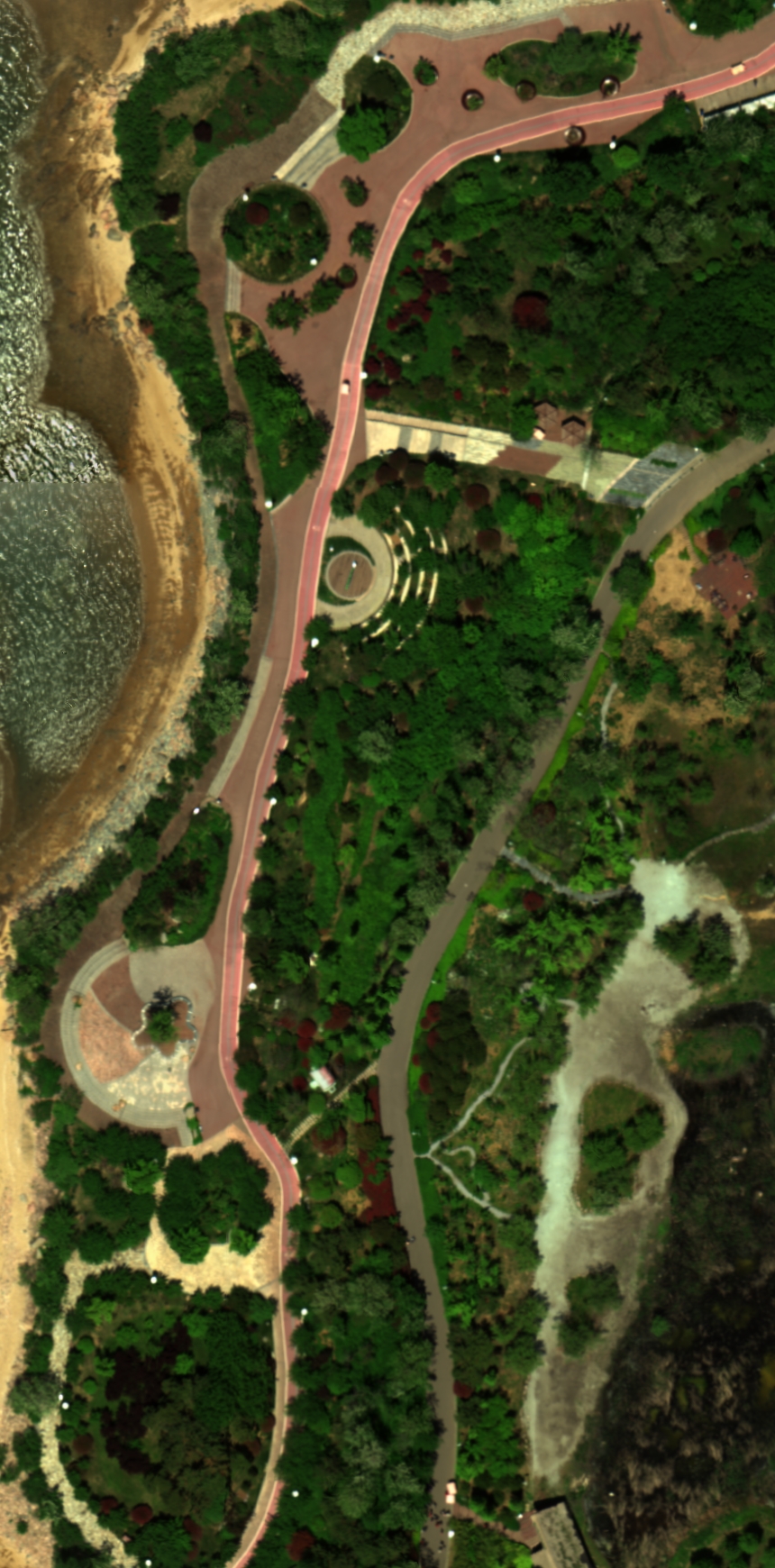}
		\caption{RGB image}
	\end{subfigure}
	\begin{subfigure}{0.12\textwidth}
		\includegraphics[width=0.99\textwidth]{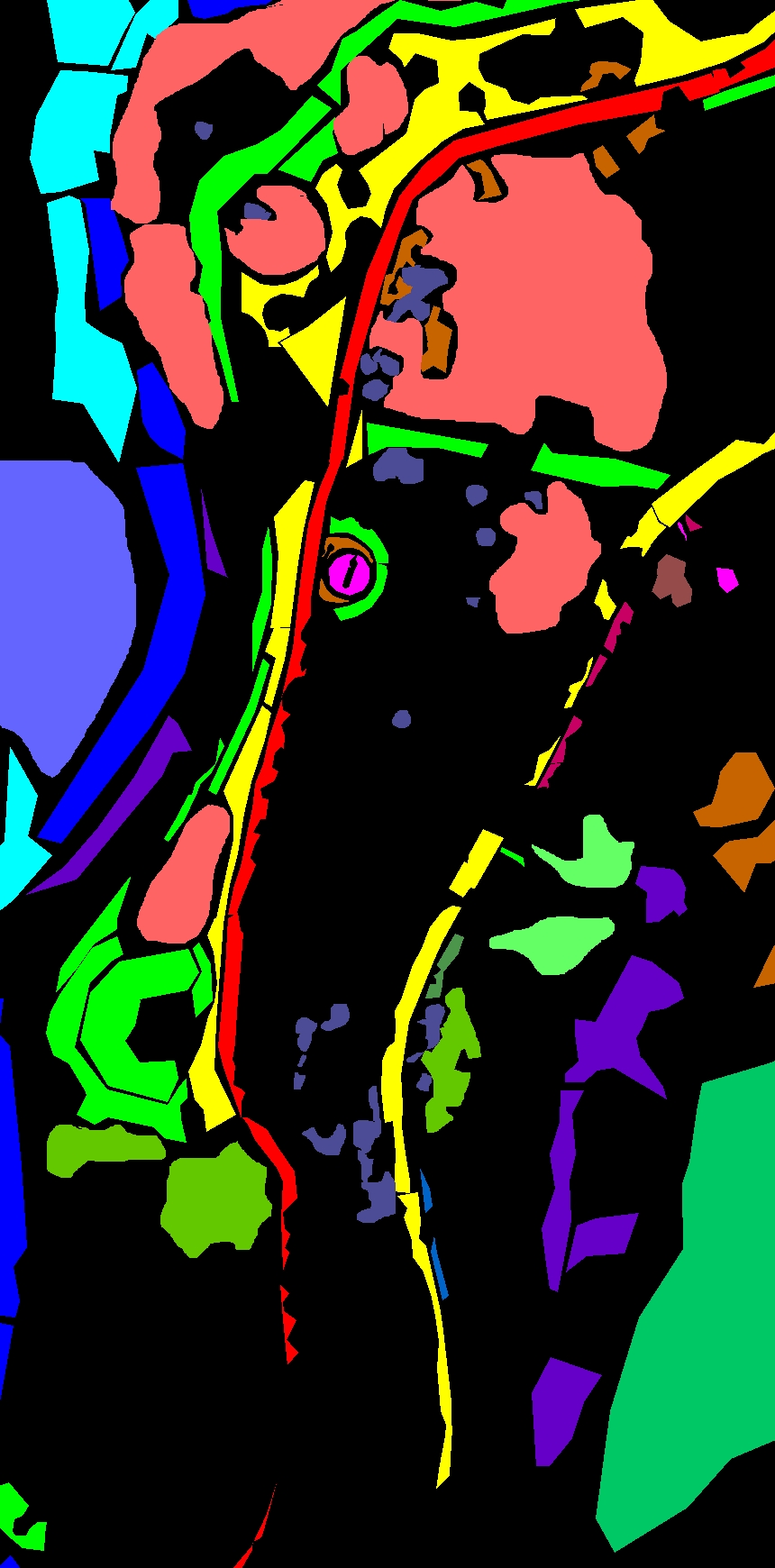}
		\caption{GT}
	\end{subfigure}
	\begin{subfigure}{0.12\textwidth}
		\includegraphics[width=0.99\textwidth]{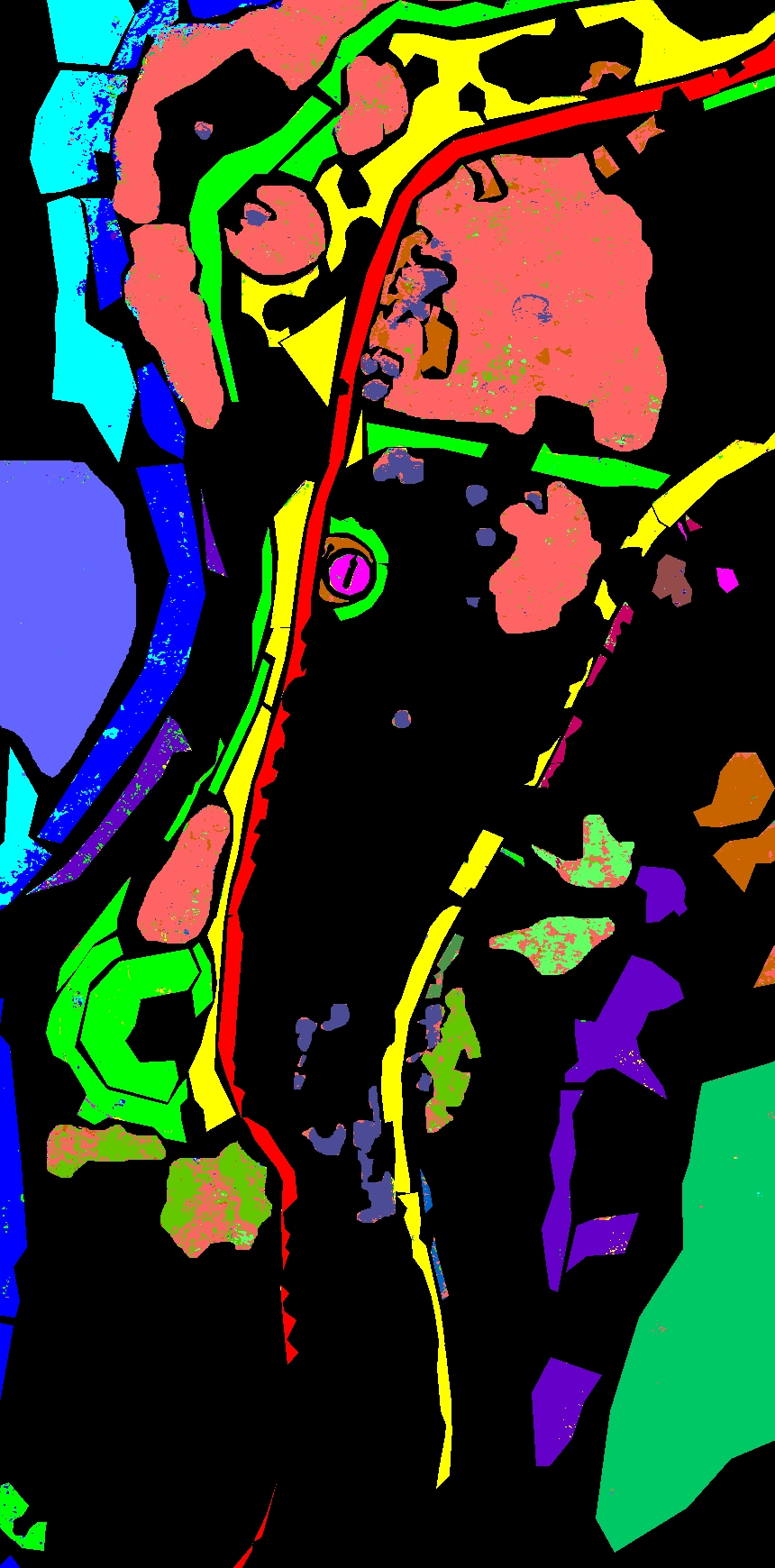}
		\caption{1D-CNN}
	\end{subfigure}
	\begin{subfigure}{0.12\textwidth}
		\includegraphics[width=0.99\textwidth]{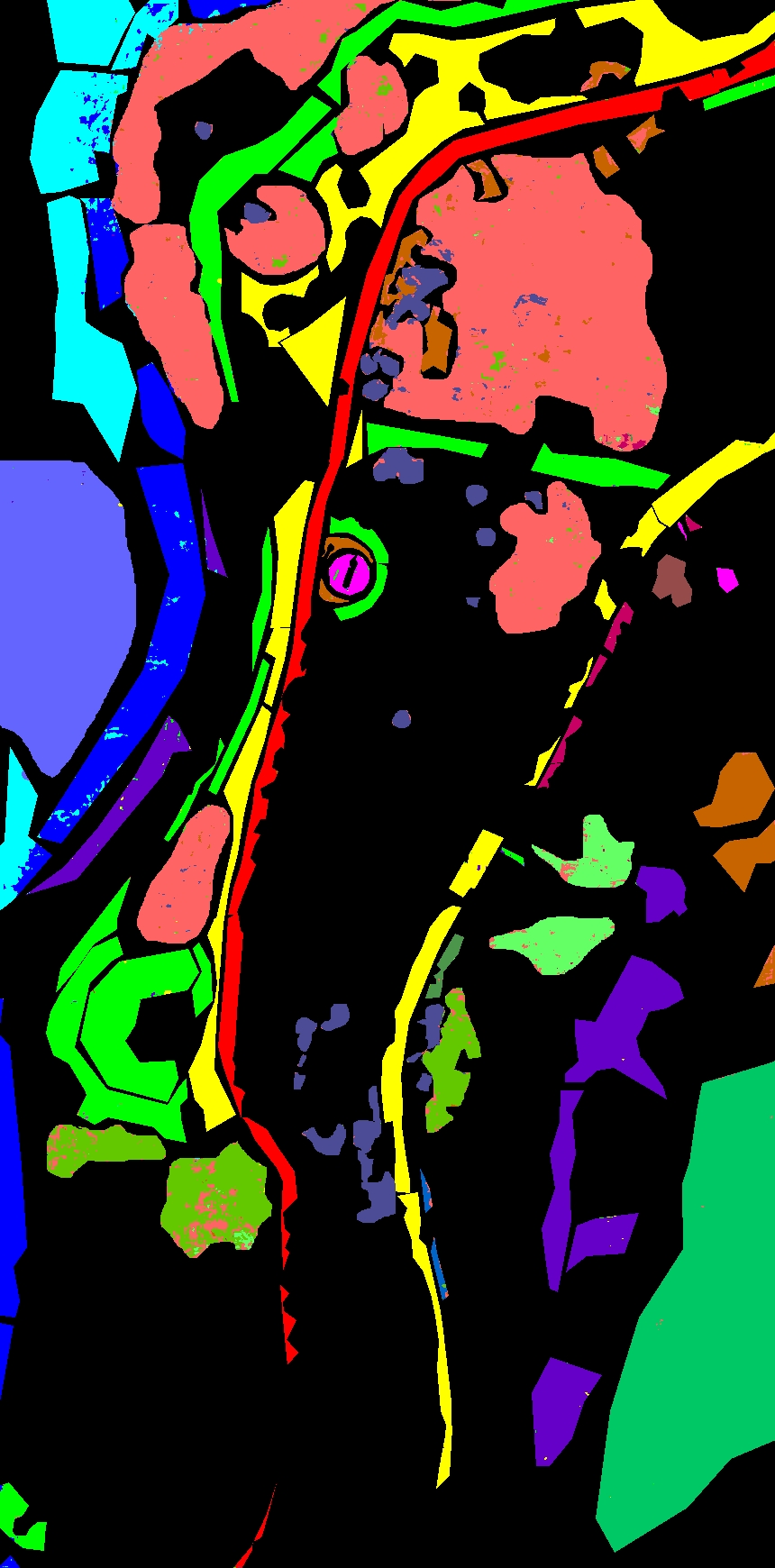}
		\caption{2D-CNN}
	\end{subfigure}
 	\begin{subfigure}{0.12\textwidth}
		\includegraphics[width=0.99\textwidth]{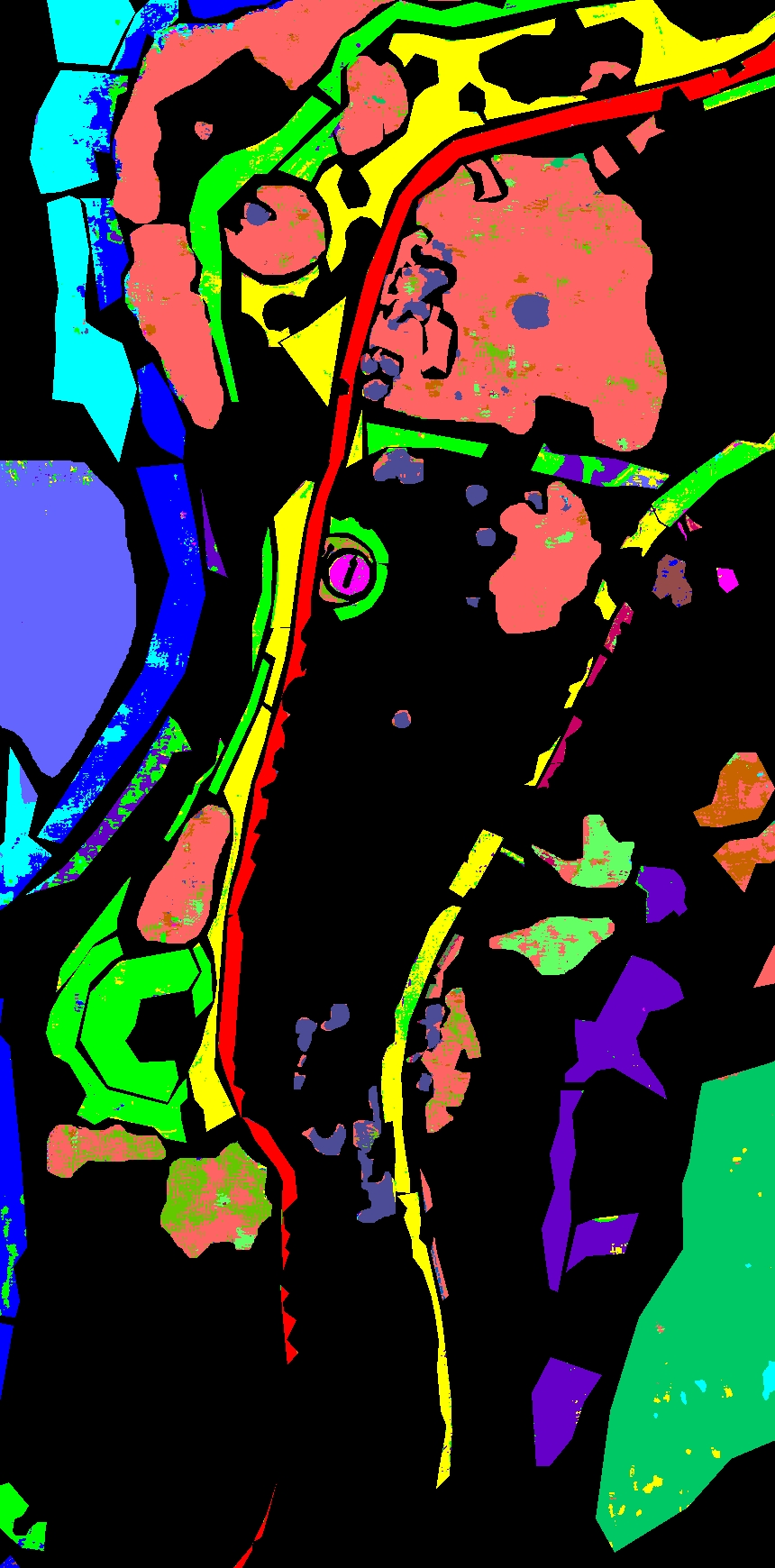}
		\caption{3D-CNN}
	\end{subfigure}
	\begin{subfigure}{0.12\textwidth}
		\includegraphics[width=0.99\textwidth]{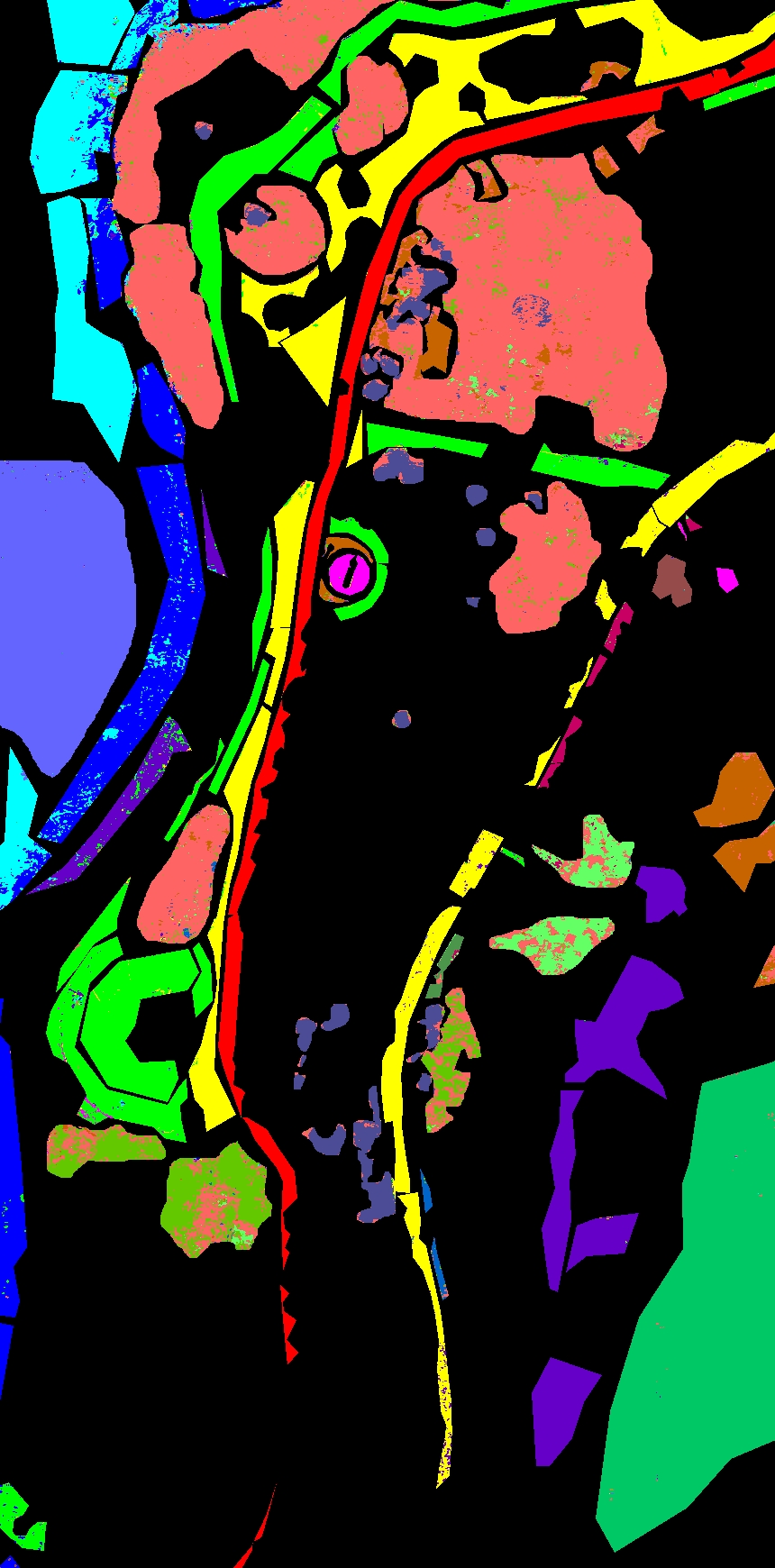}
		\caption{DRNN} 
	\end{subfigure}
	\begin{subfigure}{0.12\textwidth}
		\includegraphics[width=0.99\textwidth]{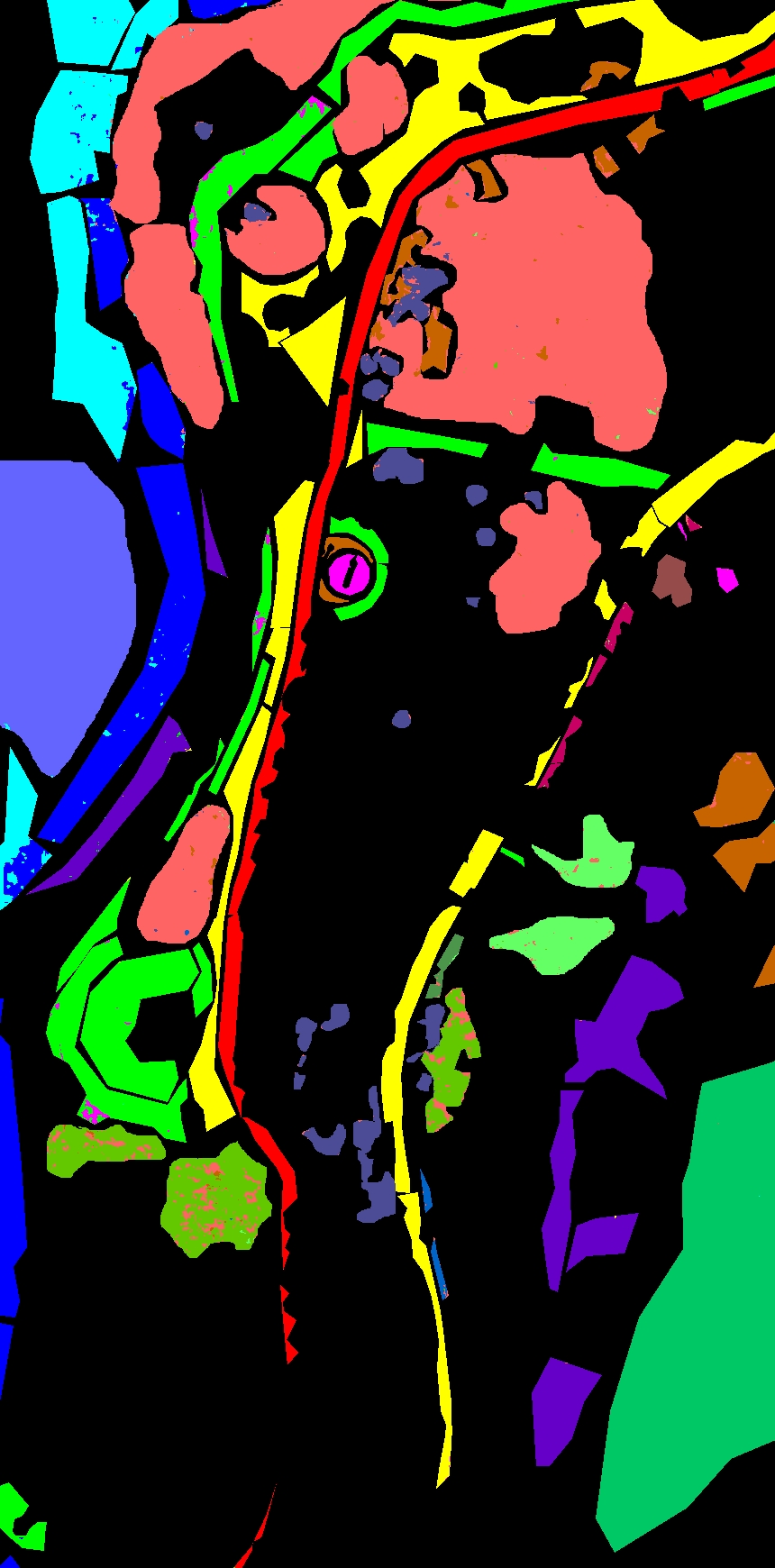}
		\caption{ResNet50} 
	\end{subfigure}
    \begin{subfigure}{0.12\textwidth}
		\includegraphics[width=0.99\textwidth]{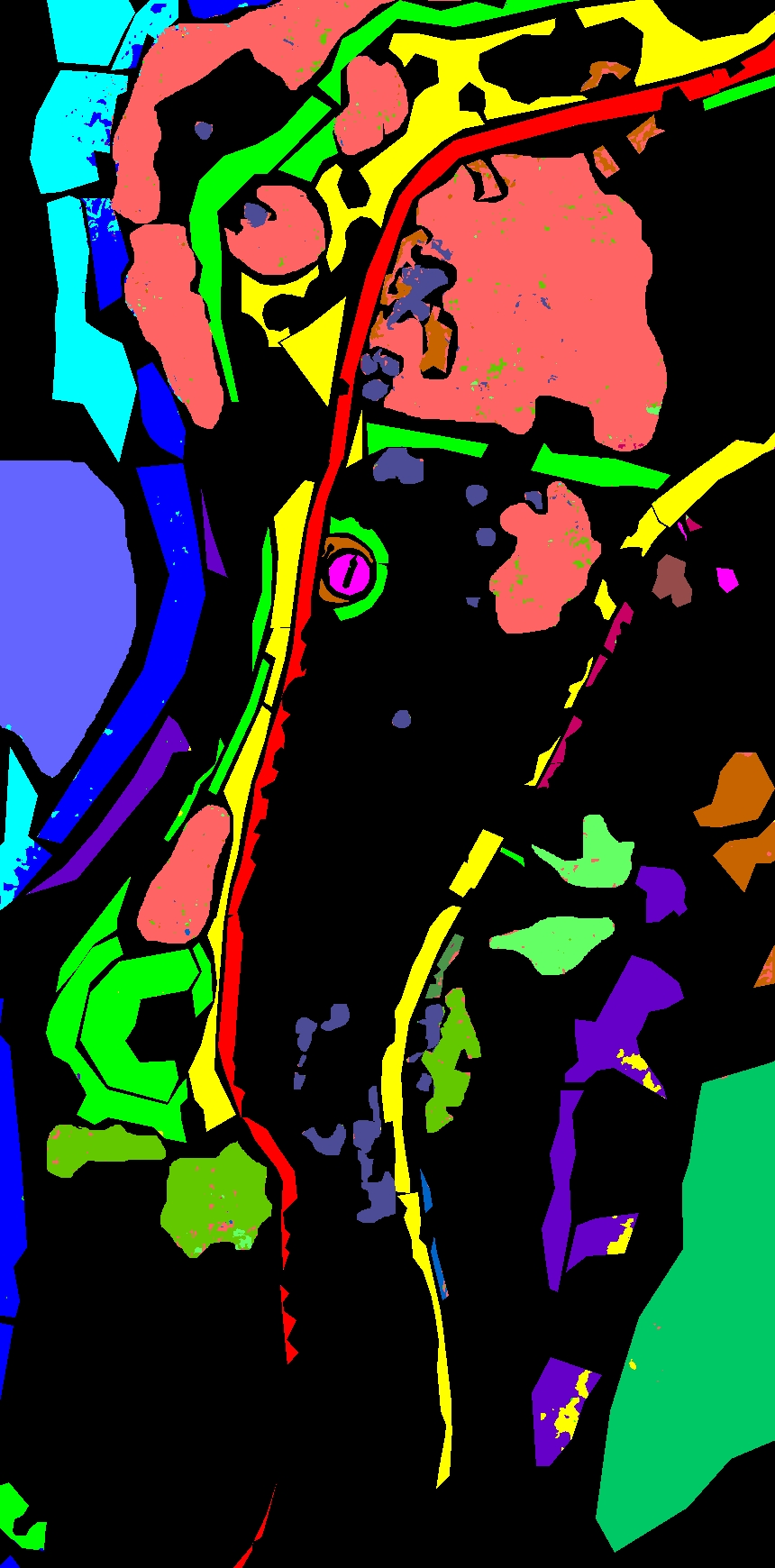}
		\caption{VGG-16}
	\end{subfigure}
    \begin{subfigure}{0.12\textwidth}
		\includegraphics[width=0.99\textwidth]{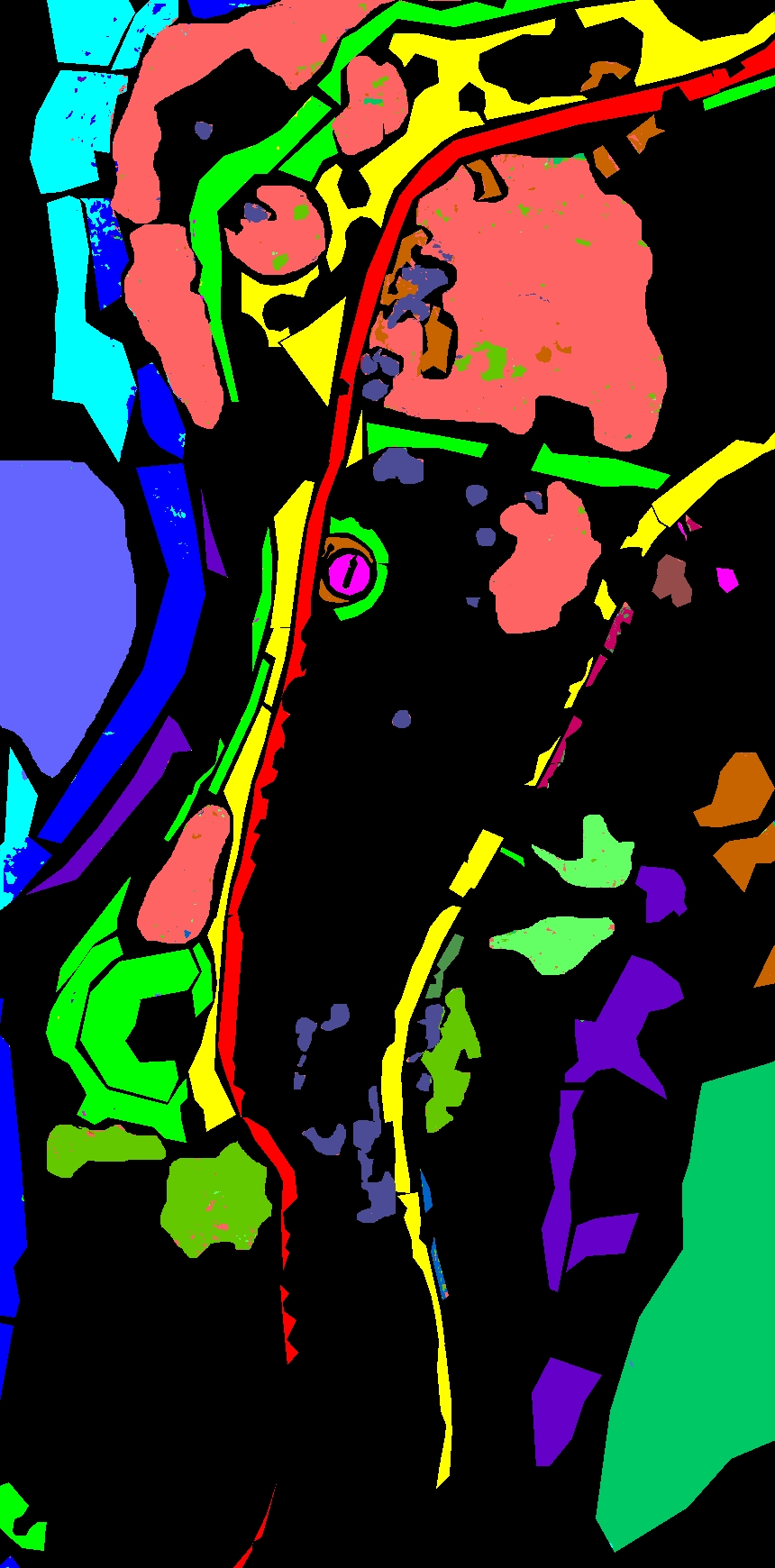}
		\caption{EfficientNet}
	\end{subfigure}
	\begin{subfigure}{0.12\textwidth}
		\includegraphics[width=0.99\textwidth]{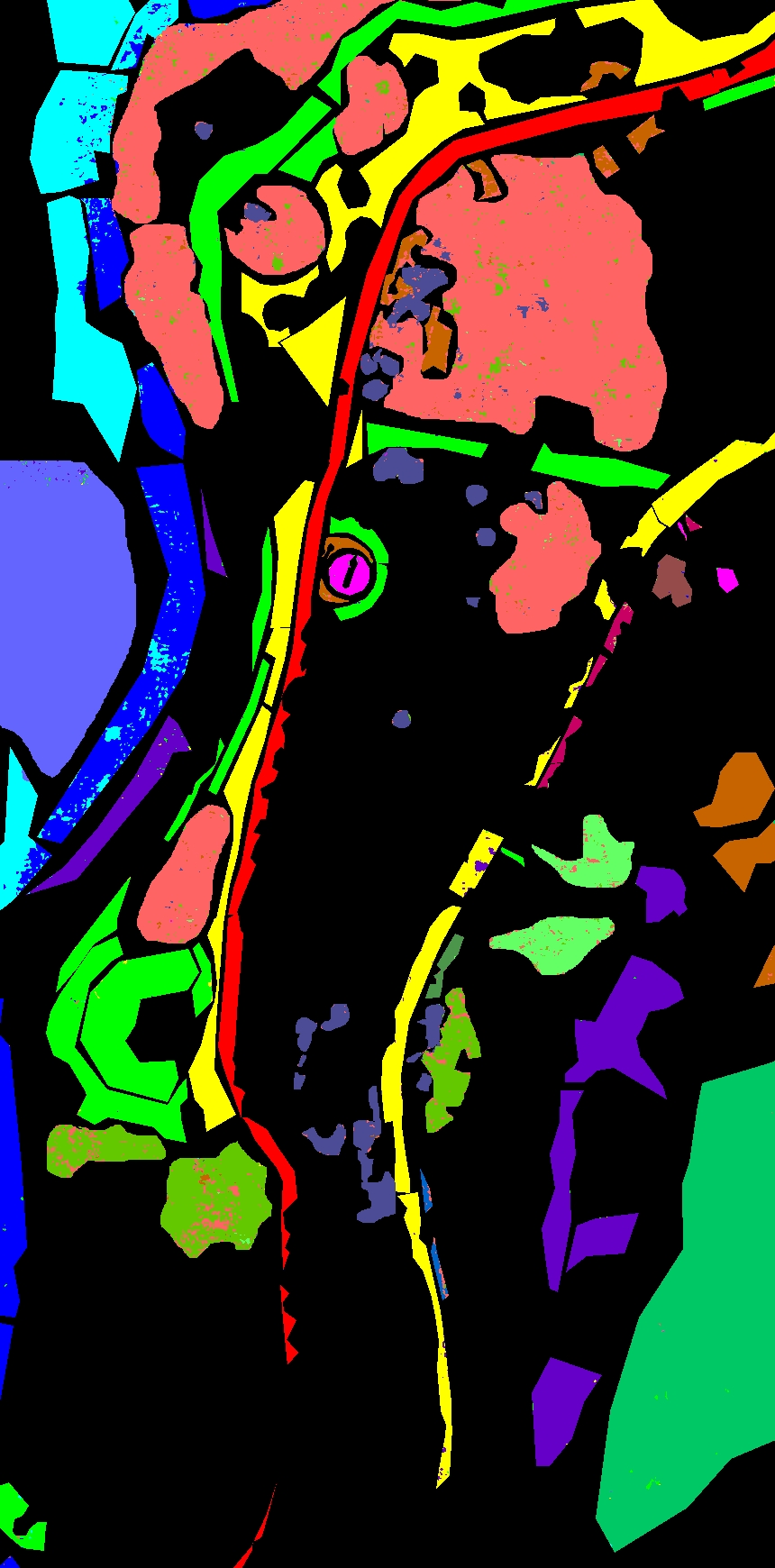}
		\caption{ViT}
	\end{subfigure}
	\begin{subfigure}{0.12\textwidth}
		\includegraphics[width=0.99\textwidth]{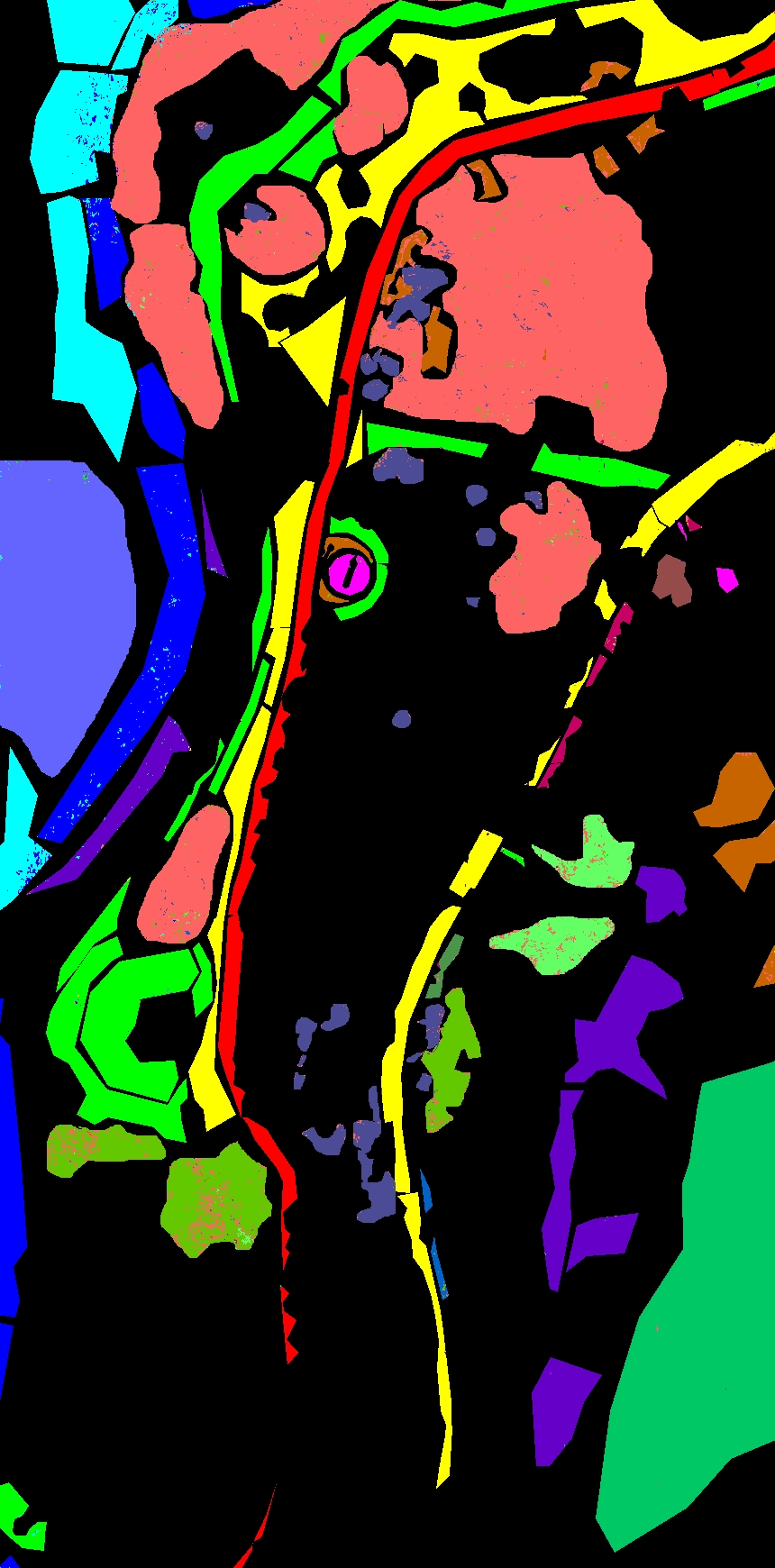}
		\caption{1D-KAN}
	\end{subfigure}
	\begin{subfigure}{0.12\textwidth}
		\includegraphics[width=0.99\textwidth]{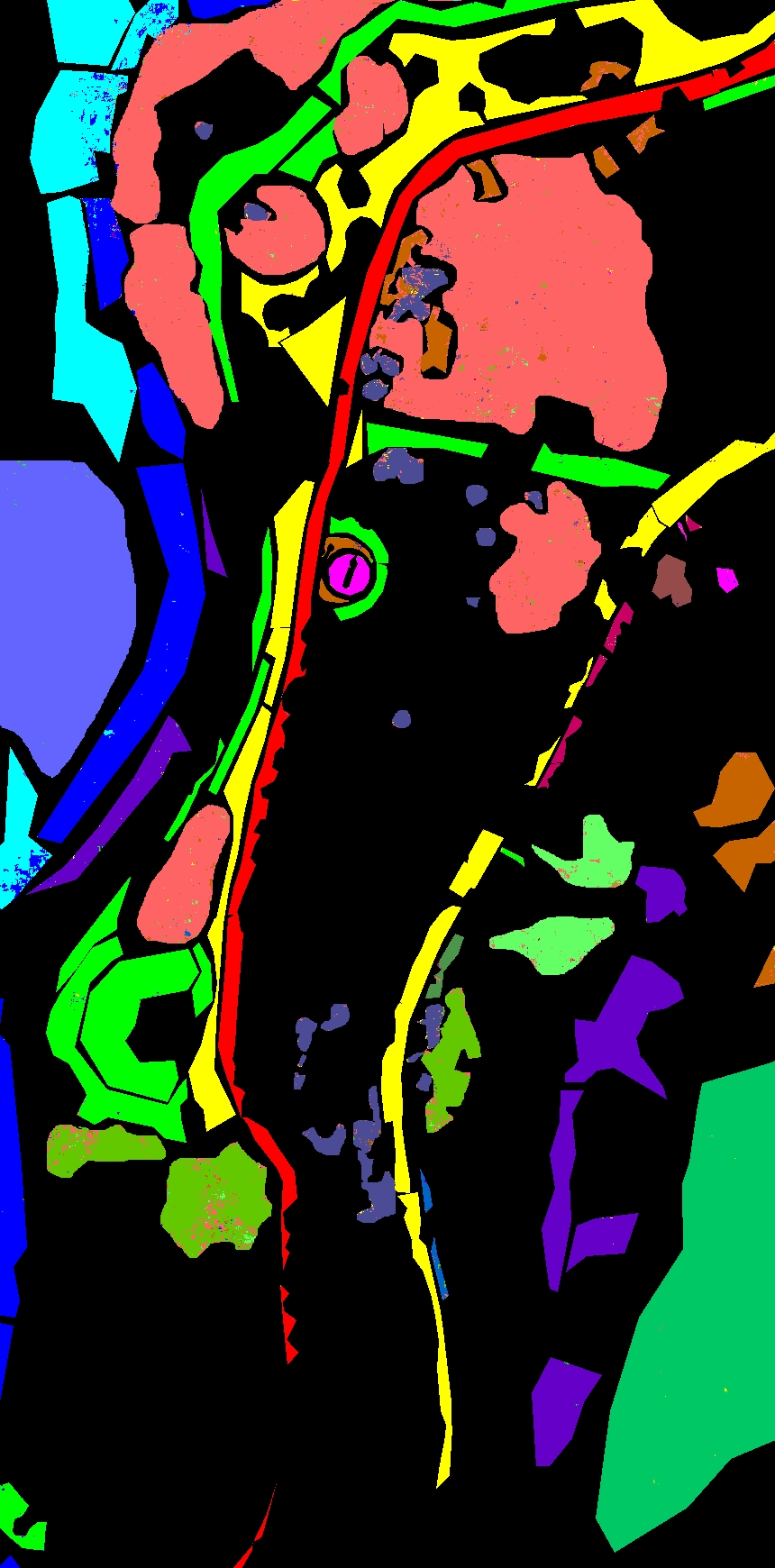}
		\caption{2D-KAN}
	\end{subfigure}
 	\begin{subfigure}{0.12\textwidth}
		\includegraphics[width=0.99\textwidth]{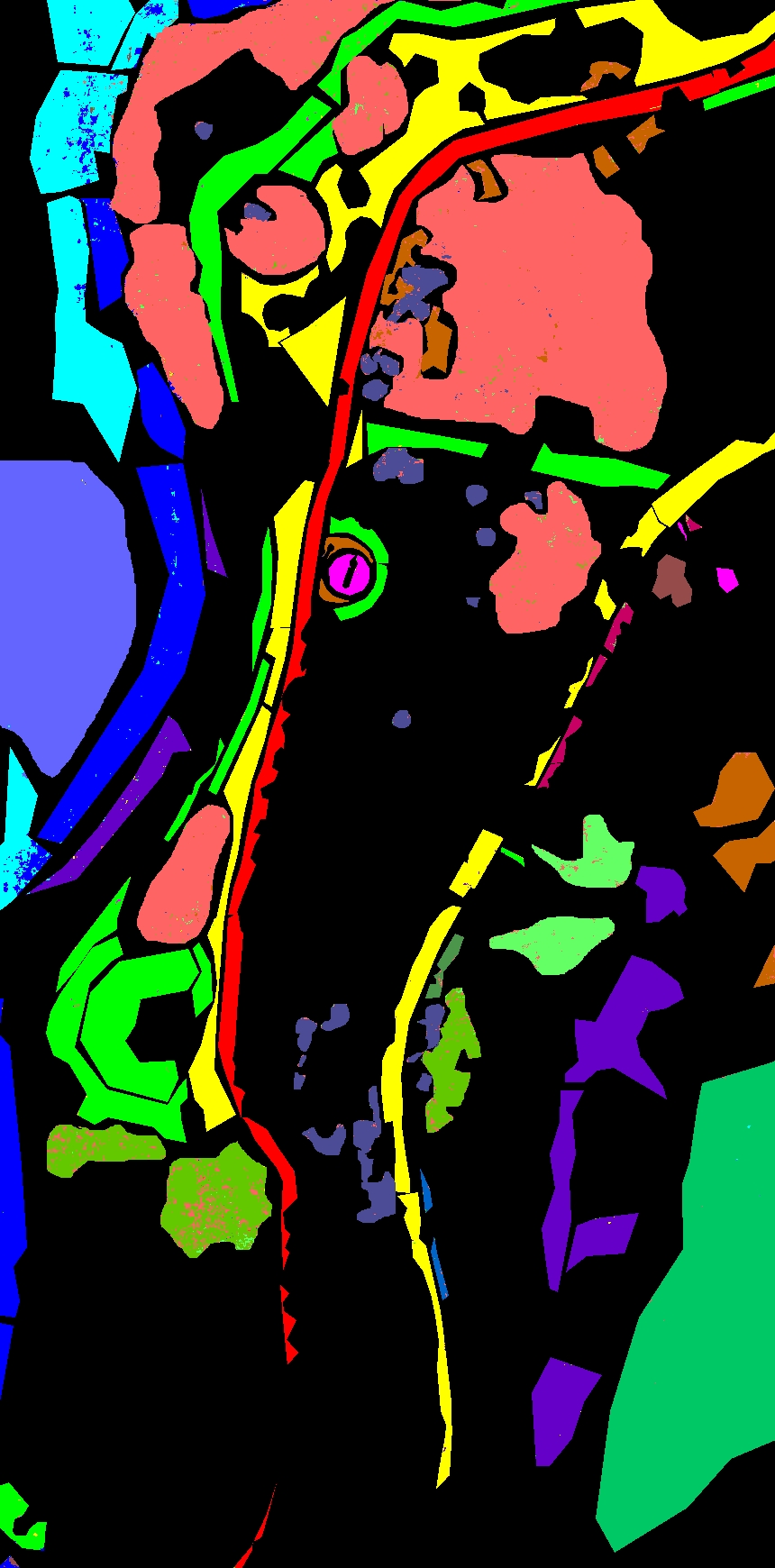}
		\caption{3D-KAN}
	\end{subfigure}
    \begin{subfigure}{0.12\textwidth}
		\includegraphics[width=0.99\textwidth]{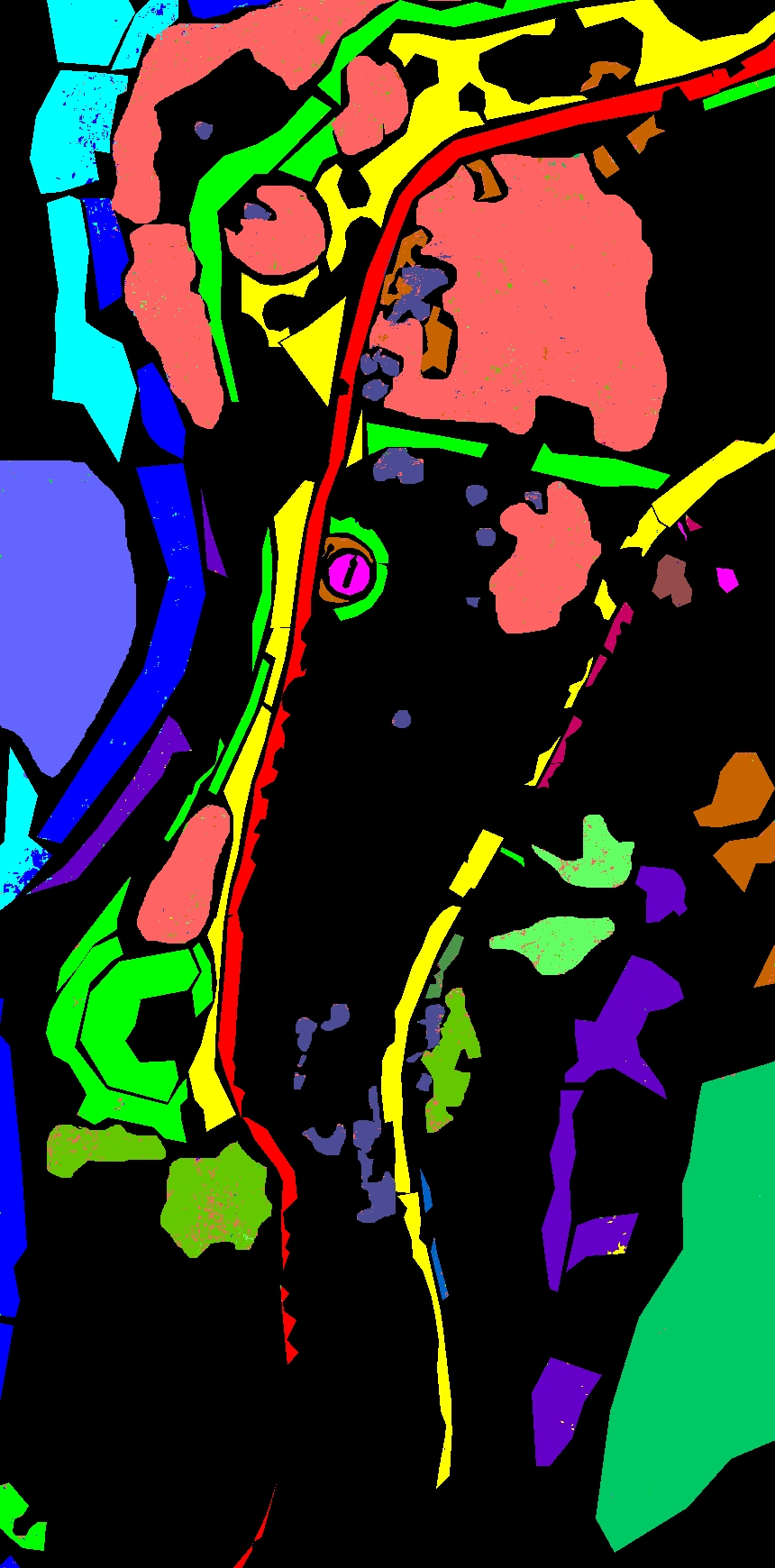}
		\caption{HybridKAN}
	\end{subfigure}
\caption{The predicted land cover maps obtained for the Tangdaowan HSI data set.}
\label{fig:Tangdaowan}
\end{figure*}

Table\ref{tab:Tangdaowan} and Fig\ref{fig:Tangdaowan} illustrate the HSI classification results and maps produced by the developed CNN- and transformer-based architectures in the Tangdaowan HSI dataset. The results revealed that the KAN models, specifically the developed HybridKAN architecture obtained a competitive HSI classification accuracy as compared to other well-known CNNs and ViTs. The developed HybridKAN achieved the highest average accuracy (97.12\%), while the ResNet-50 model achieved the best overall accuracy (98.09\%) and kappa value (97.82\%). The HSI data classification results underscored the effectiveness of the KAN models as compared to other classification models. The 3D KAN models similar to its counterpart of 3D CNN demonstrated the least HSI classification accuracy with an average accuracy of 76.14\% compared to that of 1D KAN (96.15\%), 2D KAN (96.47\%), and the HybridKAN (97.12\%). As seen in Fig\ref{fig:Tangdaowan}, the HybridKAN illustrated the most homogeneous classification map with much less noise as compared to the other classification architectures, showcasing its high capability in accurate HSI data classification.

\begin{figure*}[!ht]
\centering
	\begin{subfigure}{0.12\textwidth}
		\includegraphics[width=0.99\textwidth]{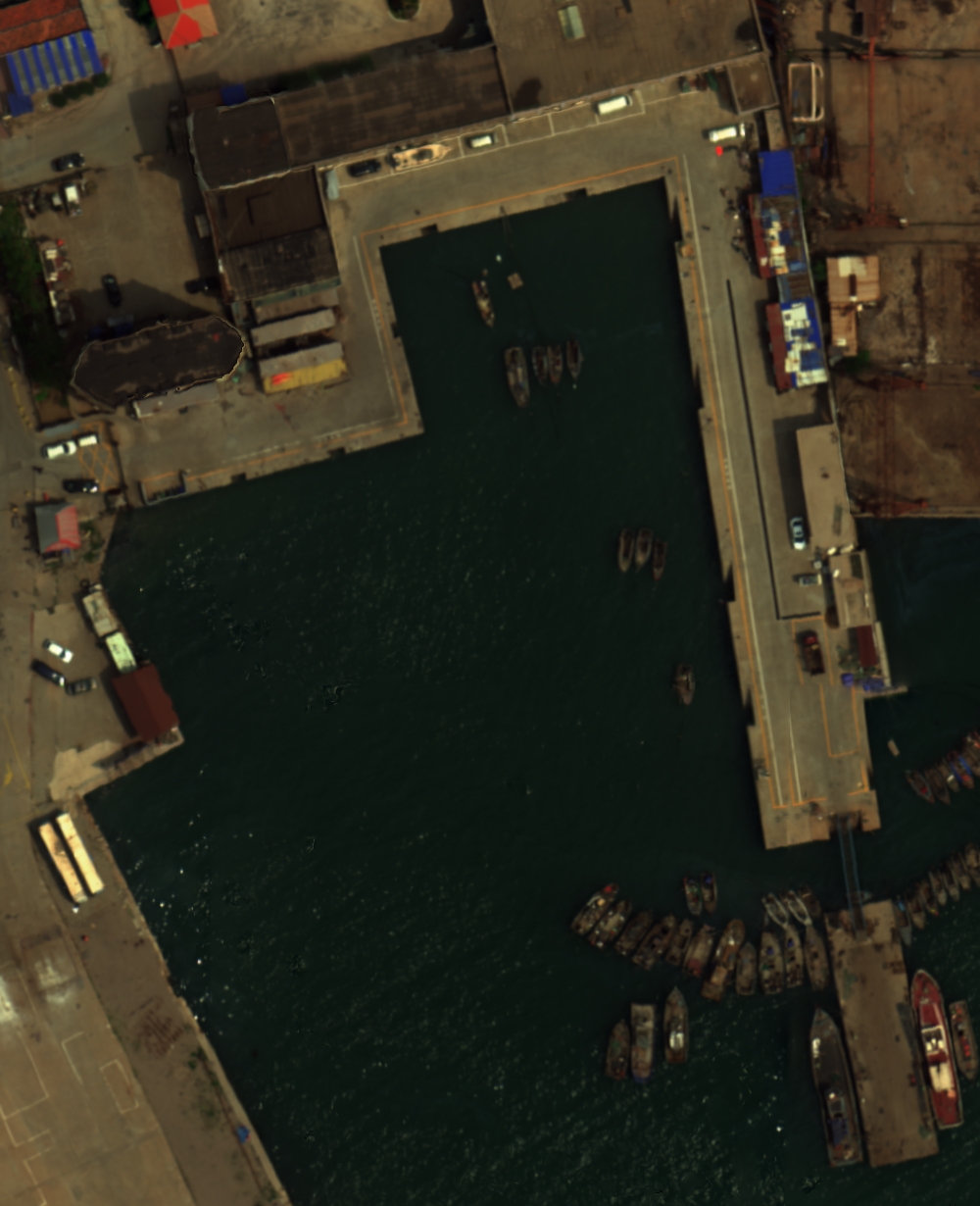}
		\caption{RGB image}
	\end{subfigure}
	\begin{subfigure}{0.12\textwidth}
		\includegraphics[width=0.99\textwidth]{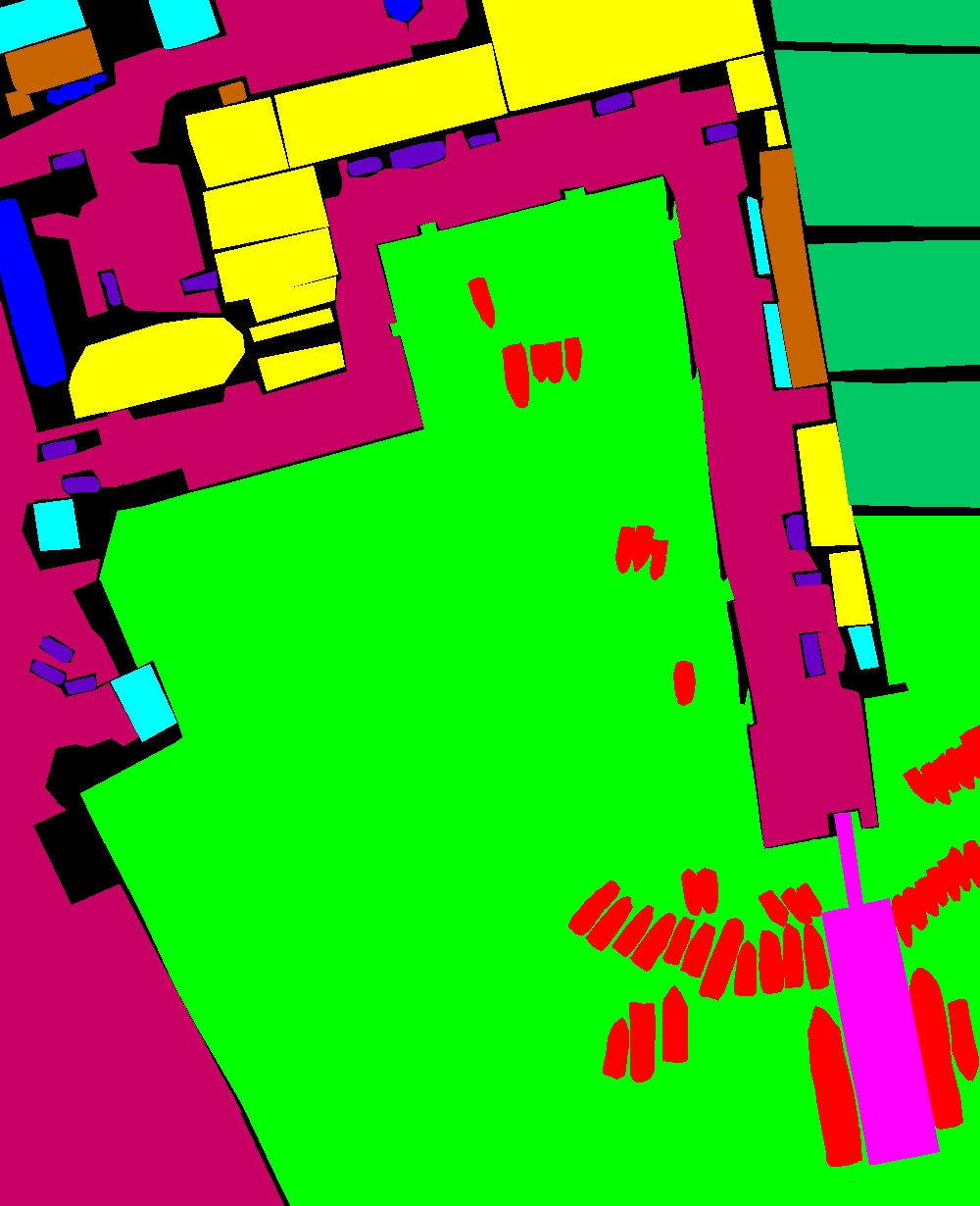}
		\caption{Ground Truth}
	\end{subfigure}
	\begin{subfigure}{0.12\textwidth}
		\includegraphics[width=0.99\textwidth]{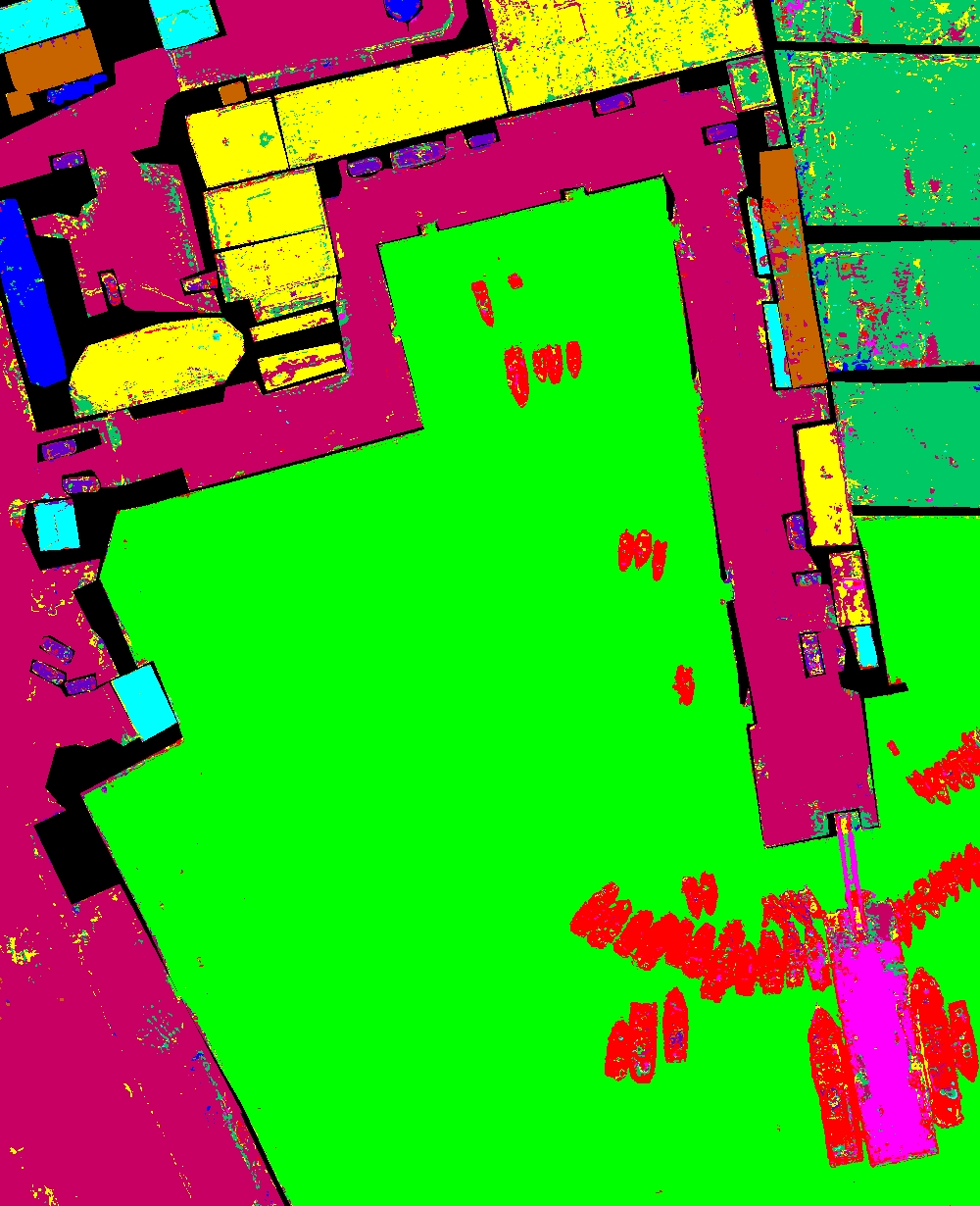}
		\caption{1D-CNN}
	\end{subfigure}
	\begin{subfigure}{0.12\textwidth}
		\includegraphics[width=0.99\textwidth]{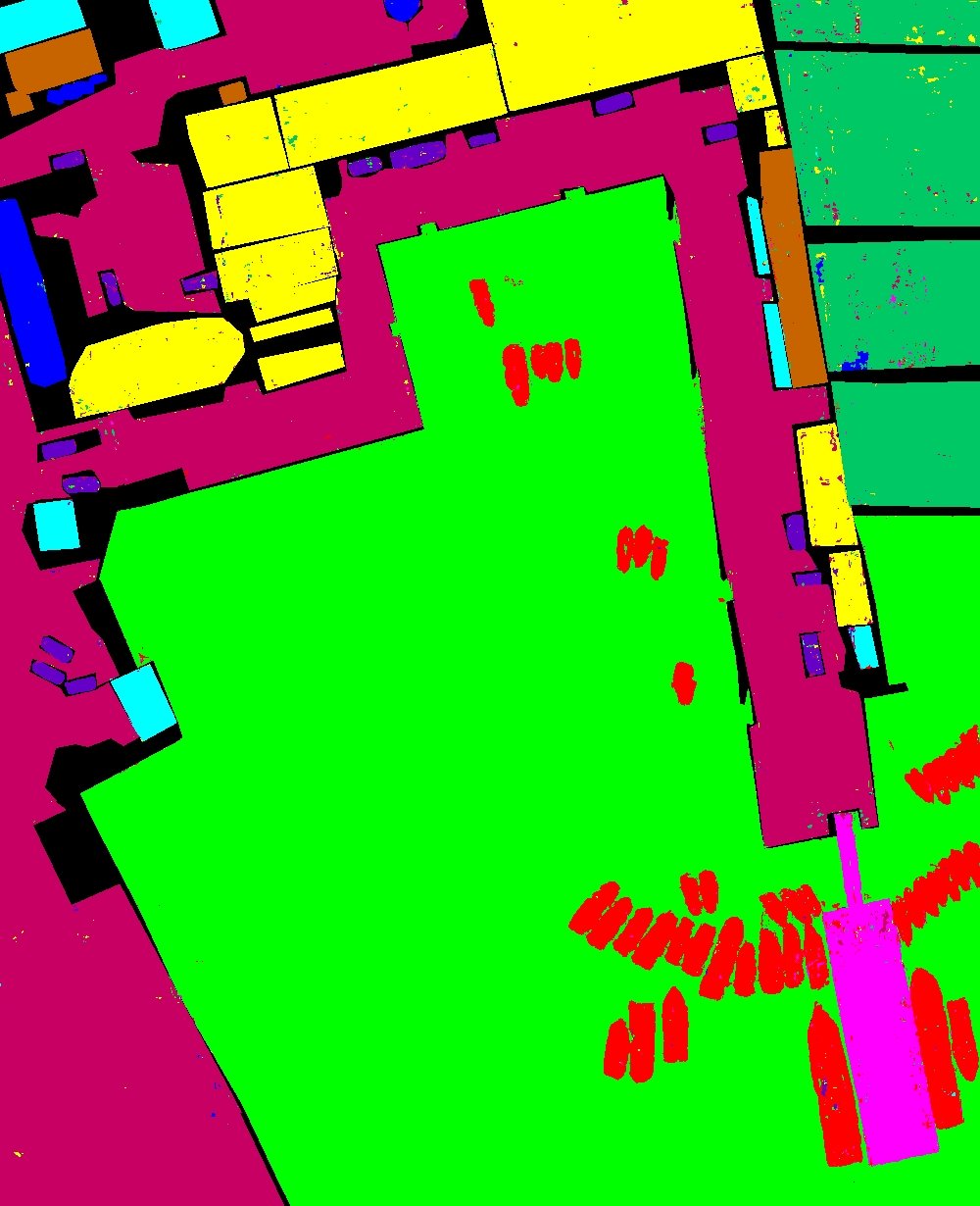}
		\caption{2D-CNN}
	\end{subfigure}
 	\begin{subfigure}{0.12\textwidth}
		\includegraphics[width=0.99\textwidth]{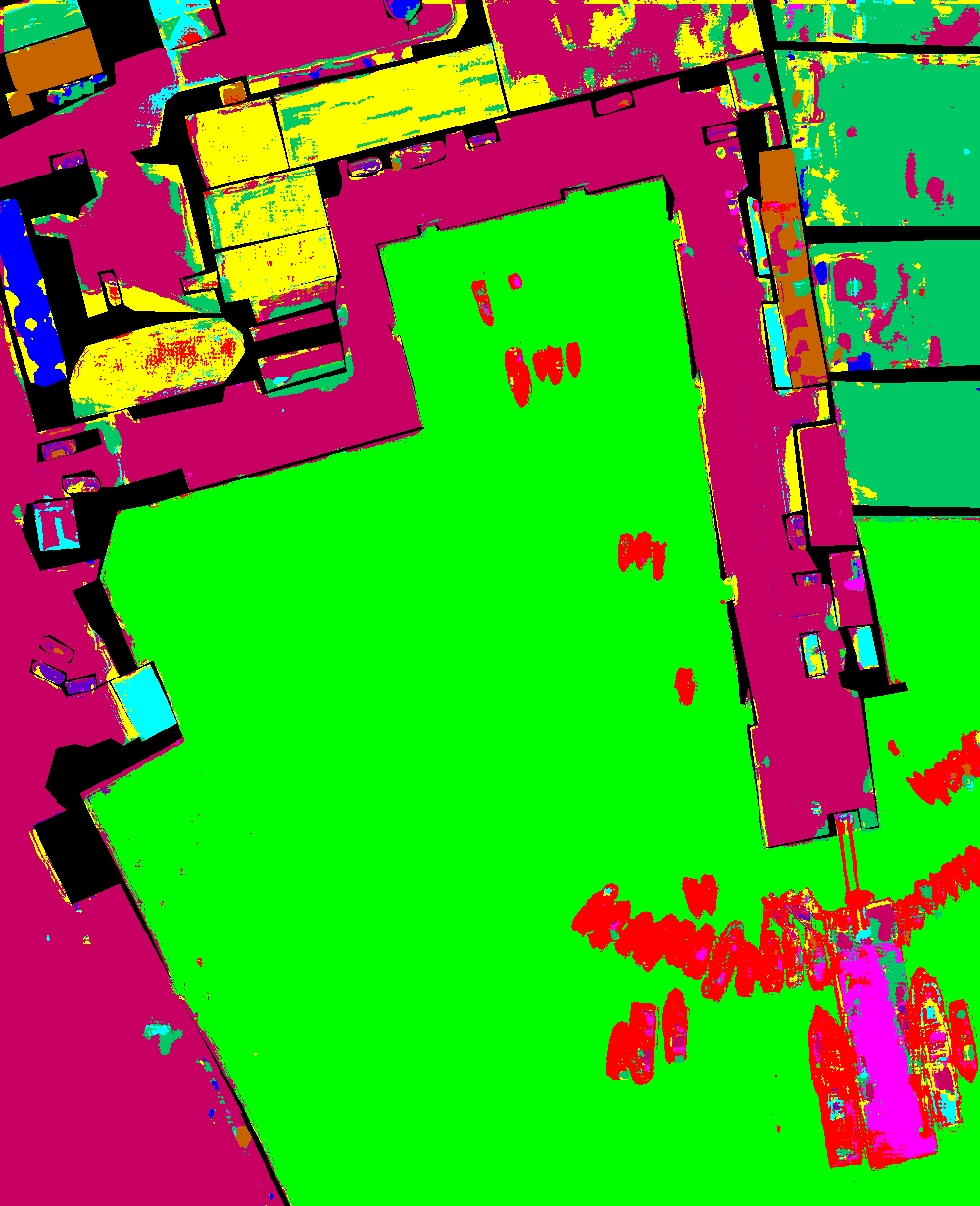}
		\caption{3D-CNN}
	\end{subfigure}
	\begin{subfigure}{0.12\textwidth}
		\includegraphics[width=0.99\textwidth]{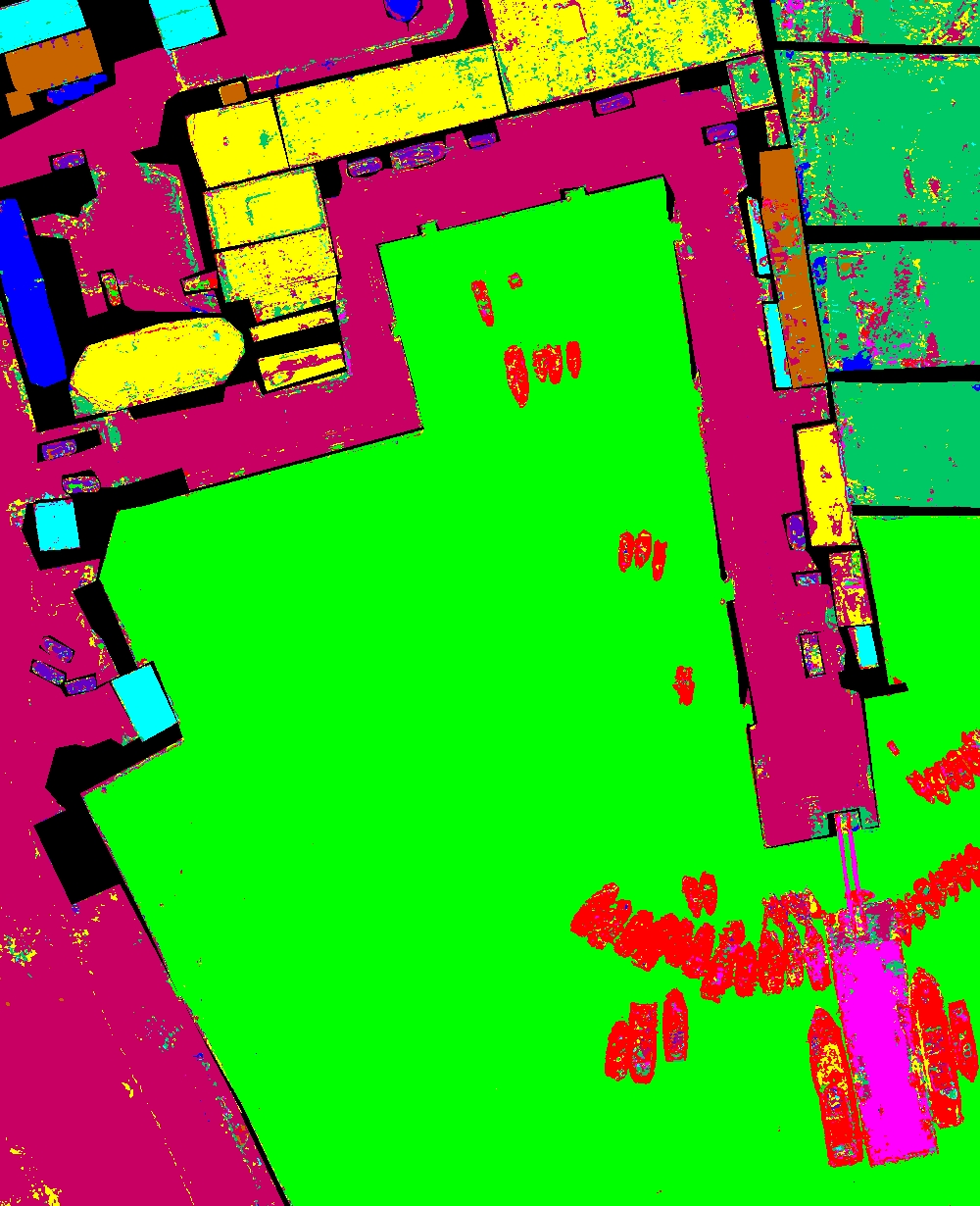}
		\caption{DRNN} 
	\end{subfigure}
	\begin{subfigure}{0.12\textwidth}
		\includegraphics[width=0.99\textwidth]{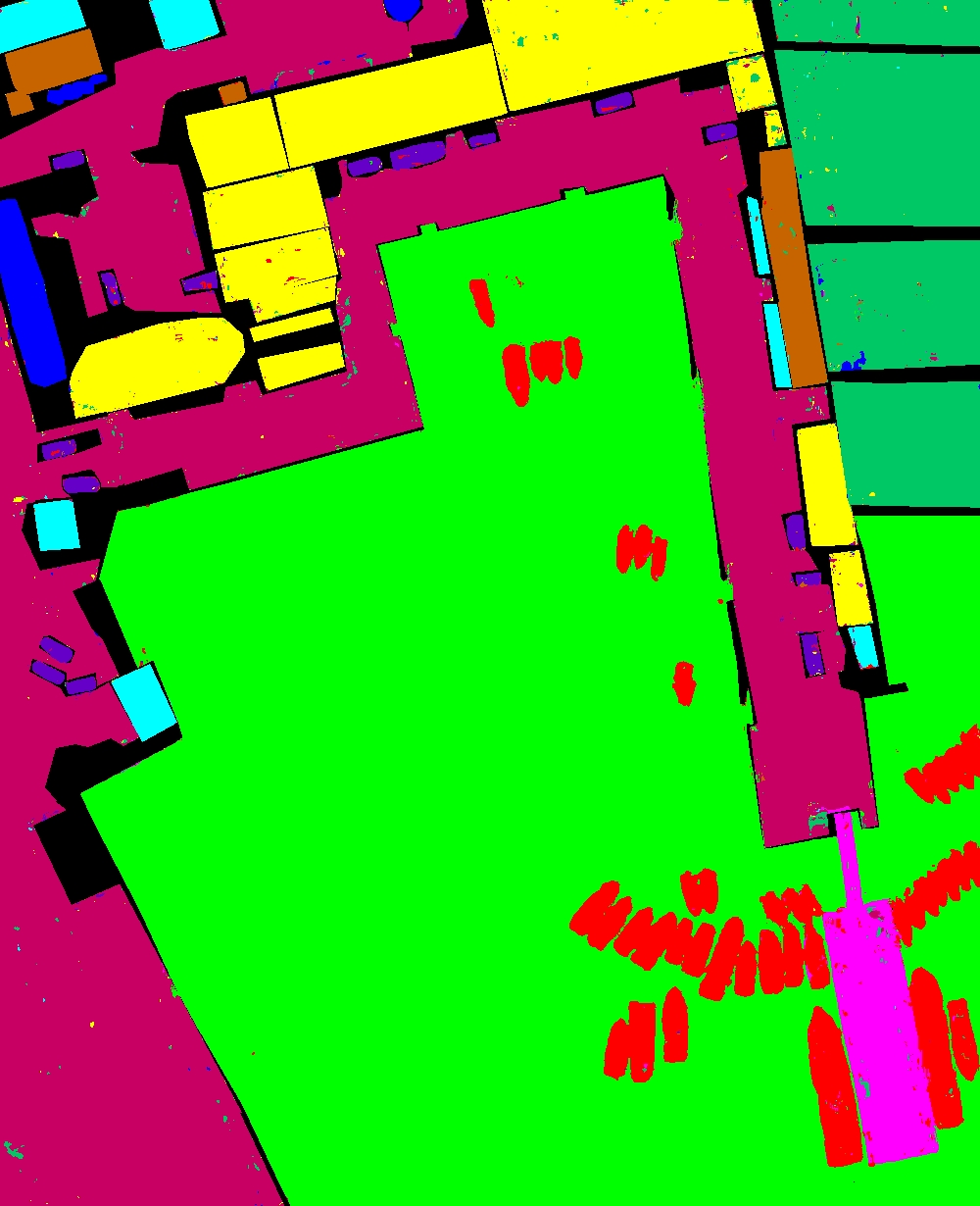}
		\caption{ResNet50} 
	\end{subfigure}
    \begin{subfigure}{0.12\textwidth}
		\includegraphics[width=0.99\textwidth]{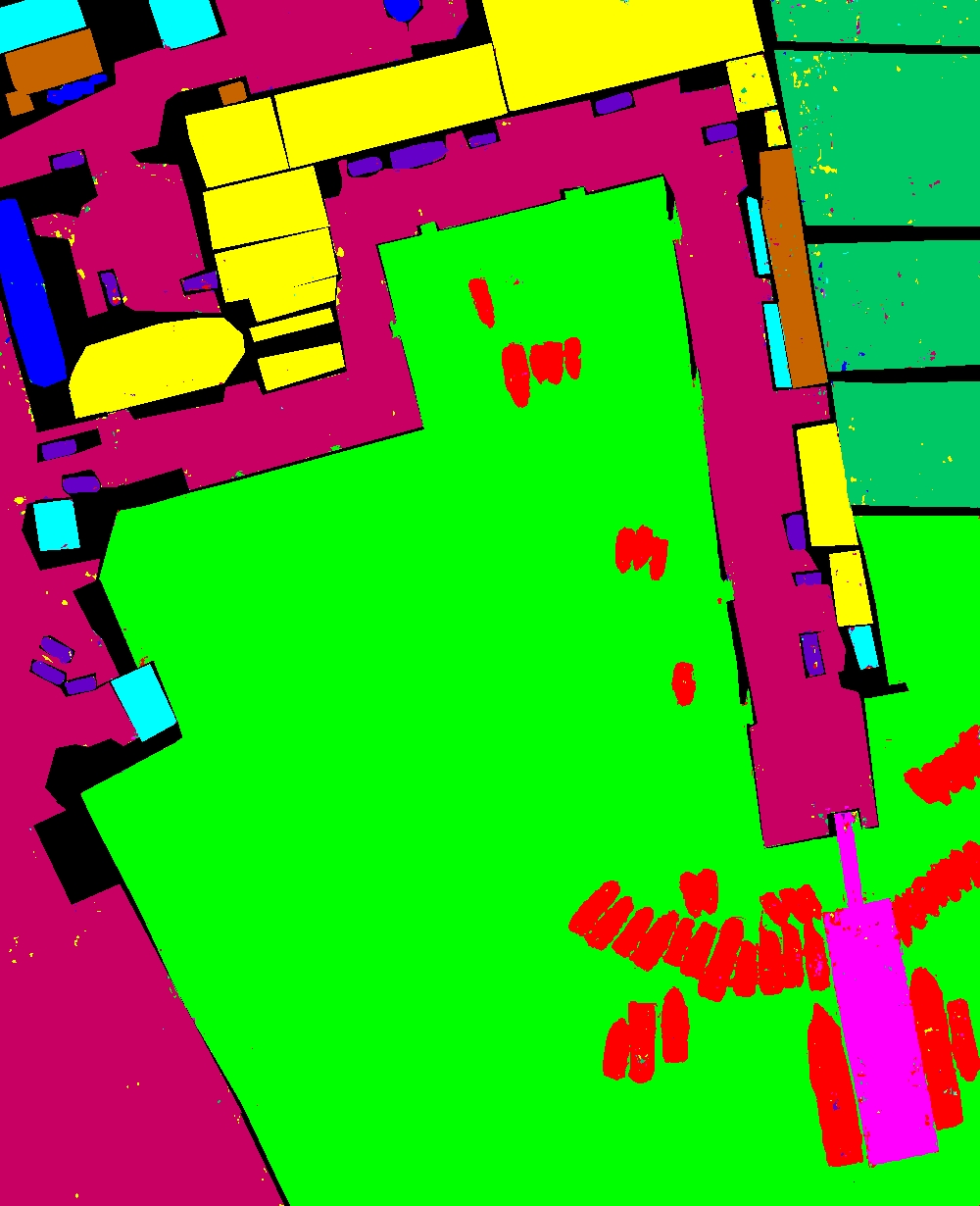}
		\caption{VGG-16}
	\end{subfigure}
    \begin{subfigure}{0.12\textwidth}
		\includegraphics[width=0.99\textwidth]{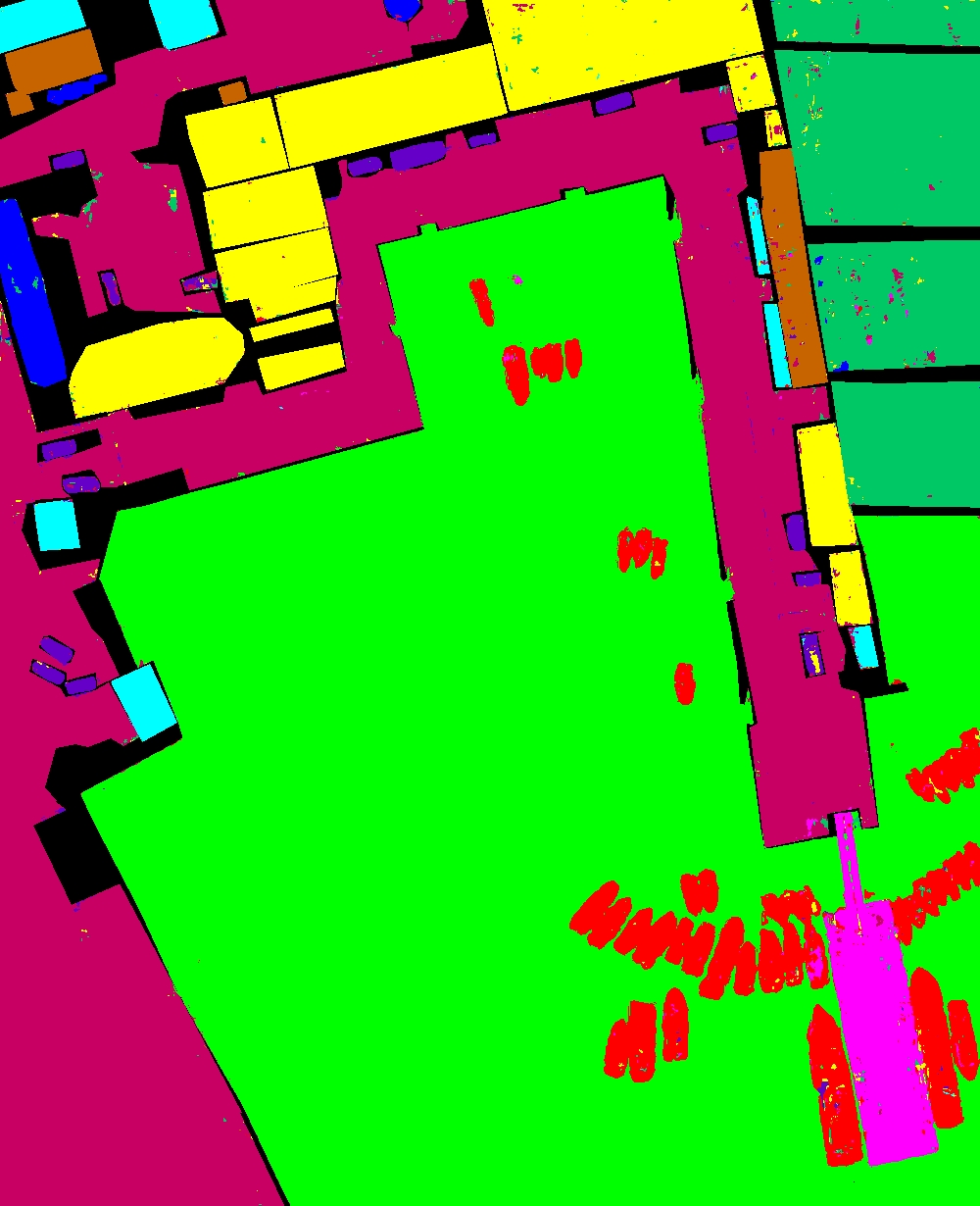}
		\caption{EfficientNet}
	\end{subfigure}
	\begin{subfigure}{0.12\textwidth}
		\includegraphics[width=0.99\textwidth]{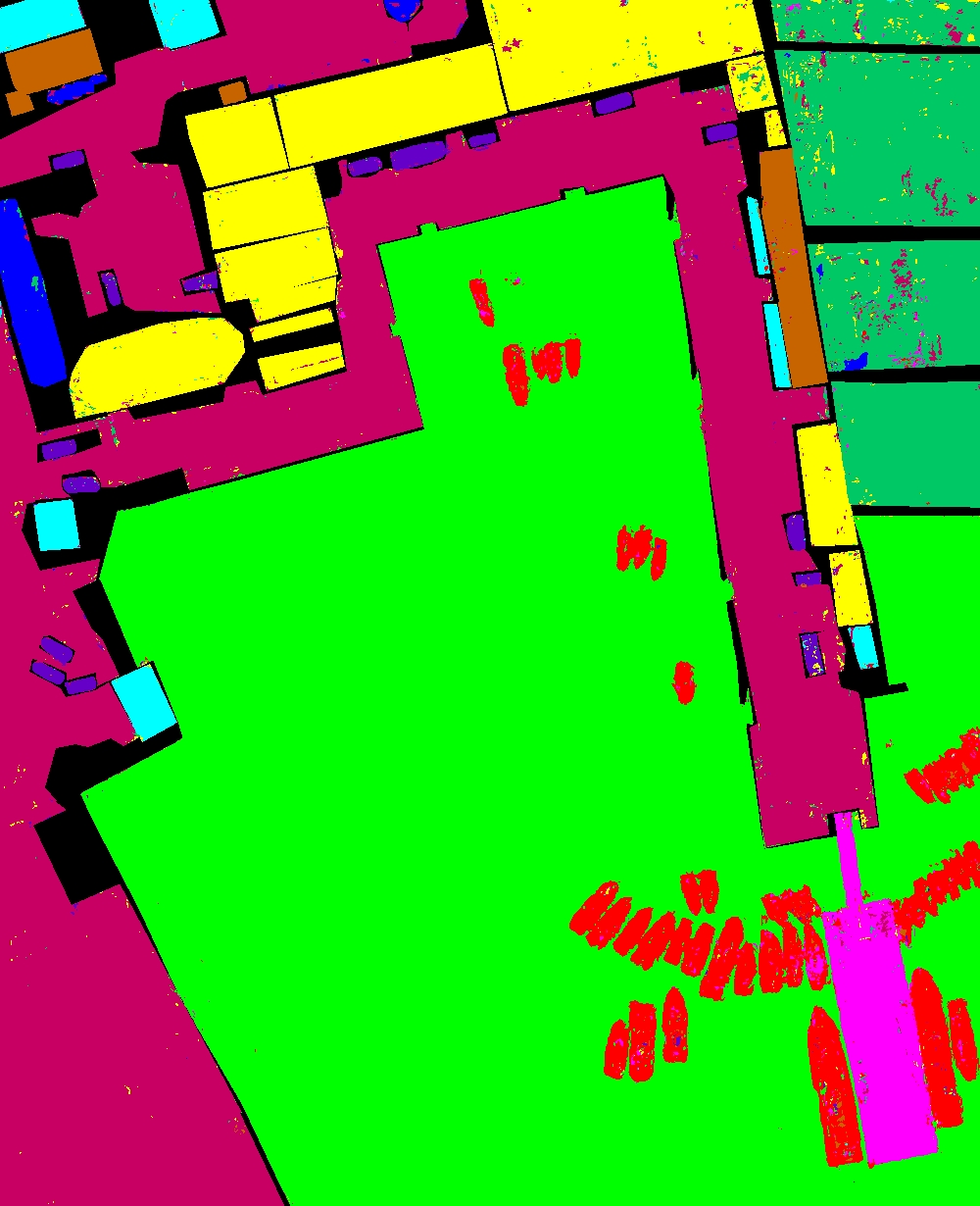}
		\caption{ViT}
	\end{subfigure}
	\begin{subfigure}{0.12\textwidth}
		\includegraphics[width=0.99\textwidth]{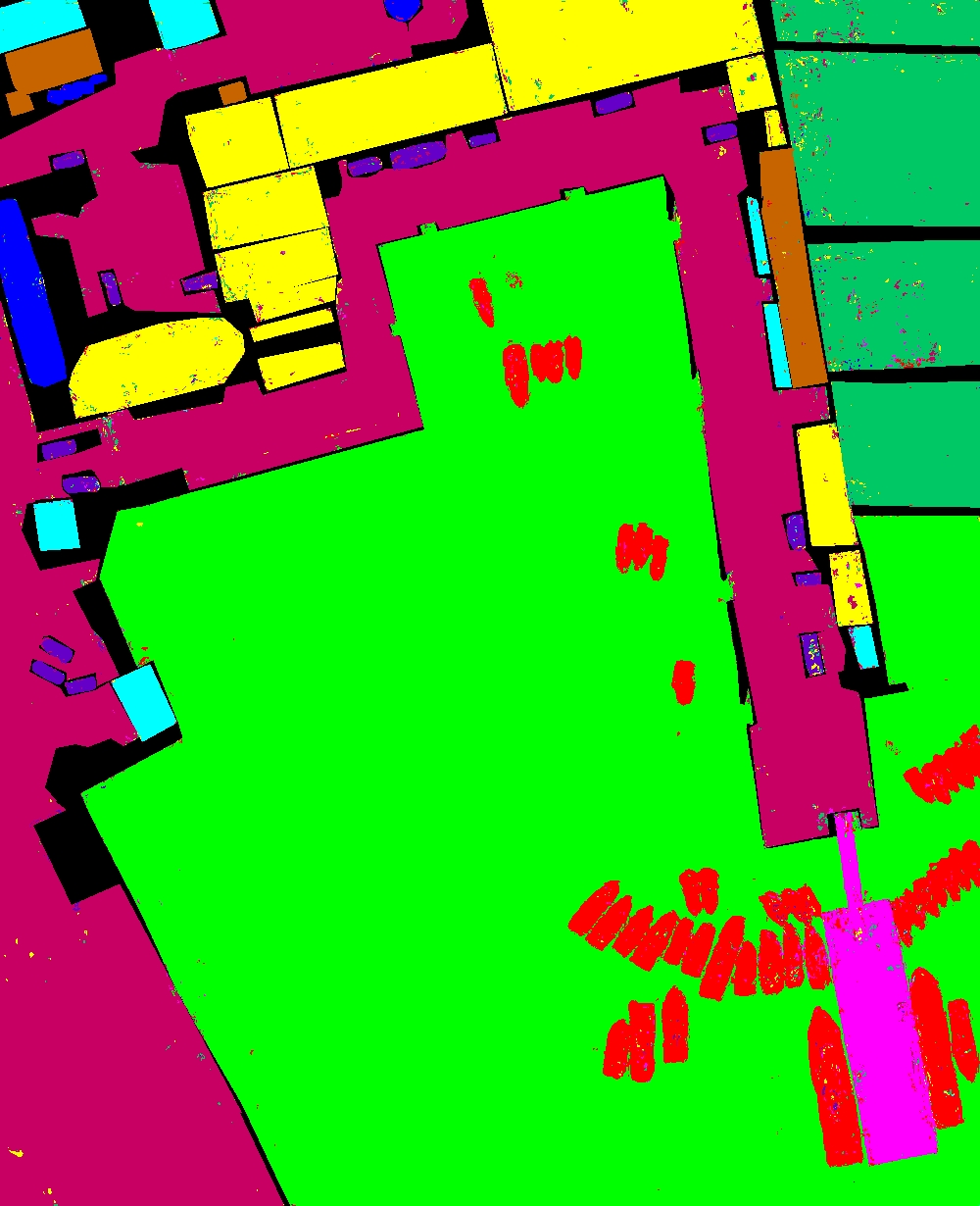}
		\caption{1D-KAN}
	\end{subfigure}
	\begin{subfigure}{0.12\textwidth}
		\includegraphics[width=0.99\textwidth]{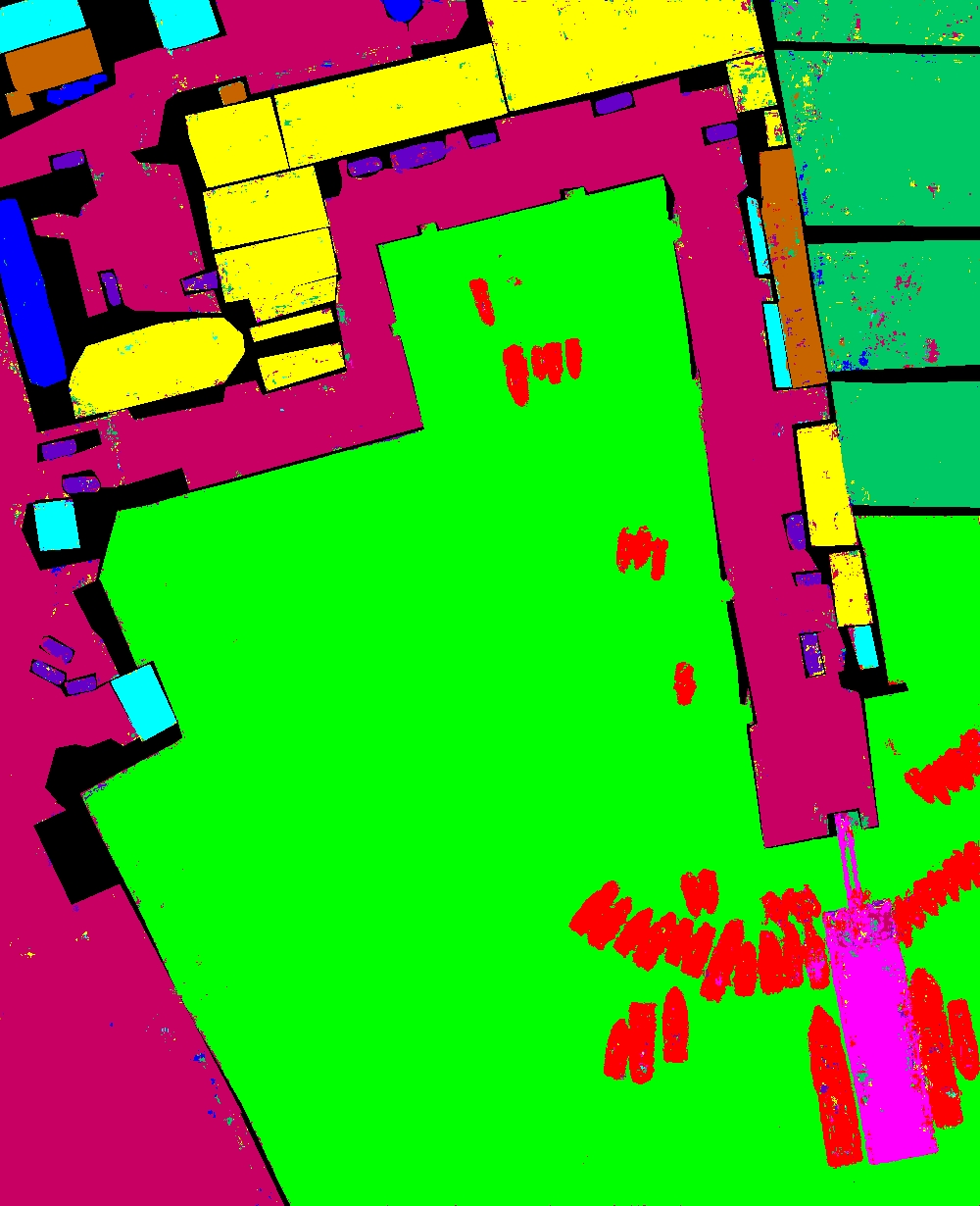}
		\caption{2D-KAN}
	\end{subfigure}
 	\begin{subfigure}{0.12\textwidth}
		\includegraphics[width=0.99\textwidth]{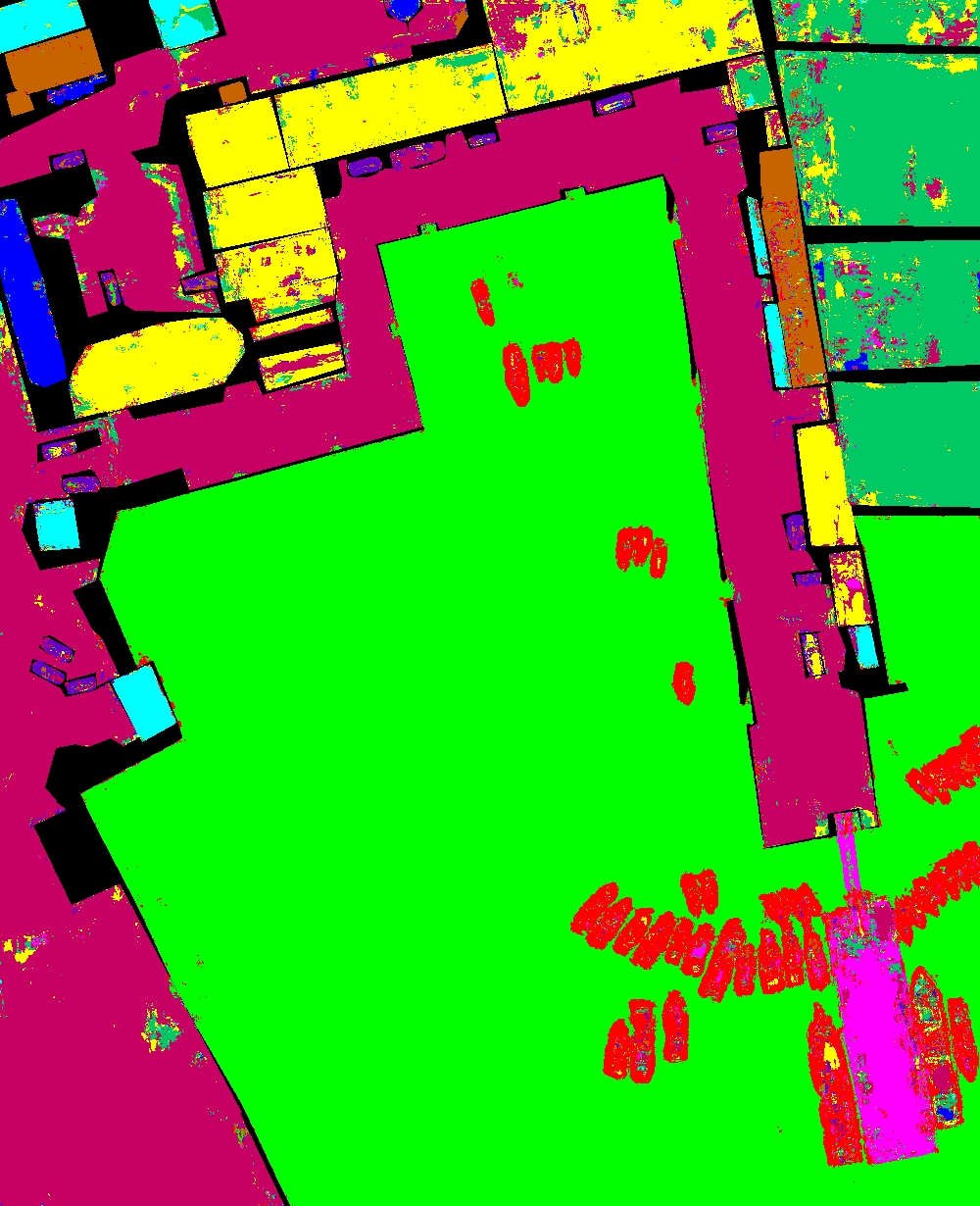}
		\caption{3D-KAN}
	\end{subfigure}
    \begin{subfigure}{0.12\textwidth}
		\includegraphics[width=0.99\textwidth]{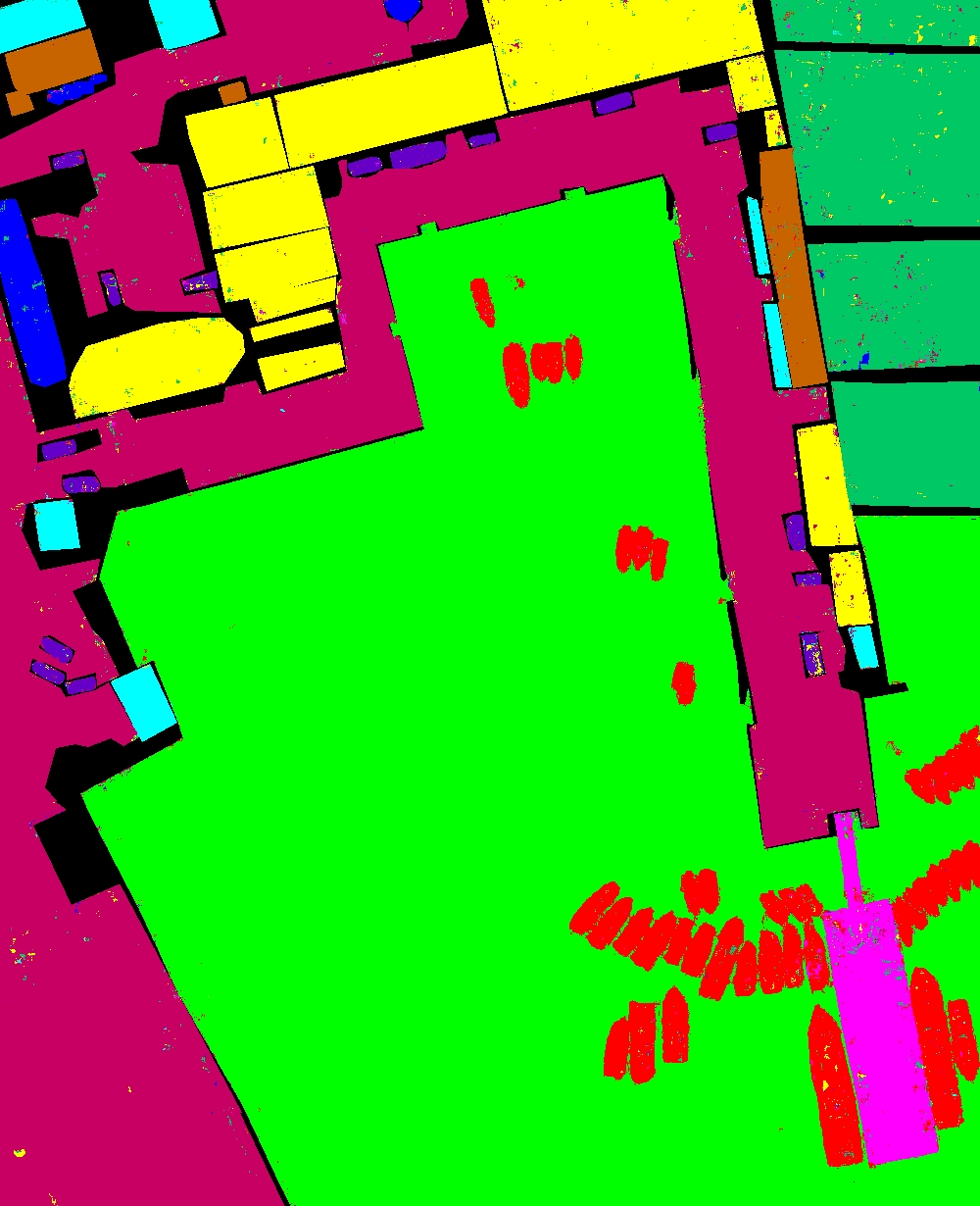}
		\caption{HybridKAN}
	\end{subfigure}
\caption{The predicted land cover maps were created for the Pingan data set.}
\label{fig:Pingan}
\end{figure*}

On the other hand, in the Pingan HSI data set, as seen in Table\ref{tab:Pingan} and Fig\ref{fig:Pingan}, the highest HSI data classification result was obtained by the VGG-16 CNN model with overall accuracy, kappa value, and average accuracy of 99.06\%, 98.61\%, and 98.09\%, respectively. In this HSI dataset, the developed HybridKAN architecture demonstrated a competitive HSI classification accuracy compared to other models with an average accuracy, kappa value, and overall accuracy of 95.95\%, 97.74\%, and 98.48\%, respectively. Similar to the Tangdaowan dataset, the 3D KAN model with an average accuracy of 81.53\% illustrated the least classification accuracy over the 2DKAN (94.14\%). 1DKAN (95.42\%), and the HybridKAN (95.95\%). While the statistical results showed a slightly better classification accuracy by the VGG-16 over the HybridKAN, as seen in  Fig\ref{fig:Pingan}, the HybridKAN architecture produced much less noise and a more homogeneous classification map.

\begin{figure*}[!t]
\centering
	\begin{subfigure}{0.12\textwidth}
		\includegraphics[width=0.99\textwidth]{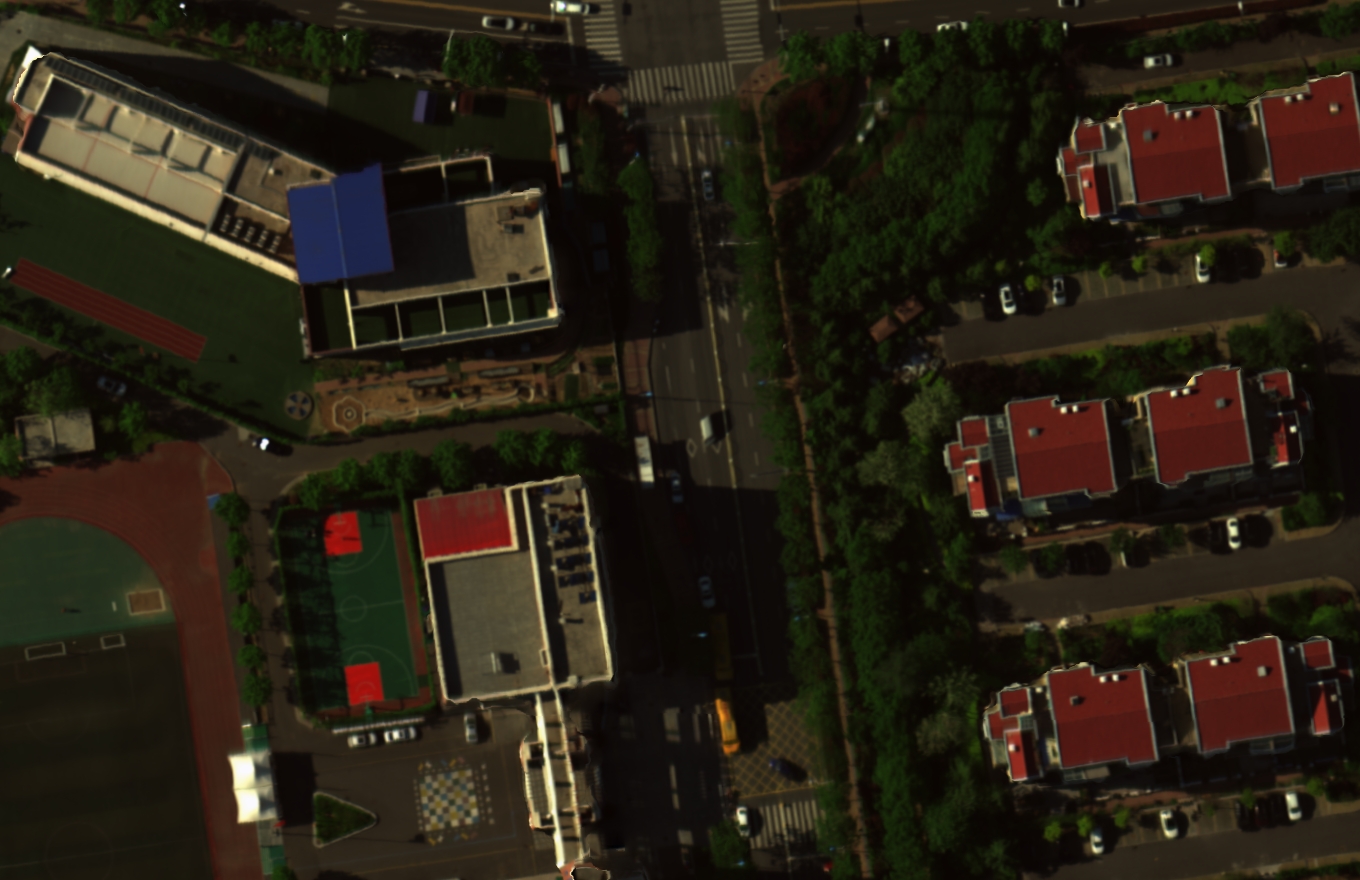}
		\caption{RGB}
	\end{subfigure}
         \begin{subfigure}{0.12\textwidth}
		\includegraphics[width=0.99\textwidth]{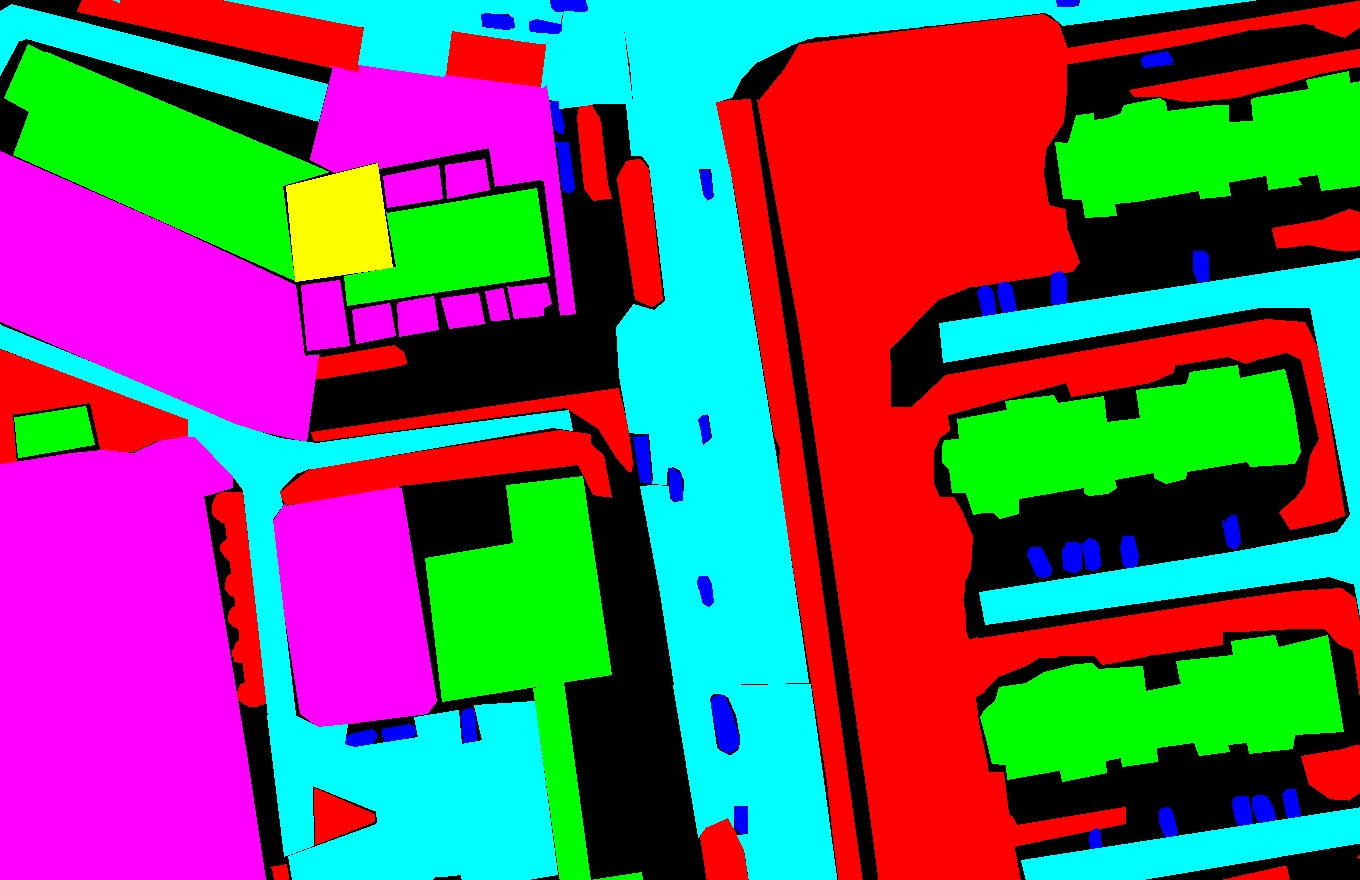}
		\caption{Ground Truth}
	\end{subfigure}
	\begin{subfigure}{0.12\textwidth}
		\includegraphics[width=0.99\textwidth]{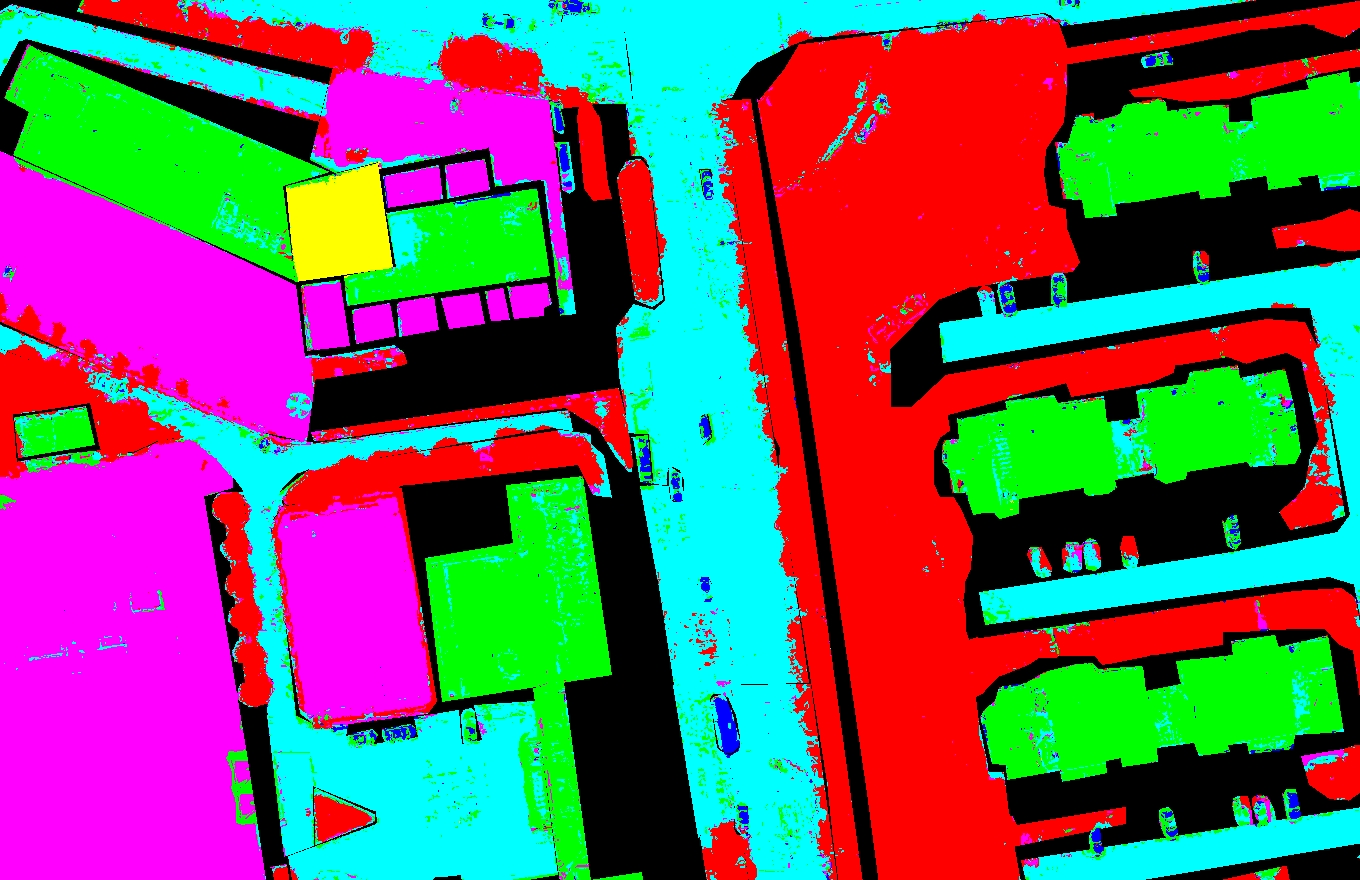}
		\caption{1D-CNN}
	\end{subfigure}
	\begin{subfigure}{0.12\textwidth}
		\includegraphics[width=0.99\textwidth]{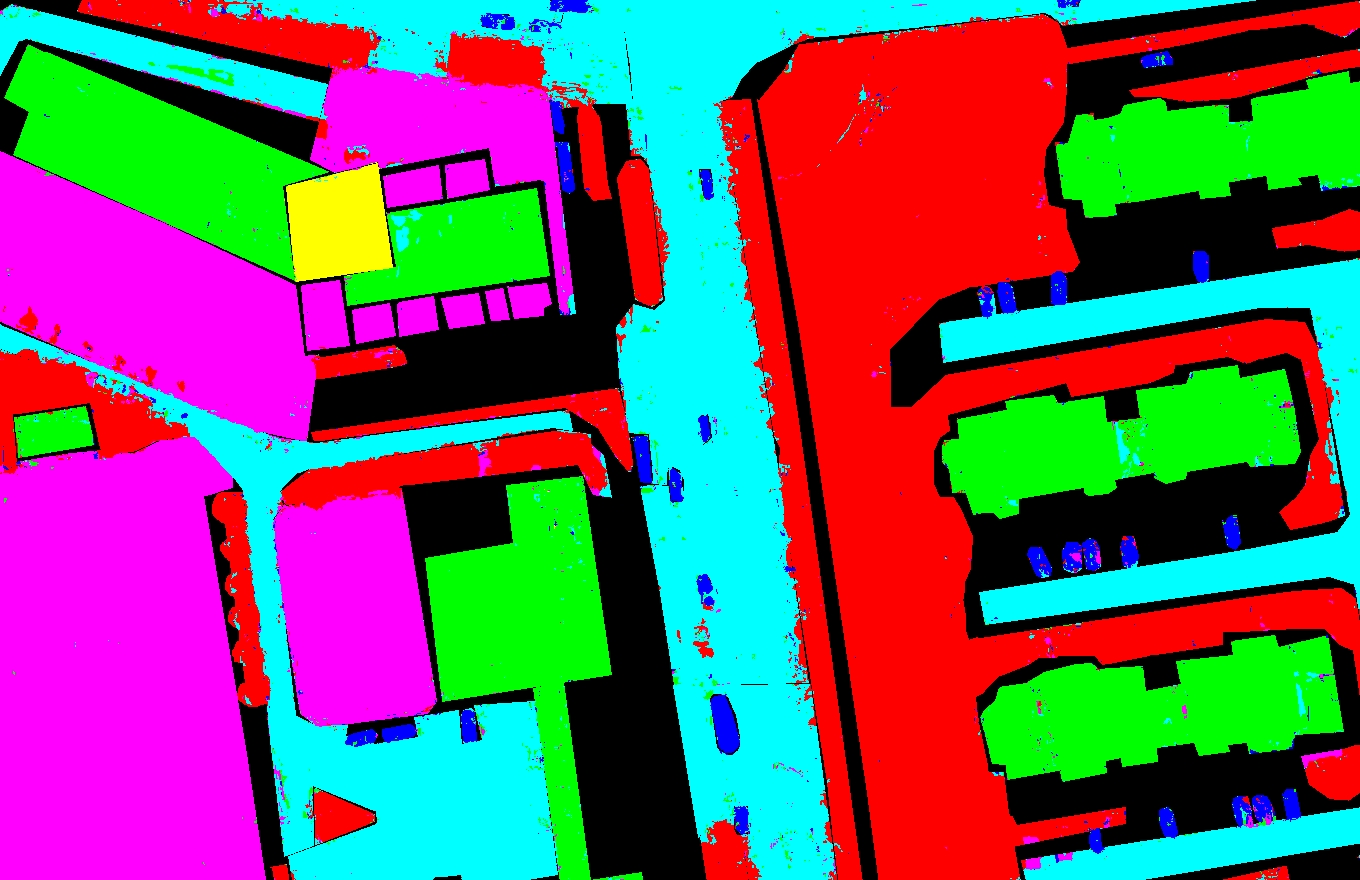}
		\caption{2D-CNN}
	\end{subfigure}
 	\begin{subfigure}{0.12\textwidth}
		\includegraphics[width=0.99\textwidth]{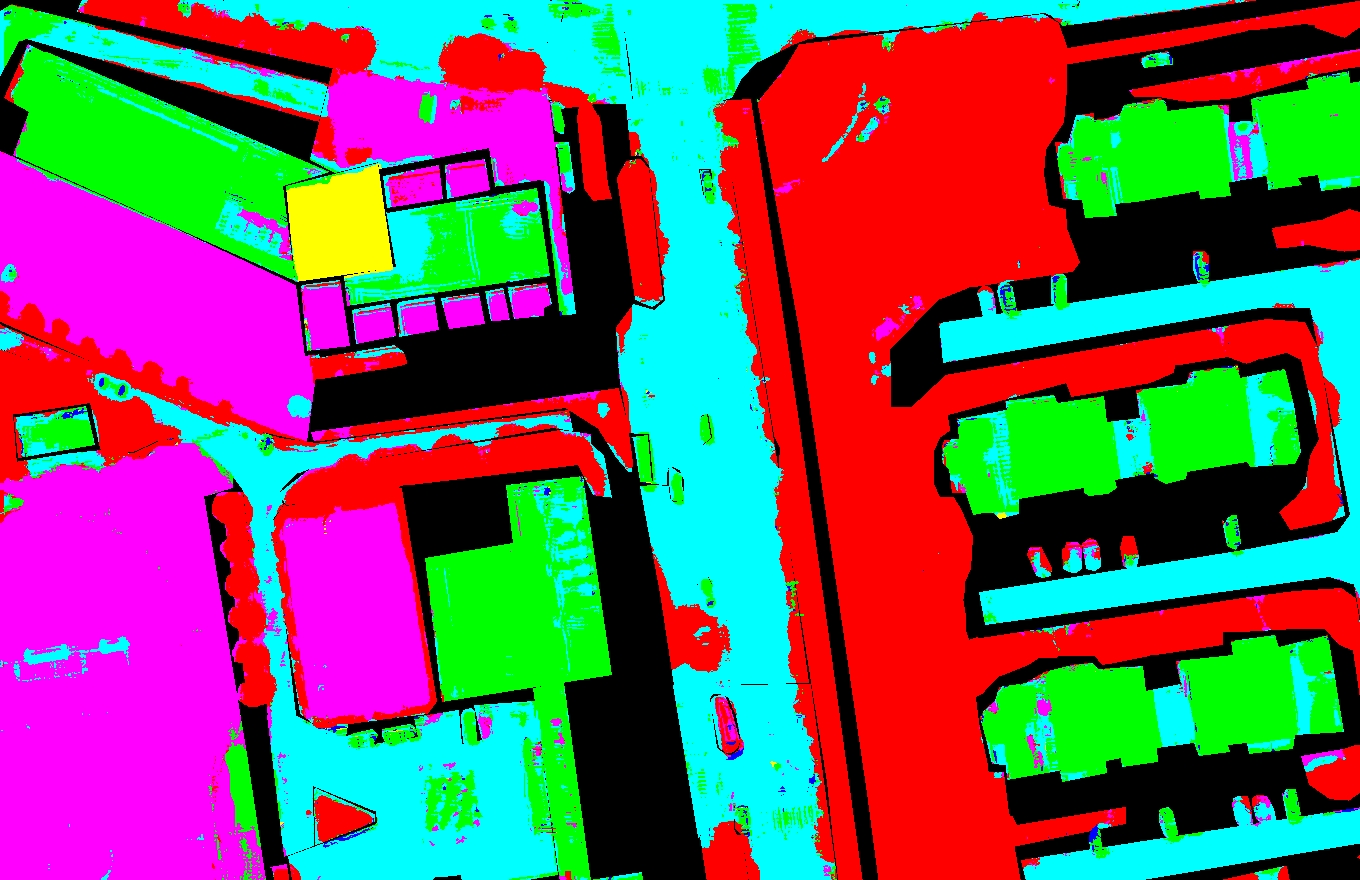}
		\caption{3D-CNN}
	\end{subfigure}
	\begin{subfigure}{0.12\textwidth}
		\includegraphics[width=0.99\textwidth]{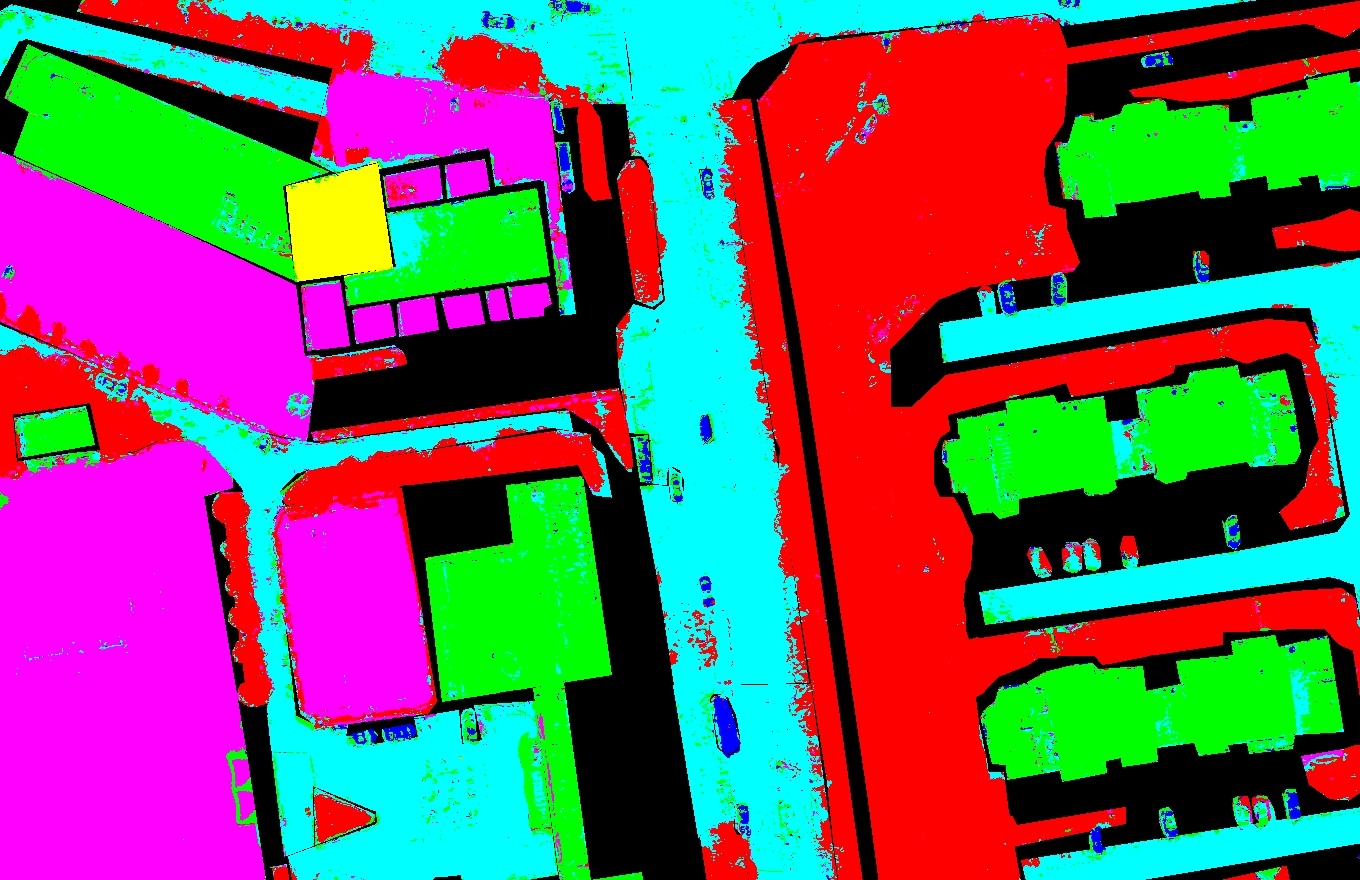}
		\caption{DRNN} 
	\end{subfigure}
	\begin{subfigure}{0.12\textwidth}
		\includegraphics[width=0.99\textwidth]{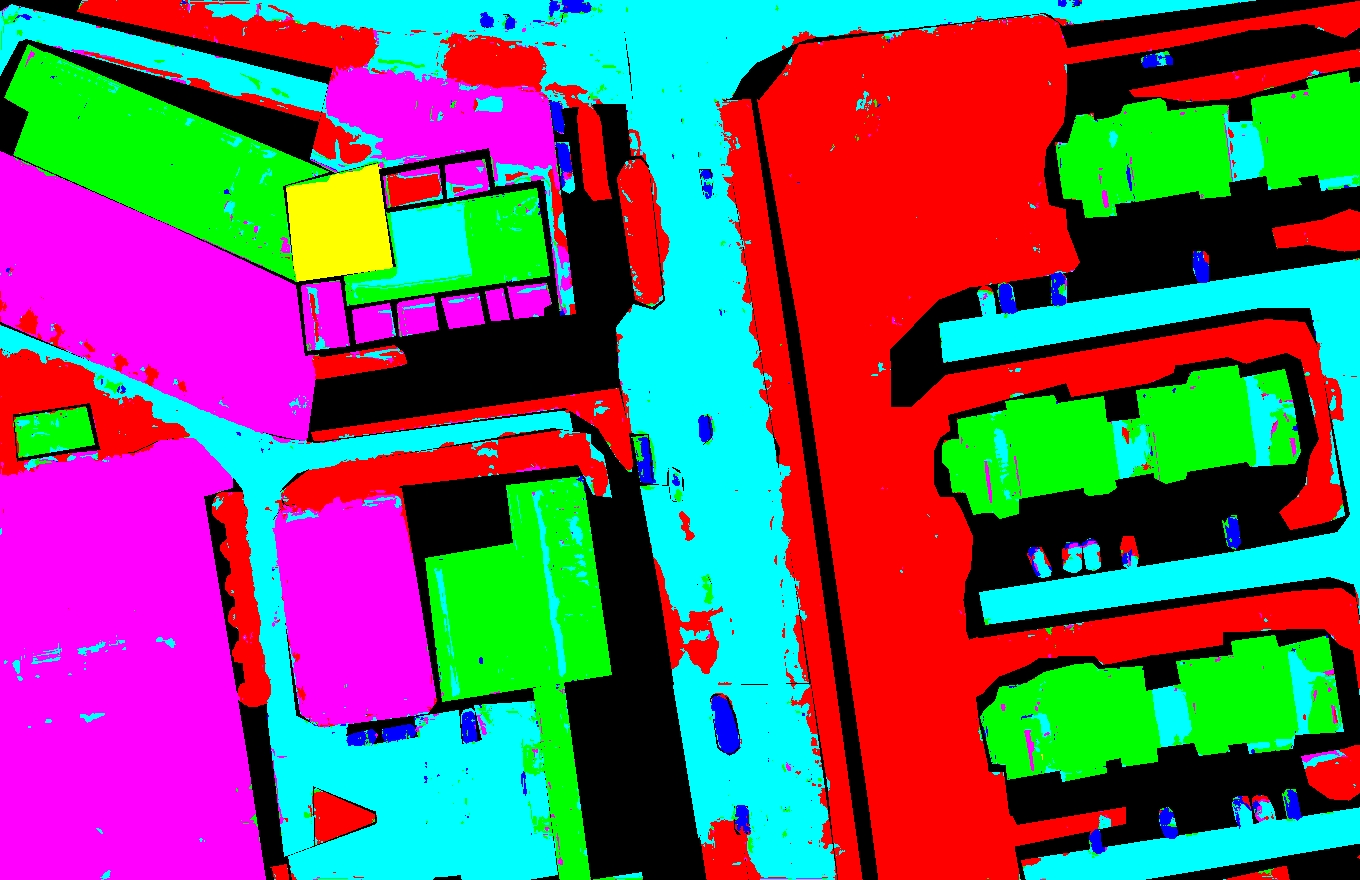}
		\caption{ResNet50} 
	\end{subfigure}
    \begin{subfigure}{0.12\textwidth}
		\includegraphics[width=0.99\textwidth]{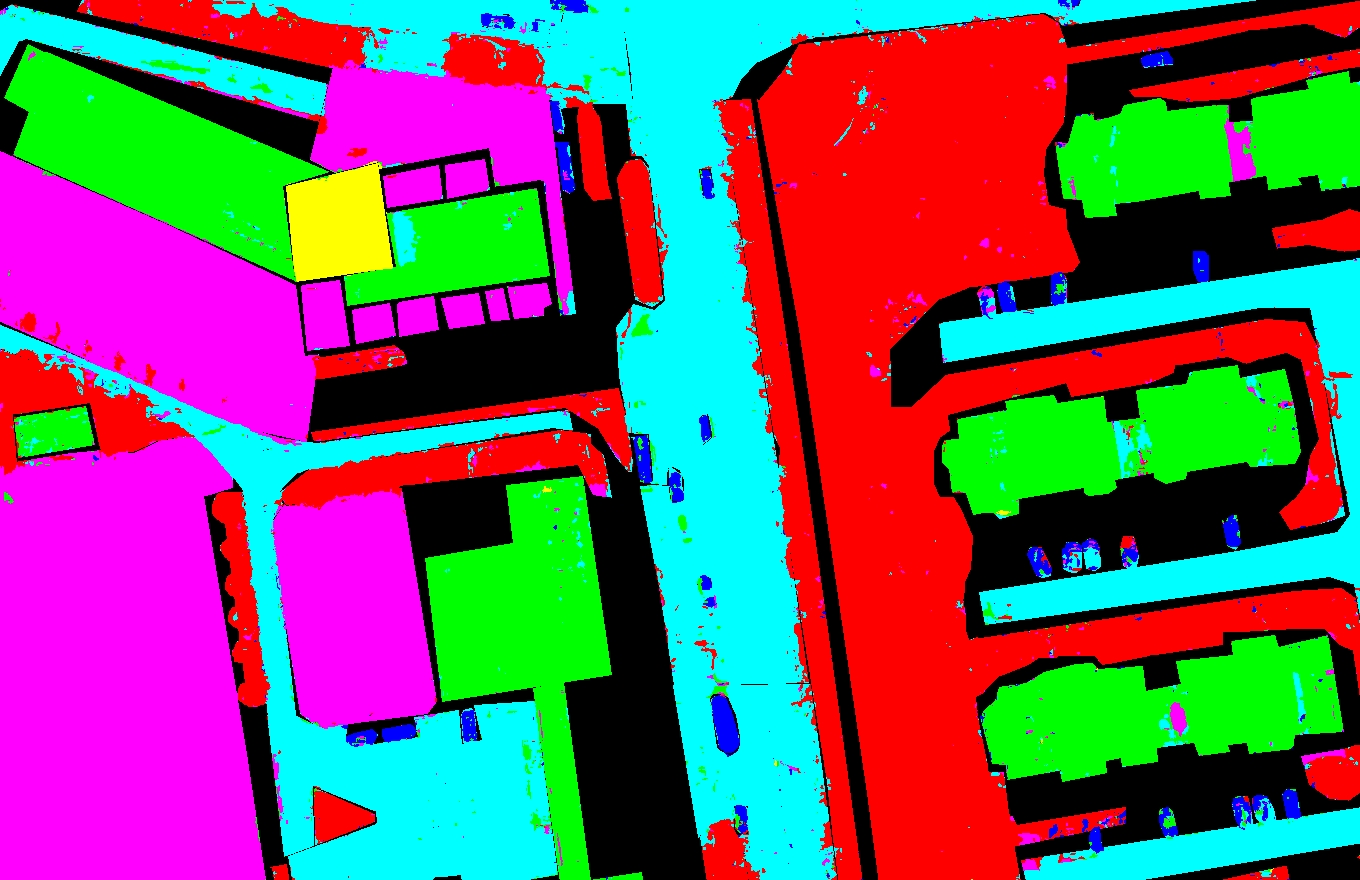}
		\caption{VGG-16}
	\end{subfigure}
    \begin{subfigure}{0.12\textwidth}
		\includegraphics[width=0.99\textwidth]{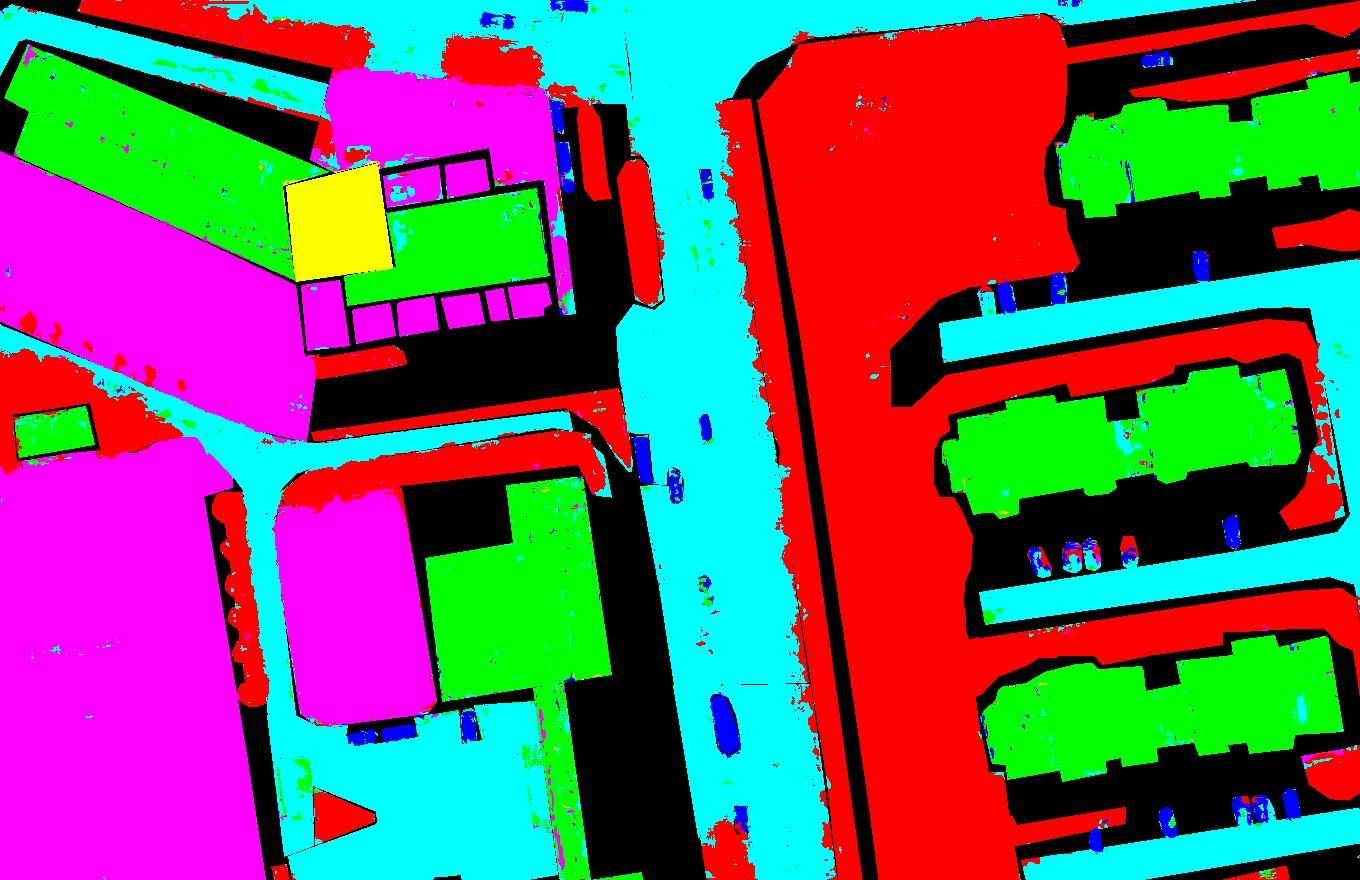}
		\caption{EfficientNet}
	\end{subfigure}
	\begin{subfigure}{0.12\textwidth}
		\includegraphics[width=0.99\textwidth]{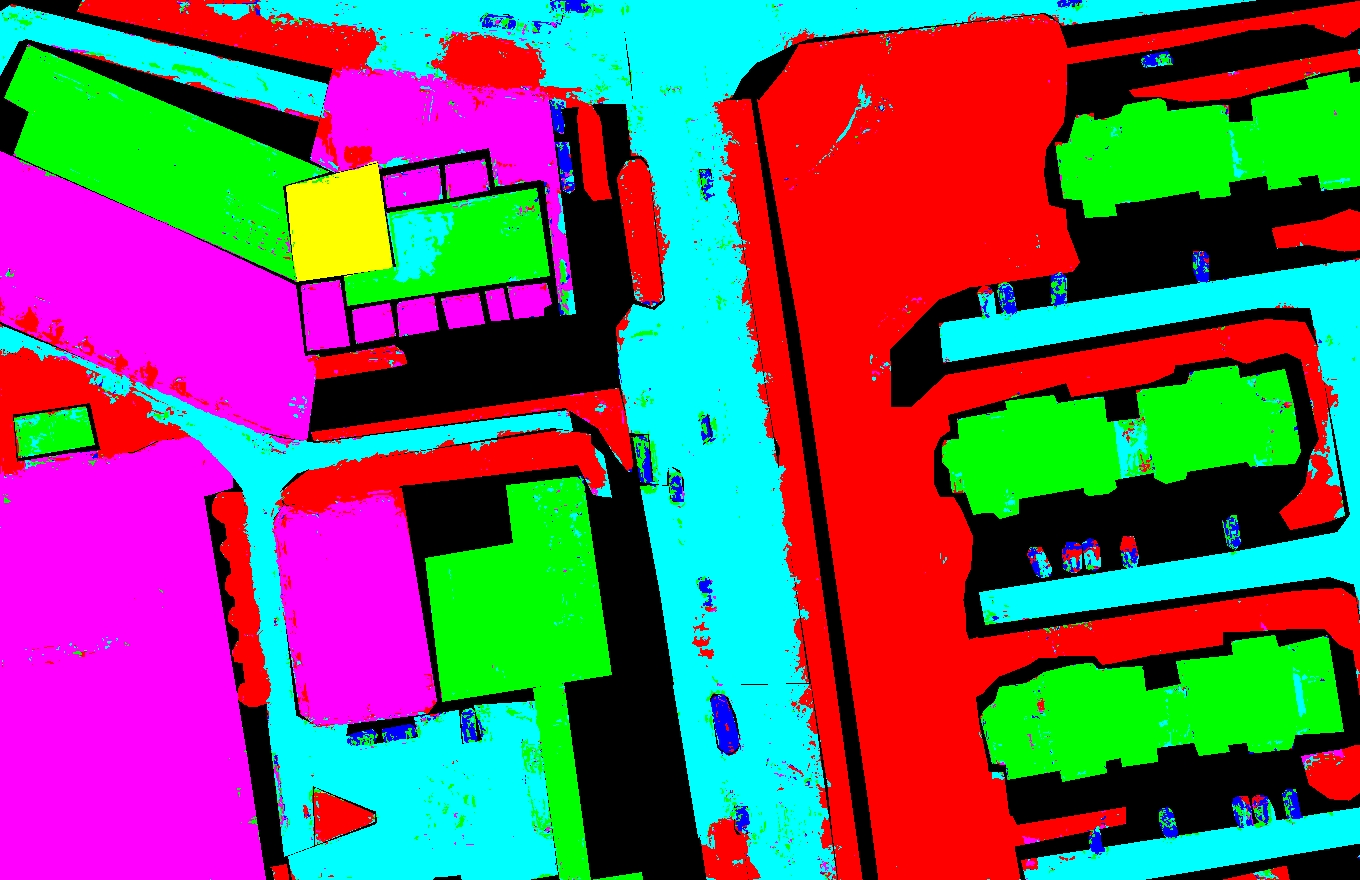}
		\caption{ViT}
	\end{subfigure}
	\begin{subfigure}{0.12\textwidth}
		\includegraphics[width=0.99\textwidth]{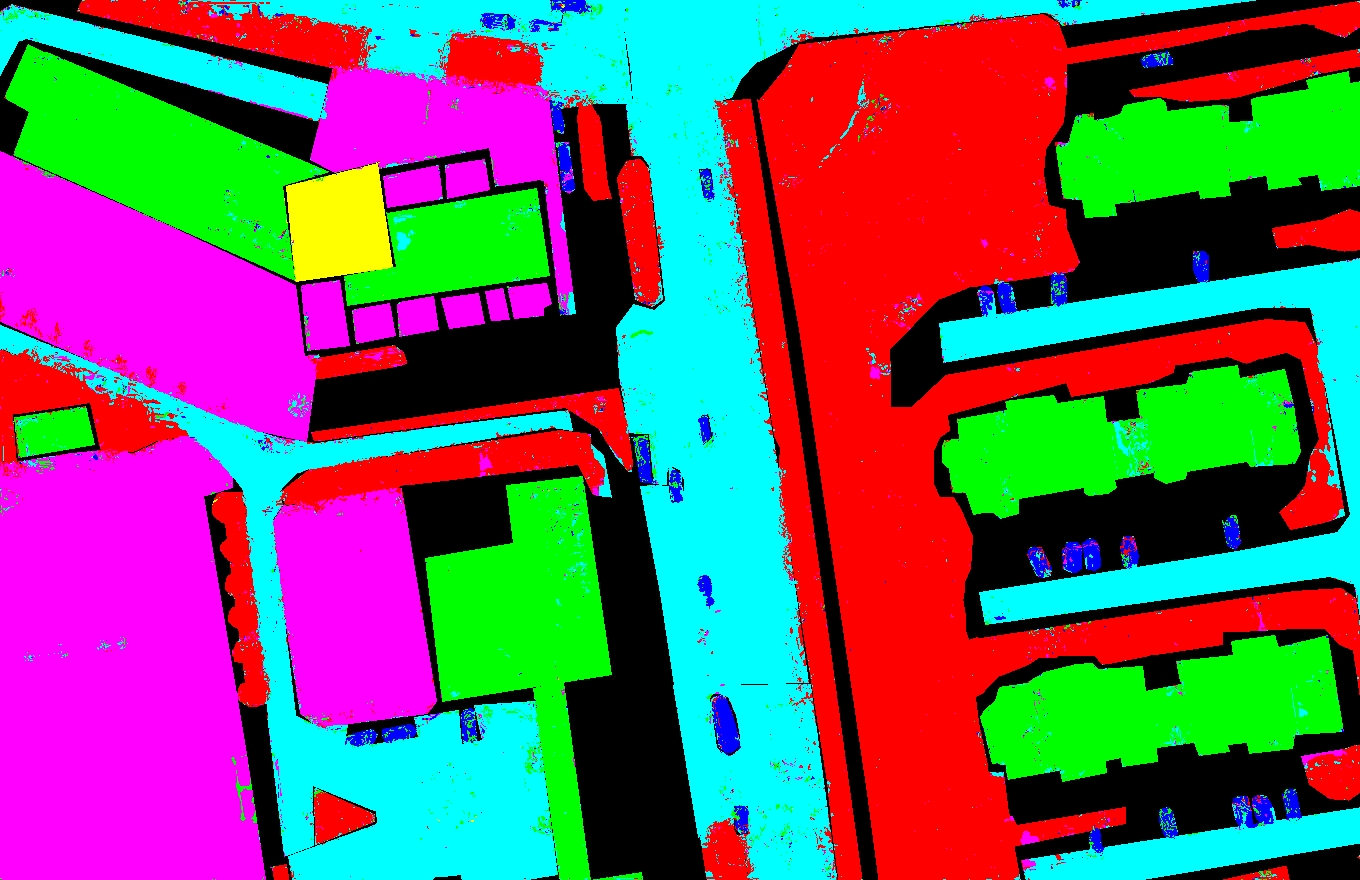}
		\caption{1D-KAN}
	\end{subfigure}
	\begin{subfigure}{0.12\textwidth}
		\includegraphics[width=0.99\textwidth]{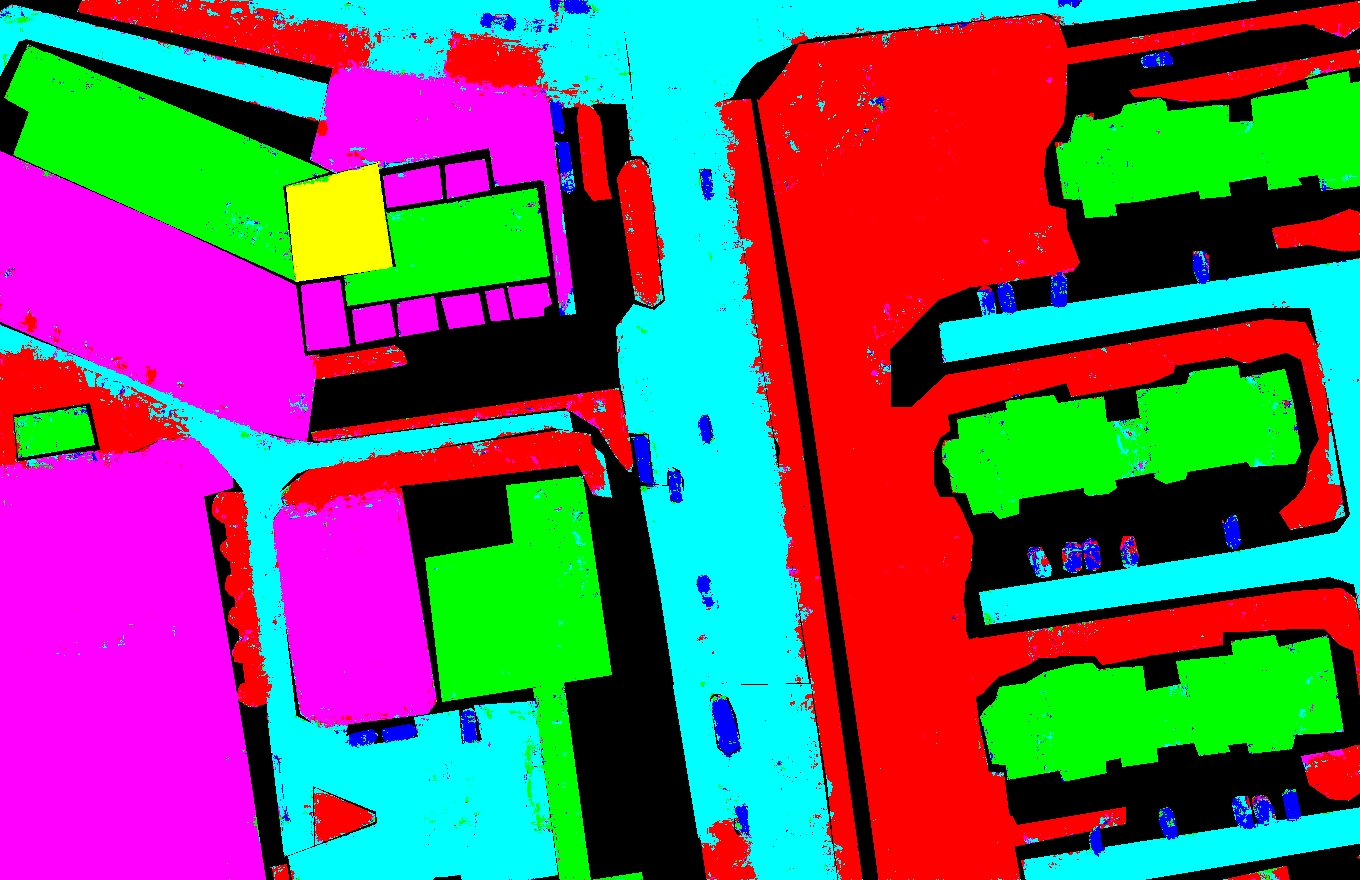}
		\caption{2D-KAN}
	\end{subfigure}
 	\begin{subfigure}{0.12\textwidth}
		\includegraphics[width=0.99\textwidth]{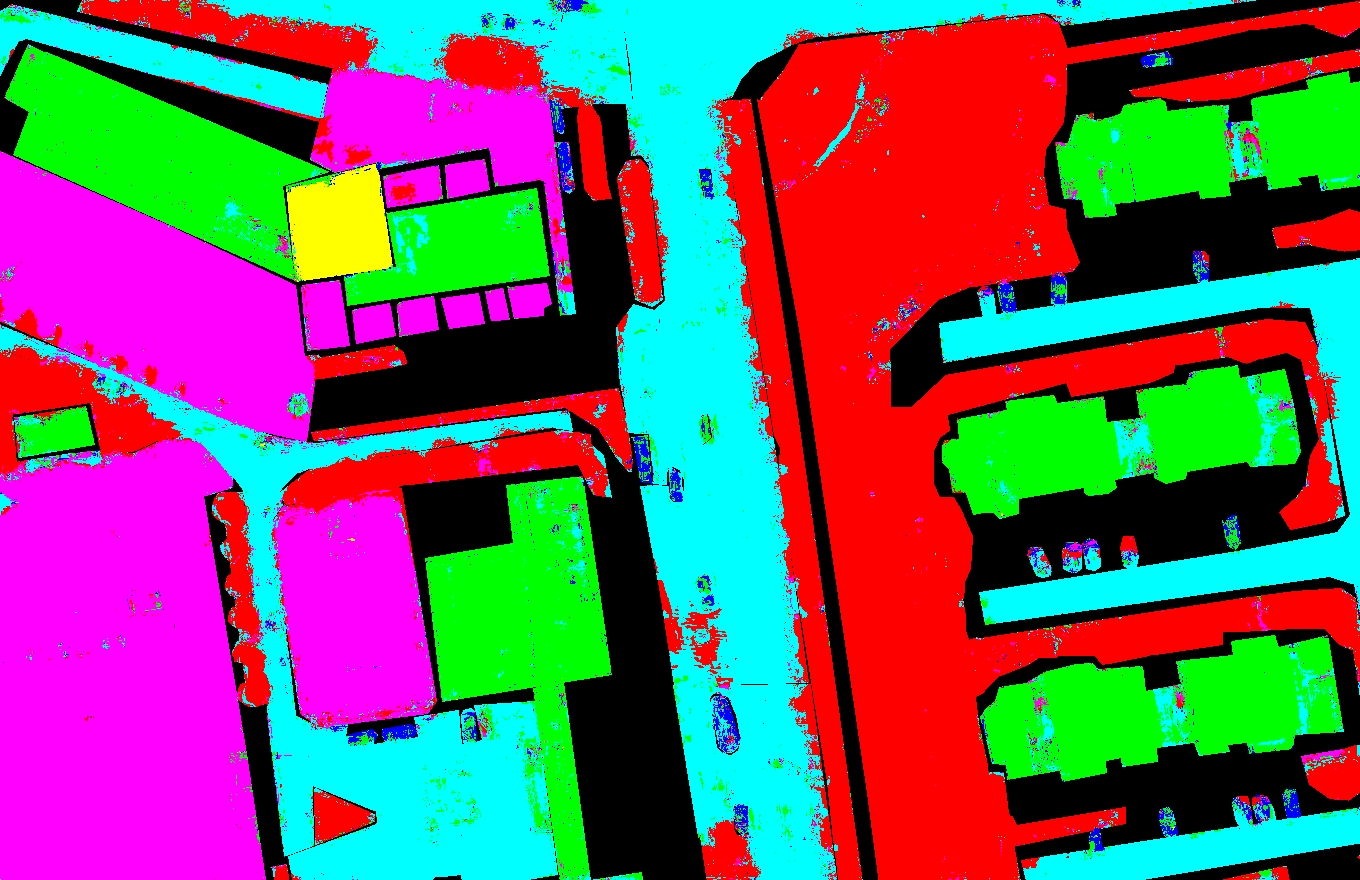}
		\caption{3D-KAN}
	\end{subfigure}
    \begin{subfigure}{0.12\textwidth}
		\includegraphics[width=0.99\textwidth]{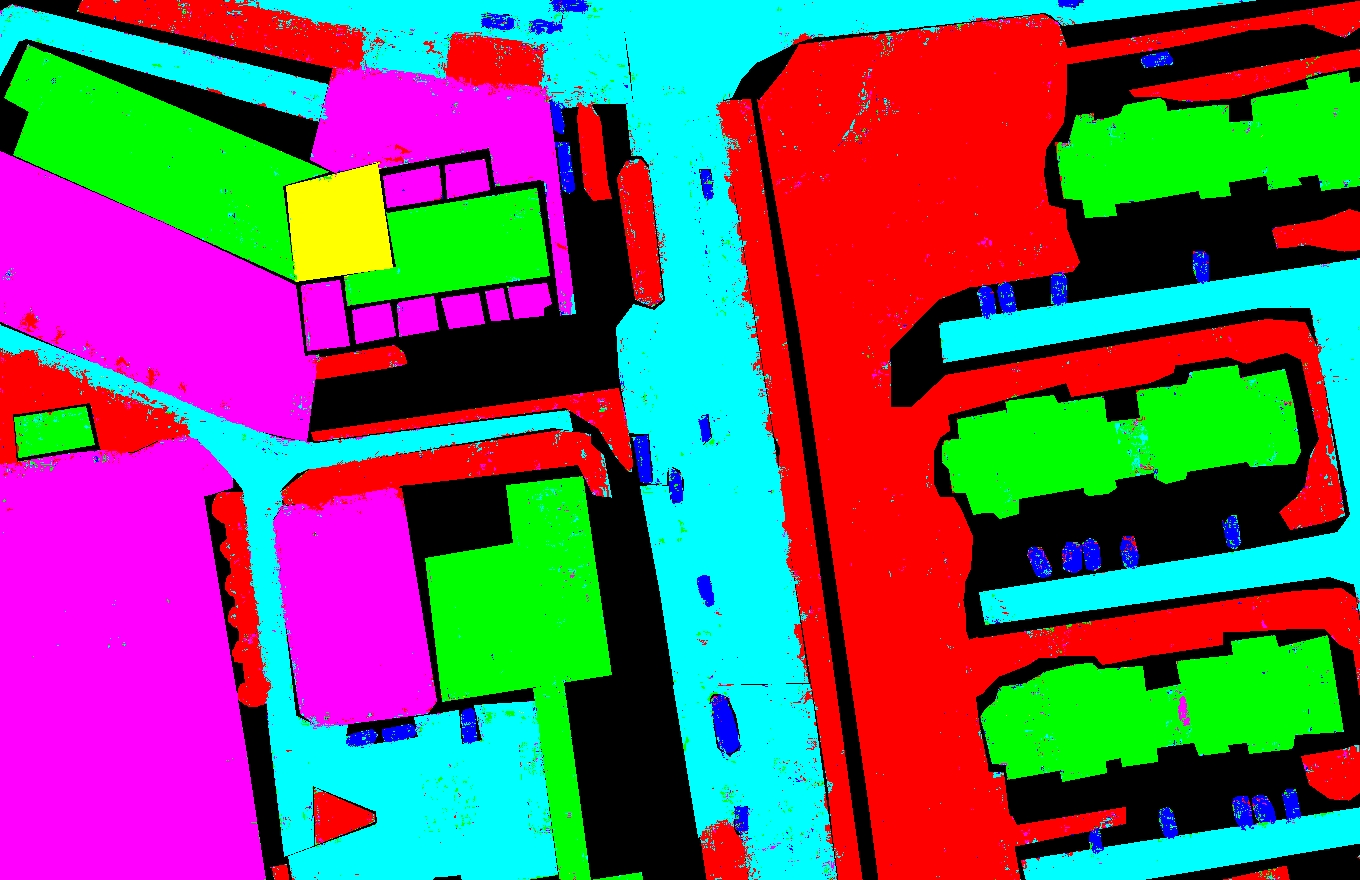}
		\caption{HybridKAN}
	\end{subfigure}
\caption{The predicted land cover maps were created for the Qingyun data set.}
\label{fig:Qingyun}
\end{figure*}

Moreover, as seen in Table\ref{tab:Qingyun} and Fig\ref{fig:Qingyun}, the best HSI data classification accuracy was obtained by the developed HybridKAN architecture in terms of overall accuracy (97.06\%) and kappa value (96.11\%) in the Qingyun HSI dataset. The highest average accuracy was achieved by the 2DCNN model (95.60\%) over the other developed classification models. The proposed HybridKAN architecture obtained the highest average accuracy (94.91\%) as compared to 3DKAN (84.69\%), 1DKAN (92.11\%), and 2DKAN network (92.93\%). Overall, the obtained results showed significant capability of KAN models for complex land cover land use mapping using HSI data. We used a simple and straightforward architecture similar to traditional CNN-based models (e.g., Hybrid SN \cite{Roy6016}), yet the developed model based on the KAN architectures illustrated competitive or better HSI data classification capability compared to other developed CNN- and ViT-based classification models.

\begin{table*}[!t]
\centering
\caption{Classification results in terms of OA, AA, and Kappa (in \%) obtained on the Qingyun dataset.}
\resizebox{0.99\linewidth}{!}{
\begin{tabular}{|c|c|c|c|c|c|c|c|c|c|c|c|c|c|c|}
\hline
Class No. & 1DCNN & 2DCNN & 3DCNN& VGG16\cite{simonyan2015deep} & ResNet50\cite{He_2016_CVPR} & EfficientNet\cite{Koonce2021} & RNN\cite{Mou752} & ViT\cite{Alexey20} & 1DKAN \cite{liu2024kan} &2DKAN  & 3DKAN  & HybridKAN \\ 
\hline
\hline
1& 94.62& 97.53&   90.96   & 96.82& 97.28& \textbf{97.64} & 95.36& 97.42&  96.45 & 96.50      & 93.72 &  97.17 \\

2 & 92.28 & \textbf{97.89}&   84.23   & 96.01 & 87.62& 96.04& 92.88& 95.02&  96.73 & \textbf{97.89}    &   92.94&   97.66\\

3 & 31.82& \textbf{85.48}&  10.11   &  71.31& 55.82& 69.04& 38.55& 53.76& 73.01  & 73.07&    47.39&  82.62\\

4 & 97.03& 99.24&   95.66   & 99.52 & 97.62& \textbf{99.83}& 98.54& 98.15&  99.10 & 98.62    &   96.89&  99.07\\

5& 92.71& 98.20 &   91.21   & \textbf{98.54} & 93.09& 96.95& 92.35& 95.82&  97.38 & 97.91&   95.85&  98.05\\

6& 90.02& 95.23&    83.80  & 95.05 & 93.13 & 95.69& 90.63& 93.13  & 95.32  & 95.68  &  90.80 &  \textbf{96.35} \\ 

\hline
\hline
OA& 91.63 & 96.98 &  87.19 & 96.24& 92.80& 96.27 & 92.15& 94.83 &  96.13 & 96.55 &  92.74 &  \textbf{97.06}  \\

AA & 83.08& \textbf{95.60} & 75.44 & 92.87& 87.43& 92.53 & 84.72& 88.88& 92.11  & 92.93 & 84.69 & 94.91\\

 $\kappa (\times 100)$& 88.89& 96.00&   82.94  &  95.03 & 90.43& 95.05& 89.59& 93.15  &  94.87 & 95.43&  90.37 &  \textbf{96.11} \\
\hline
\end{tabular}}
\label{tab:Qingyun}
\end{table*}

\subsection{Convergence graph between HybridSN and its KAN version}

\begin{figure*}[!ht]
\centering
\begin{subfigure}{.8\columnwidth}
\centering
\includegraphics[clip=true, trim = 10 0 10 10, width=0.98\textwidth]{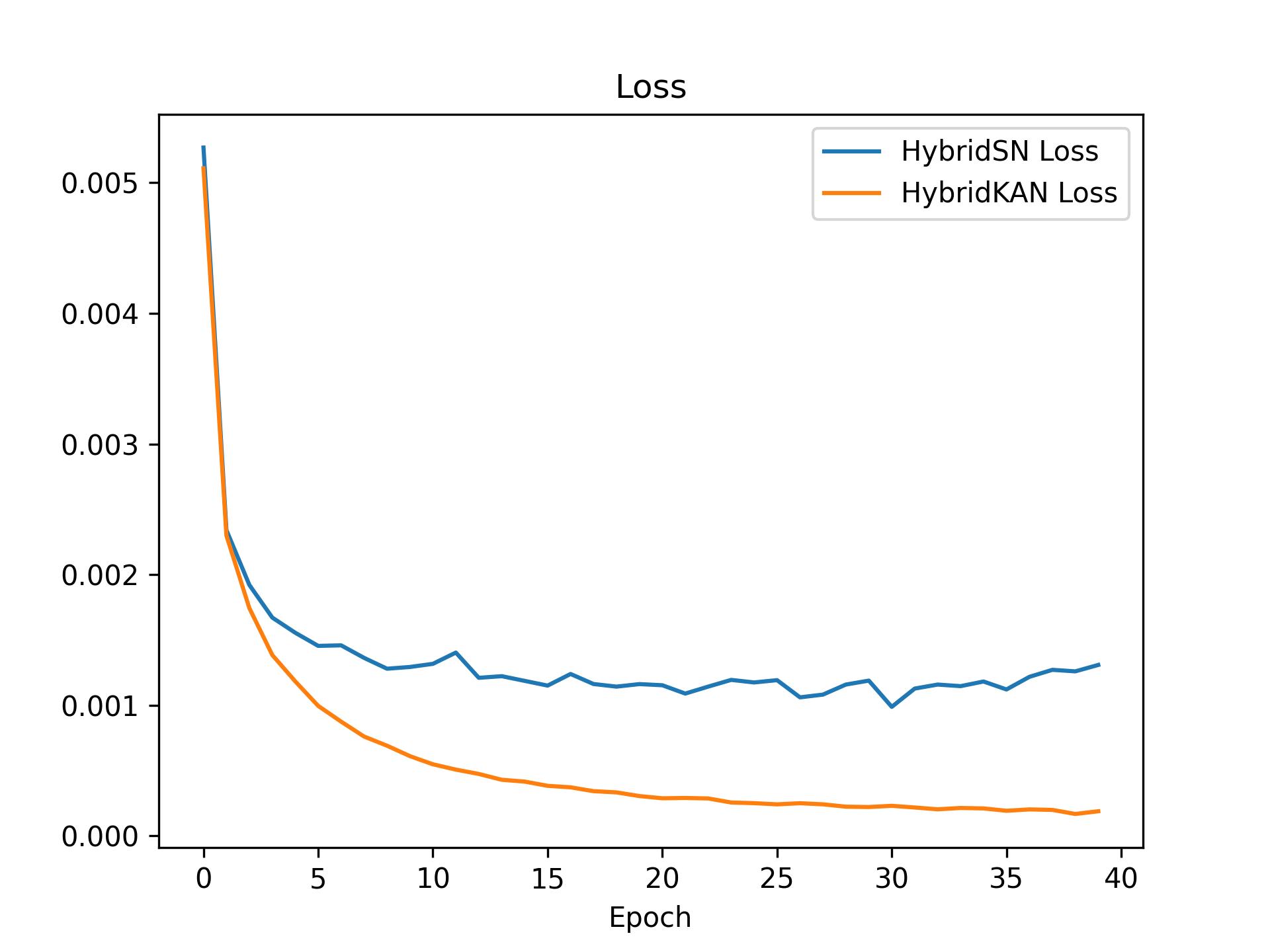} 
\caption{}
\end{subfigure}%
\begin{subfigure}{.8\columnwidth}
\centering
\includegraphics[clip=true, trim = 10 0 10 10, width=0.98\textwidth]{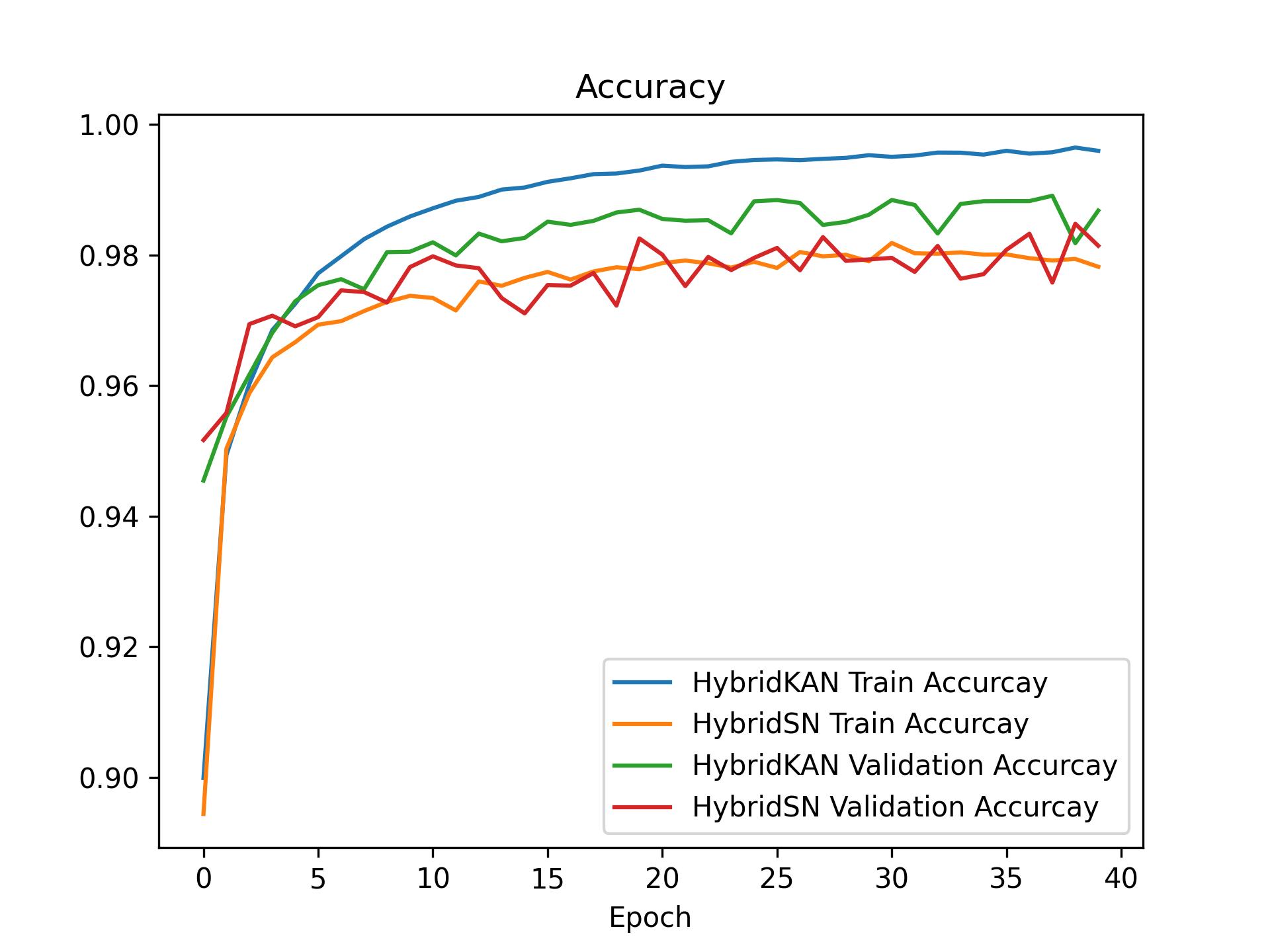} 
\caption{}
\end{subfigure}%
\caption{The convergence graph between HybridSN and its KAN version over the Tangdaowan HSI benchmark dataset for 40 epochs.}
\label{fig:convergenceTangdaowan}
\end{figure*}

\begin{figure*}[!ht]
\centering
\begin{subfigure}{.8\columnwidth}
\centering
\includegraphics[clip=true, trim = 10 0 10 10, width=0.98\textwidth]{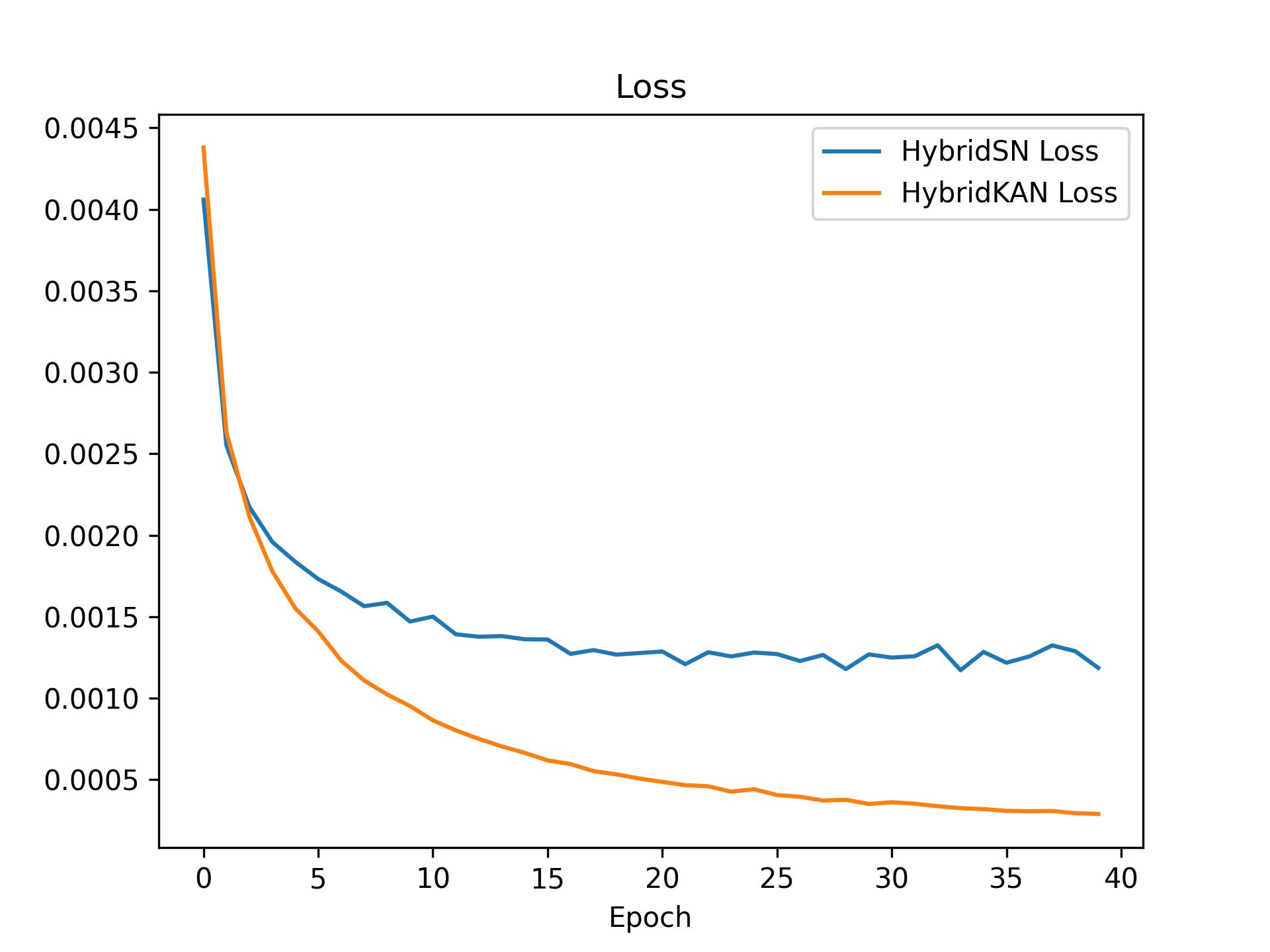} 
\caption{}
\end{subfigure}%
\begin{subfigure}{.8\columnwidth}
\centering
\includegraphics[clip=true, trim = 10 0 10 10, width=0.98\textwidth]{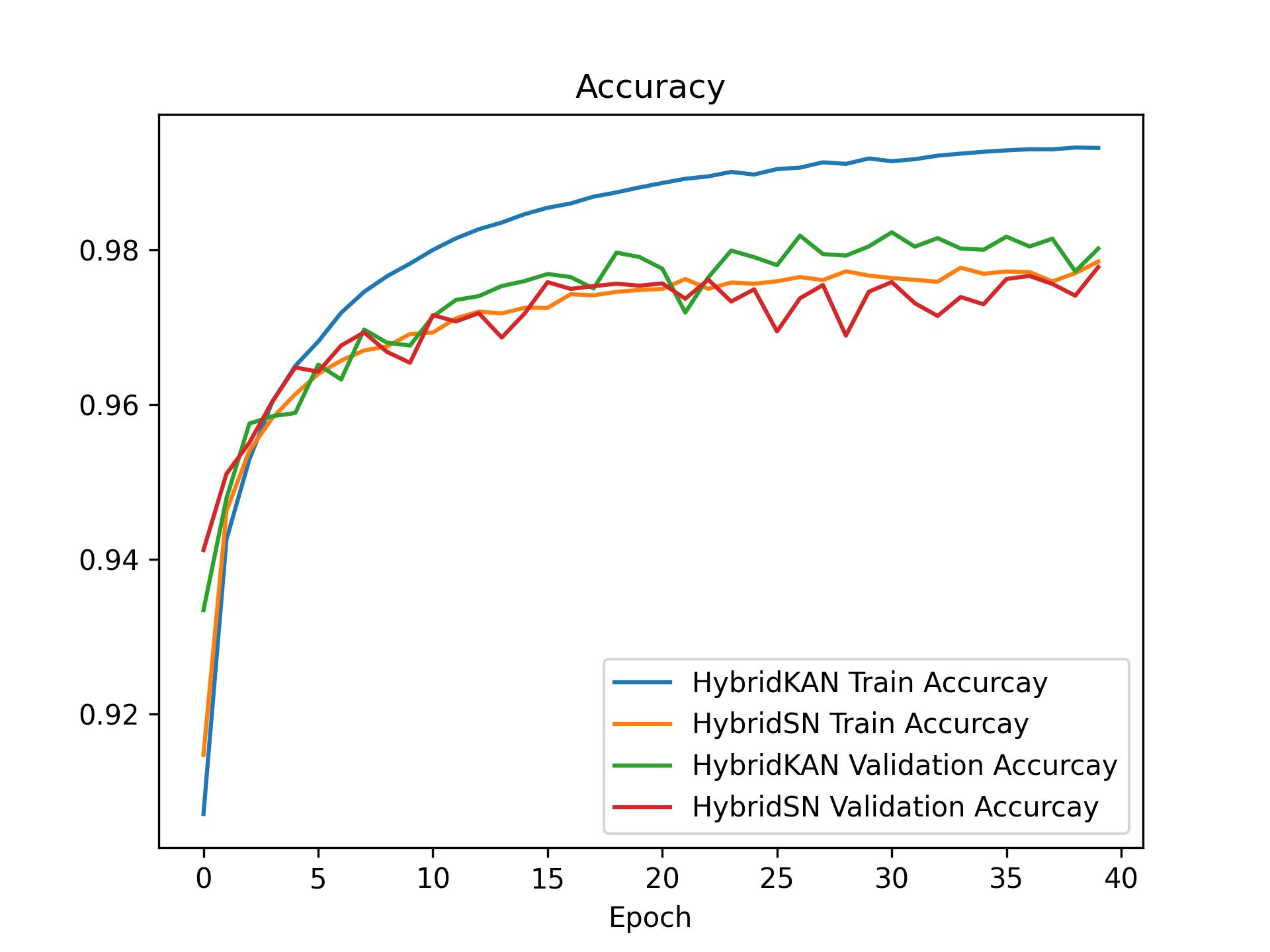} 
\caption{}
\end{subfigure}%
\caption{The convergence graph between HybridSN and its KAN version over the Qingyun HSI benchmark dataset for 40 epochs.}
\label{fig:convergenceQingyun}
\end{figure*}

\begin{figure*}[!ht]
\centering
\begin{subfigure}{.8\columnwidth}
\centering
\includegraphics[clip=true, trim = 10 0 10 10, width=0.98\textwidth]{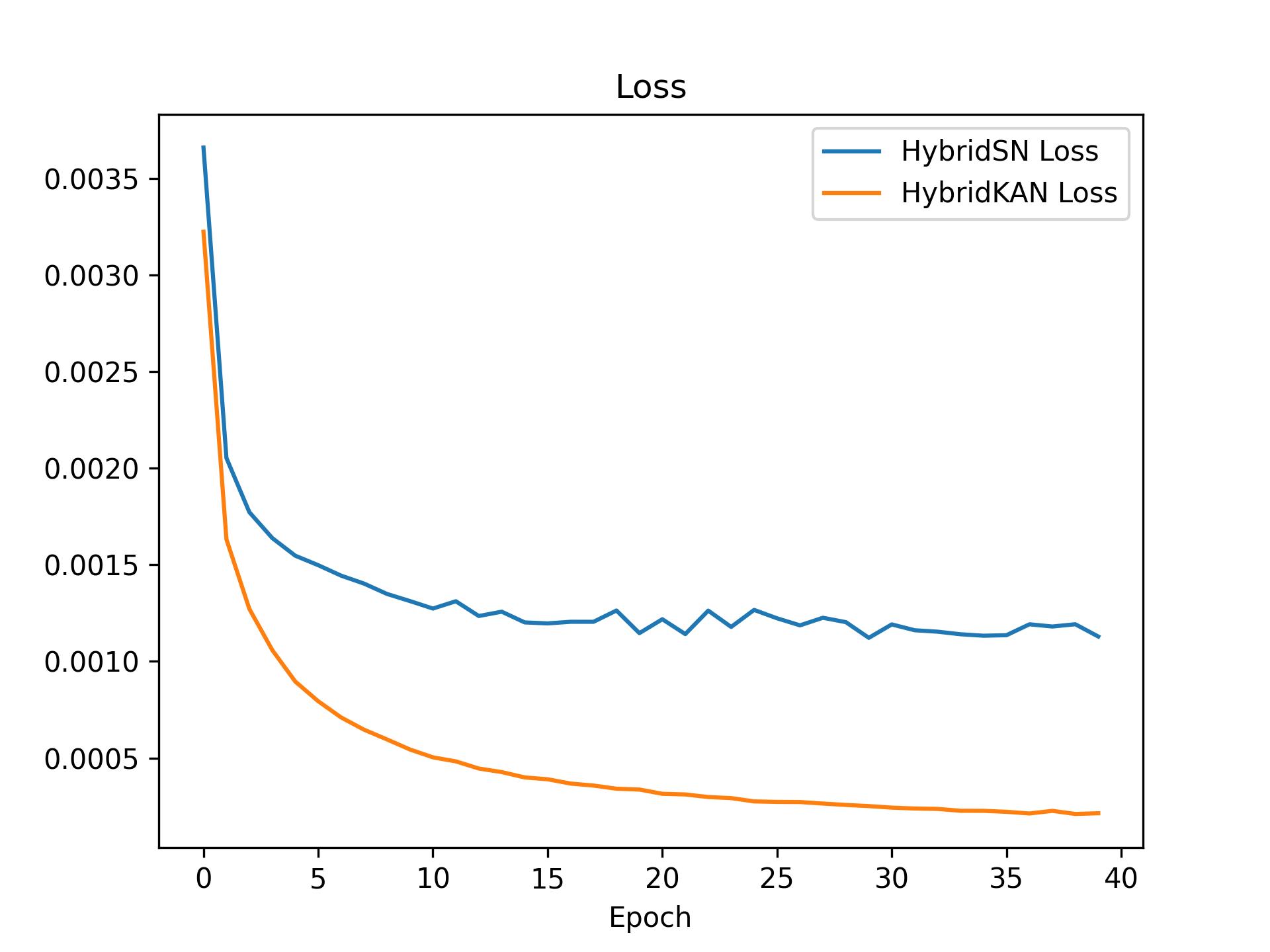} 
\caption{}
\end{subfigure}%
\begin{subfigure}{.8\columnwidth}
\centering
\includegraphics[clip=true, trim = 10 0 10 10, width=0.98\textwidth]{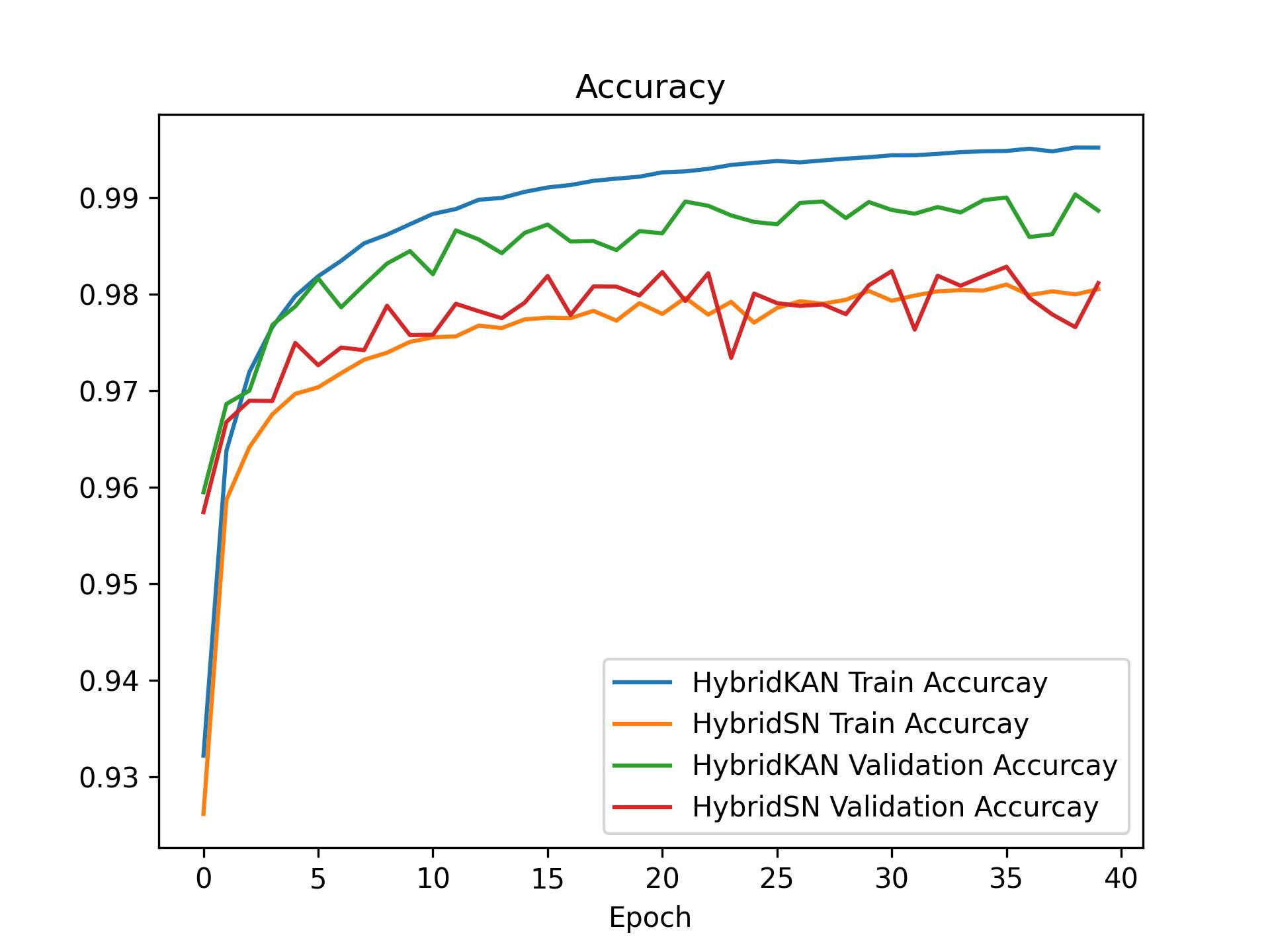} 
\caption{}
\end{subfigure}%
\caption{The convergence graph between HybridSN and its KAN version over the Pingan HSI benchmark dataset for 40 epochs.}
\label{fig:convergencePingan}
\end{figure*}

Due to the training process in deep learning may prove time-consuming and it's not always evident when the network has acquired sufficient information, convergence is an important tool. Nevertheless, when validation and training error ceases to decrease, a deep learning model is considered to have been converged. An ideal solution is not always guaranteed by convergence; this relies on several variables, including the network's architecture, the hyperparameters, and the quality of the HSI data. As seen in Figs \ref{fig:convergenceTangdaowan}, \ref{fig:convergenceQingyun}, and \ref{fig:convergencePingan}, the HybridKAN architecture, which utilizes KAN layers, is superior over the HybridSN using convolutional layers in terms of lower loss, higher train accuracy, and higher validation accuracy. The HyperKAN model requires a smaller number of epochs for its convergence which is vital in the remote sensing filed with the existence of high-dimensional and complex data. This proves the better capability of the developed HyperKAN model over the HybridSN classification algorithm.


\subsection{Feature Visualization of KAN using t-SNE}

\begin{figure*}[!ht]
\centering
\begin{subfigure}{.7\columnwidth}
\centering
\includegraphics[clip=true, trim = 40 40 40 40, width=0.98\textwidth]{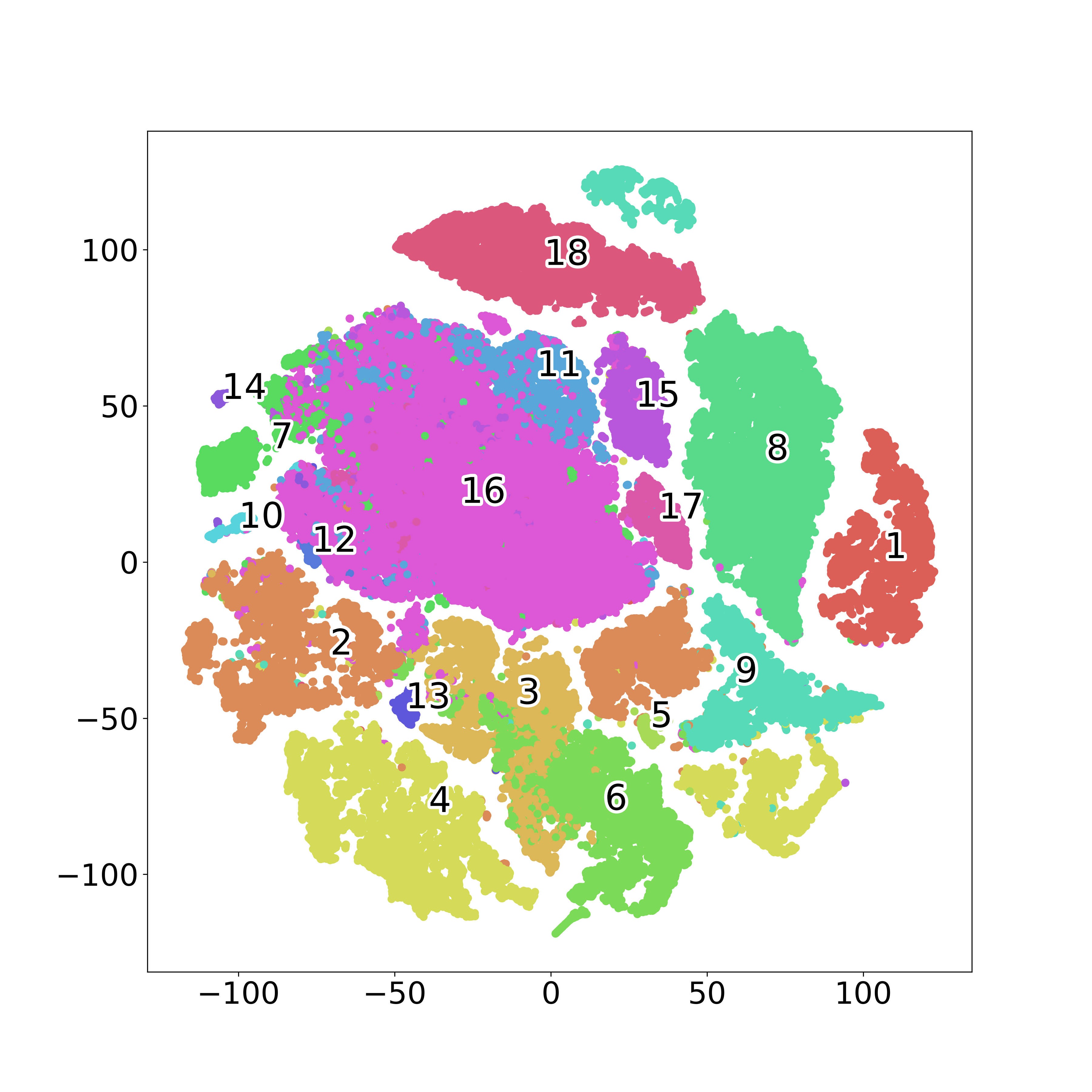} 
\caption{}
\end{subfigure}%
\begin{subfigure}{.7\columnwidth}
\centering
\includegraphics[clip=true, trim = 40 40 40 40, width=0.98\textwidth]{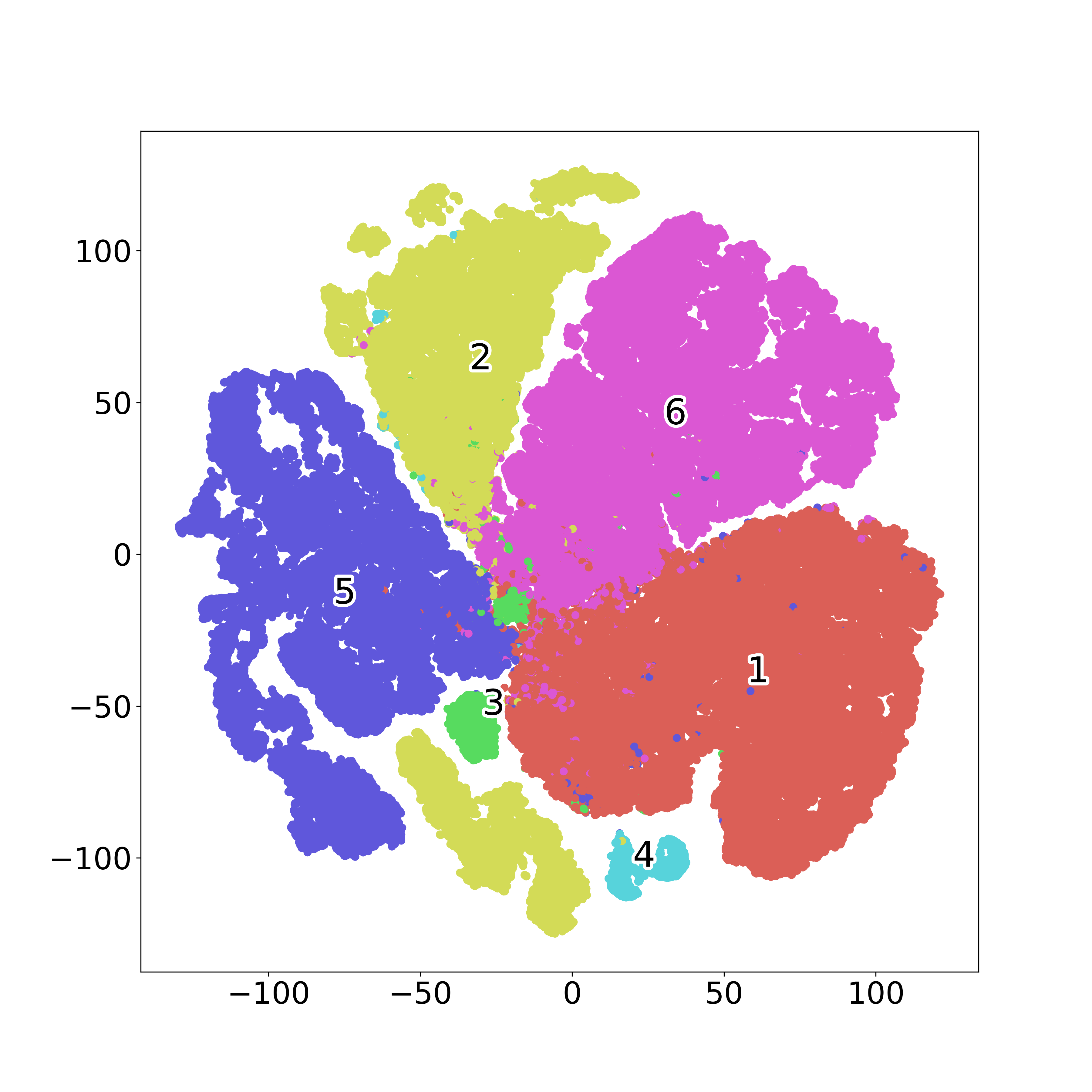} 
\caption{}
\end{subfigure}%
\begin{subfigure}{.7\columnwidth}
\centering
\includegraphics[clip=true, trim = 40 40 40 40, width=0.98\textwidth]{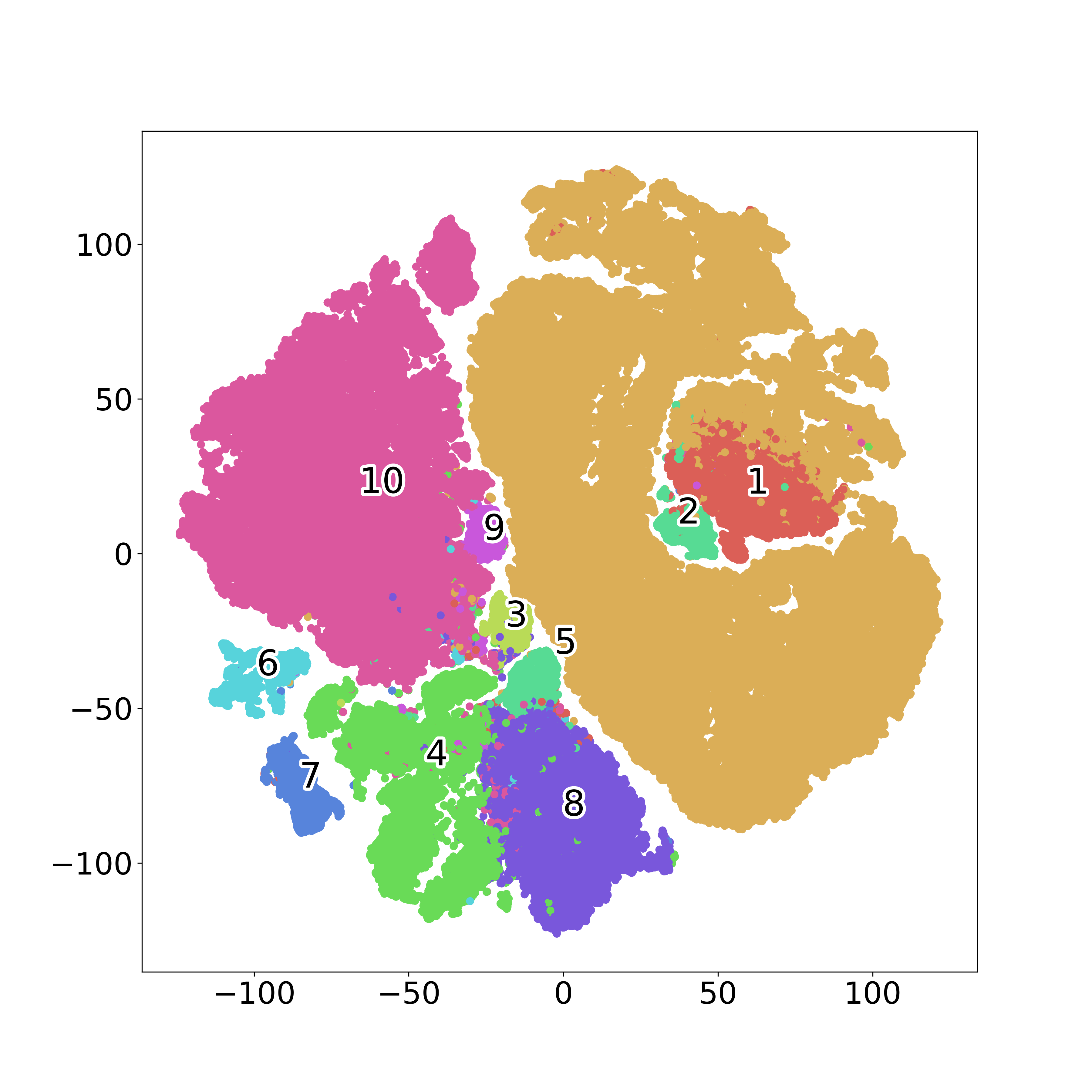} 
\caption{}
\end{subfigure}%
\caption{The t-SNE Visual presentation of the HybridKAN in data benchmark of a) Tangdaowan, b) Qingyun, and c) Pingan.}
\label{fig:tSNE}
\end{figure*}

The many spectral ranges that makeup HS data allow for the comprehensive capture of details over a large range of electromagnetic wavelengths. As such, it can be difficult to visualize these high-dimensional characteristics. Nevertheless, t-Distributed Stochastic Neighbour Embedding (t-SNE) \cite{van_der2008} may make it easier to observe the complex spectral-spatial features that the developed HybridKAN extracts in a two-dimensional space. To analyze the representational abilities of our model, this visualization is essential because it provides insights that may not be immediately clear from a direct examination of the raw data. The feature distributions for HybridKAN in 2D feature space by t-SNE are shown in Fig. \ref{fig:tSNE}. As can be seen in Fig.~\ref{fig:tSNE}, the developed KAN-based architecture demonstrated excellent feature separation capability in recognizing complex land covers in all three HSI data benchmarks, according to results obtained by the t-SNE algorithm. Furthermore, due to its weighted non-linear function rather than traditional MLPs with fixed non-linear activation functions, the HybridKAN's classification map, as shown in Fig. \ref{fig:Pingan}, \ref{fig:Qingyun}, and \ref{fig:Tangdaowan}, showed the least amount of noise and the most homogeneous classification map when compared to the other implemented algorithms. Furthermore, compared to the conventional ViT architecture, it is clear that HybridKAN's classification maps are far less noisy.

\subsection{Hyperparameter Sensitivity Analysis:}

The complexity and quantity of parameters in a classification algorithm are important factors to consider in the remote sensing field. In comparison with 2-D and 3-D KAN models, the HybridKAN model has more parameters, but this increase is justified. Higher OA, AA, and \emph{k} in Tables\ref{tab:Pingan}, \ref{tab:Qingyun}, and \ref{tab:Tangdaowan} demonstrate the considerable boost to the classification performance of the Hybrid model, which justifies the trade-off. Furthermore, as shown in Figs \ref{fig:Pingan}, \ref{fig:Tangdaowan}, and \ref{fig:Qingyun}, the visual classification maps produced by the HybridKAN architecture yield less noise and more homogeneous classification maps. This perspective emphasizes the notion that the trade-off of greater model complexity (as indicated by the greater number of parameters in Table \ref{tab:ComputationComplexity}) is matched with a demonstrable and supported improvement in the model's capacity to categorize high-spectrum imagery correctly.

\begin{table*}[!t]
\centering
\caption{Number of parameters in the developed classification algorithms. ( * To reduce the number of trainable parameters, VGG-16, EfficientNet, and ResNet-50 have been modified.)}
\resizebox{0.99\linewidth}{!}{
\begin{tabular}{|c|c|c|c|c|c|c|c|c|c|c|c|c|c|c|}
\hline
Model & 1DCNN & 2DCNN & 3DCNN& VGG16* \cite{simonyan2015deep}& ResNet50* \cite{He_2016_CVPR} & EfficientNet* \cite{Koonce2021} & RNN\cite{Mou752} & ViT\cite{Alexey20} & 1DKAN \cite{liu2024kan}& 2DCKAN  & 3DKAN  & HybridKAN\\ 
\hline
\hline
Total number of parameters &  29,170& 60,902 & 4,282 &  1,174,162& 211,826 & 177,406 &7,686  & 152,586& 565,458 & 14,743 & 50,826 & 135,090\\
Number of trainable parameters & 29,170 &  60,902& 4,282 &1,174,162  & 211,826 &  177,406& 7,686 & 152,586 &565,458& 14,743   &50,826  &135,090\\
Number of non-trainable parameters & 0 &  0&0  & 0 &  0&  0& 0 & 0& 0 & 0 &0  & 0\\
\hline
\hline
Forward/backward pass size (MB) & 0.01 & 0.04 & 0.01 &  0.64 & 0.07 &   0.29& 0.11 & 18.63 &0.07&  0.12 & 0.03 & 0.12\\
Params size (MB) &  0.11&   0.23&0.02  & 4.48 & 0.81 & 0.68 &0.03  &   0.58& 2.16 & 0.06&0.19 &0.52\\
Estimated Total Size (MB) & 0.12 &   0.27&0.03  & 5.12  & 0.88 &0.97  &  0.14&  19.22& 2.23&0.18 &0.23  &0.63\\

\hline
\end{tabular}}
\label{tab:ComputationComplexity}
\end{table*}

\section{Conclusion}
\label{sec:Concl}

This research proposed and discussed a KAN models-based architecture for complex land use land cover mapping using HSI data, which employs 1D, 2D, and 3D KAN models. The classification results on three highly complex HSI datasets demonstrate that the developed classification model, HybridKAN, was competitive or better statistically and visually over several other CNN- and ViT-based algorithms, including 1D-CNN, 2DCNN, 3D CNN, VGG-16, ResNet-50, EfficientNet, RNN, and ViT. The obtained results underscored the significant potential use of KAN models in complex remote sensing tasks. The HSI data classification ability of the proposed Hybrid KAN architecture compared to other CNN-and ViT-based classification models is shown over three HSI benchmark datasets: QUH-Pingan, QUH-Tangdaowan, and QUH-Qingyun. The results underscored the competitive or better capability of the developed hybrid model across these benchmark datasets compared to state-of-the-art classification architectures.

\ifCLASSOPTIONcaptionsoff
  \newpage
\fi

\bibliographystyle{IEEEtran}
\bibliography{IEEEabrv,Ref}

\end{document}